%% file: main.tex
\documentclass{article} %
\usepackage{titletoc}
\usepackage{iclr2023_conference,times}

\input{math_commands.tex}

\usepackage{hyperref}
\usepackage{url}
\usepackage{url}
\usepackage{amsmath, amssymb, amsthm, cleveref}
\usepackage{wrapfig}
\crefname{equation}{}{}
\Crefname{equation}{}{}

\newcommand{\comment}[1]{}
\usepackage[T1]{fontenc}    %
\usepackage{hyperref}       %
\usepackage{url}            %
\usepackage{booktabs}       %
\usepackage{amsfonts}       %
\usepackage{nicefrac}       %
\usepackage{microtype}      %
\usepackage{algorithm}
\usepackage{algorithmic}

\usepackage[utf8]{inputenc}
\usepackage{mathtools}
\usepackage{subcaption}
\usepackage{graphicx}
\usepackage{caption}
\usepackage{hyperref}
\usepackage{bbm}
\usepackage[shortlabels]{enumitem}
\usepackage{array}
\usepackage{arydshln}

\usepackage{multirow}
\crefname{equation}{}{}
\Crefname{equation}{}{}

\usepackage{url}            %
\usepackage{booktabs}       %
\usepackage{amsfonts}       %
\usepackage{nicefrac}       %
\usepackage{xcolor}
\usepackage{microtype}      %
\usepackage{amssymb,amsmath, amsthm}
\usepackage[utf8]{inputenc}
\usepackage{mathtools}
\usepackage{subcaption}
\usepackage{graphicx}
\usepackage{caption}
\usepackage{bbm}
\usepackage{bm}
\usepackage{booktabs}
\usepackage{multirow}
\usepackage{siunitx}

\usepackage[shortlabels]{enumitem}
\usepackage{array}
\usepackage{arydshln}
\newtheorem{lem}{Lemma}
\newtheorem{thm}{Theorem}

\newtheorem{assum}{Assumption}

\makeatletter
\def\thickhline{%
  \noalign{\ifnum0=`}\fi\hrule \@height \thickarrayrulewidth \futurelet
   \reserved@a\@xthickhline}
\def\@xthickhline{\ifx\reserved@a\thickhline
               \vskip\doublerulesep
               \vskip-\thickarrayrulewidth
             \fi
      \ifnum0=`{\fi}}
\makeatother

\newlength{\thickarrayrulewidth}
\def\HiLi{\leavevmode\rlap{\hbox to \hsize{\color{yellow!50}\leaders\hrule height .8\baselineskip depth .5ex\hfill}}}

\usepackage{xcolor} %
\usepackage{tikz}
\usetikzlibrary{fit,calc}
\newcommand*{\tikzmk}[1]{\tikz[remember picture,overlay,] \node (#1) {};}
\newcommand{\boxit}[1]{\tikz[remember picture,overlay]{\node[yshift=0pt,fill=#1,opacity=.25,fit={($(A)+(0pt,9pt)$)($(B)+(0,0)$)}] {};}}
\colorlet{pink}{red!40}
\colorlet{cyan}{cyan!60}

\title{FedExP: Speeding Up Federated Averaging via Extrapolation}

\author{Divyansh Jhunjhunwala\textsuperscript{1}, Shiqiang Wang\textsuperscript{2}, Gauri Joshi\textsuperscript{1} \\
\textsuperscript{1}Carnegie Mellon University, \textsuperscript{2}IBM Research \\
\texttt{\{djhunjhu, gaurij\}@andrew.cmu.edu}, \,\, \texttt{wangshiq@us.ibm.com}
\vspace{-1em}
}

\iclrfinalcopy %
\begin{document}

\maketitle

\begin{abstract}
Federated Averaging (\texttt{FedAvg}) remains the most popular algorithm for Federated Learning (FL) optimization due to its simple implementation, stateless nature, and privacy guarantees combined with secure aggregation. Recent work has sought to generalize the vanilla averaging in \texttt{FedAvg} to a generalized gradient descent step by treating client updates as pseudo-gradients and using a server step size. While the use of a server step size has been shown to provide performance improvement theoretically, the practical benefit of the server step size has not been seen in most existing works. In this work, we present \texttt{FedExP}, a method to adaptively determine the server step size in FL based on dynamically varying pseudo-gradients throughout the FL process. We begin by considering the overparameterized convex regime, where we reveal an interesting similarity between \texttt{FedAvg} and the Projection Onto Convex Sets (\texttt{POCS}) algorithm. We then show how \texttt{FedExP} can be motivated as a novel extension to the \textit{extrapolation mechanism} that is used to speed up POCS. Our theoretical analysis later also discusses the implications of \texttt{FedExP} in underparameterized and non-convex settings. Experimental results show that \texttt{FedExP} consistently converges faster than \texttt{FedAvg} and competing baselines on a range of realistic FL datasets. 
\end{abstract}

\section{Introduction}
\label{sec:introduction}

Federated Learning (FL) has emerged as a key distributed learning paradigm in which a central server orchestrates the training of a machine learning model across a network of devices. FL is based on the fundamental premise that data never leaves a clients device, as clients only communicate model updates with the server. Federated Averaging or \texttt{FedAvg}, first introduced by \citet{mcmahan2017communication}, remains the most popular algorithm in this setting due to the simplicity of its implementation, stateless nature (i.e., clients do not maintain local parameters during training) and the ability to incorporate privacy-preserving protocols such as secure aggregation \citep{bonawitz2016practical, kadhe2020fastsecagg}.
\paragraph{Slowdown Due to Heterogeneity.} One of the most persistent problems in \texttt{FedAvg} is the slowdown in model convergence due to data heterogeneity across clients. Clients usually perform multiple steps of gradient descent on their heterogeneous objectives before communicating with the server in \texttt{FedAvg}, which leads to what is colloquially known as \textit{client drift error} \citep{karimireddy2019error}. The effect of heterogeneity is further exacerbated by the constraint that only a fraction of the total number of clients may be available for training in every round \citep{kairouz2019advances}. Various techniques have been proposed to combat this slowdown, among the most popular being variance reduction techniques such as \citet{karimireddy2019error, mishchenko2022proxskip, mitra2021linear}, but they either lead to clients becoming stateful, add extra computation or communication requirements or have privacy limitations. 
\paragraph{Server Step Size.}
Recent work has sought to deal with this slowdown by using two separate step sizes in \texttt{FedAvg} -- a \textit{client step size} used by the clients to minimize their local objectives and a \textit{server step size} used by the server to update the global model by treating client updates as pseudo-gradients \citep{karimireddy2019error, reddi2020adaptive}. To achieve the fastest convergence rate, these works propose keeping the client step size as $\mathcal{O}\big(1/\tau\sqrt{T}\big)$ and the server step size as $\mathcal{O}\big(\sqrt{\tau M}\big)$, where $T$ is the number of communication rounds, $\tau$ is the number of local steps and $M$ is the number of clients. Using a small client step size mitigates client drift, and a large server step size prevents global slowdown. While this idea may be asymptotically optimal, it is not always effective in practical non-asymptotic and communication-limited settings \citep{charles2020outsized}.
\begin{wrapfigure}{r}{0.3\textwidth}
\centering 
\includegraphics[width=0.3\textwidth]{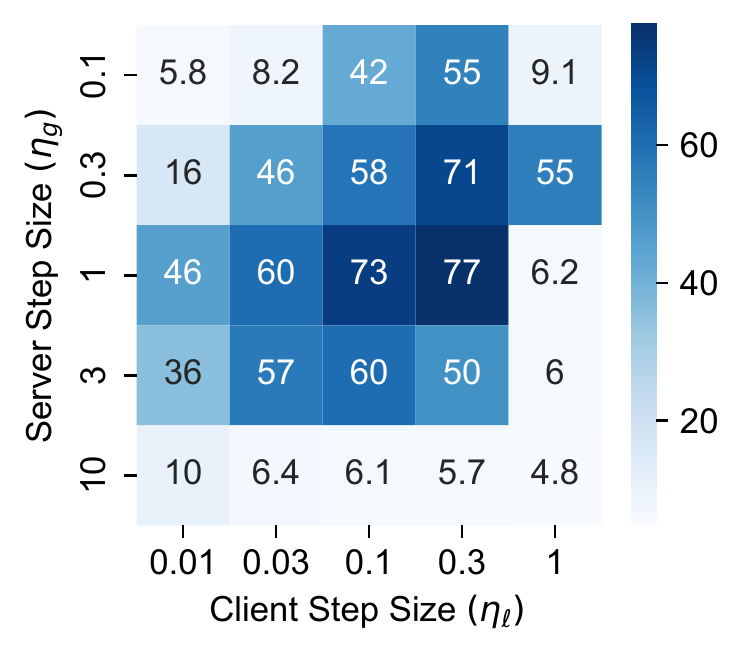}\vspace{-0.7em}\caption{Test accuracy ($\%$) achieved by different server and client step sizes on \mbox{EMNIST} dataset \citep{cohen2017emnist} after 50 rounds (details of experimental setup are in \Cref{sec:experiments} and \Cref{app:sec_experiments}).}
\vspace{-1.5em}
\label{fig:matrix_test_acc}
\end{wrapfigure}
In practice, a small client step size severely slows down convergence in the initial rounds and cannot be fully compensated for by a large server step size (see \Cref{fig:matrix_test_acc}). Also, if local objectives differ significantly, then it may be beneficial to use smaller values of the server step size \citep{malinovsky2022server}.

Therefore, we seek to answer the following question: \textit{For a moderate client step size, can we adapt the server step size according to the local progress made by the clients and the heterogeneity of their objectives?} 
In general, it is challenging to answer this question because %
it is difficult to obtain knowledge of the heterogeneity between the local objectives and appropriately use it to adapt the server step size. %

\paragraph{Our Contributions. }
In this paper, we take a novel approach to address the question posed above. We begin by considering the case where the models are \textit{overparameterized}, i.e., the number of model parameters is larger than the total number of data points across all clients. This is often true for modern deep neural network models \citep{zhang2017understanding,jacot2018neural} and the small datasets collected by edge clients in the FL setting. %
In this overparameterized regime, the global minimizer becomes a \textit{common minimizer} for all local objectives, even though they may be arbitrarily heterogeneous. Using this fact, we obtain a \textit{novel connection} between \texttt{FedAvg} and the Projection Onto Convex Sets (\texttt{POCS}) algorithm, which is used to find a point in the intersection of some convex sets. 

Based on this connection, 
we find an interesting analogy between the server step size and the \textit{extrapolation parameter} that is used to speed up \texttt{POCS} \citep{pierra1984decomposition}. We propose new extensions to the extrapolated \texttt{POCS} algorithm to support inexact and noisy projections as in \texttt{FedAvg}. In particular, we derive a \textit{time-varying} bound on the progress made by clients towards the global minimum and show how this bound can be used to adaptively estimate a good server step size at each round. 
The result is our proposed algorithm \texttt{FedExP}, which is a method to adaptively determine the server step size in each round of FL based on the pseudo-gradients in that round.

Although motivated by the overparameterized regime, our proposed \texttt{FedExP} algorithm performs well (both theoretically and empirically) in the general case, where %
the model can be either overparameterized or underparameterized. For this general case, we derive the convergence upper bounds for both convex and non-convex objectives. %
Some highlights of our work are as follows.
 \begin{itemize}[leftmargin=*]
     \item We reveal a novel connection between \texttt{FedAvg} and the \texttt{POCS} algorithm for finding a point in the intersection of convex sets. %
     \item The proposed \texttt{FedExP} algorithm is simple to implement with virtually no additional communication, computation, or storage required at clients or the server. It is well suited for both cross-device and cross-silo FL, and is compatible with partial client participation.
     \item Experimental results show that \texttt{FedExP} converges $1.4$--$2\times$ faster than \texttt{FedAvg} and most competing baselines on standard FL tasks. 
 \end{itemize}

 \paragraph{Related Work. } \hspace{-1em} Popular algorithms for adaptively tuning the step size when training neural networks include \texttt{Adagrad} \citep{duchi2011adaptive} and its variants \texttt{RMSProp} \citep{tieleman2012lecture} and \texttt{Adadelta} \citep{zeiler2012adadelta}. These algorithms consider the notion of \textit{coordinate-wise adaptivity} and adapt the step size separately for each dimension of the parameter vector based on the magnitude of the accumulated gradients. While these algorithms can be extended to the federated setting using the concept of pseudo-gradients as done by \cite{reddi2020adaptive}, these extensions are agnostic to inherent data heterogeneity across clients, which is central to FL. On the contrary, \texttt{FedExP} is explicitly designed for FL settings and uses a \textit{client-centric} notion of adaptivity that utilizes the heterogeneity of client updates in each round. The work closest to us is \cite{johnson2020adascale}, which proposes a method to adapt the step size for large-batch training by estimating the \textit{gradient diversity} \citep{yin2018gradient} of a minibatch. This result has been improved in a recent work by \cite{horvath2022adaptive}. However, both \cite{johnson2020adascale,horvath2022adaptive} focus on the centralized setting. In \texttt{FedExP}, we use a similar concept, but within a \textit{federated} environment which comes with a stronger theoretical motivation, since client data are inherently diverse in this case.  
 We defer a more detailed discussion of other adaptive step size methods and related work to \Cref{app_sec:related_work}.

\section{Problem Formulation and Preliminaries}

As in most standard federated learning frameworks, we consider the problem of optimizing the model parameters $\bw \in \mathbb{R}^{d}$ to minimize the global objective function $F(\bw)$ defined as follows:
\begin{align}
    \min_{\bw \in \mathbb{R}^d} F(\bw) := \frac{1}{M}\sum_{i=1}^M F_i(\bw),
\label{eq:prob_form}
\end{align}%
where $F_i(\bw) := \frac{1}{|\mathcal{D}_i |} \sum_{\delta_i \in \mathcal{D}_i} \ell(\bw,\delta_i)$ is the empirical risk objective computed on the local data set $\mathcal{D}_i$ at the the $i$-th client. Here, $\ell(\cdot, \cdot)$ is a loss function and $\delta_i$ represents a data sample from the empirical local data distribution $\mathcal{D}_i$. The total number of clients in the FL system is denoted by $M$. Without loss of generality, we assume that all the $M$ client objectives are given equal weight in the global objective function defined in \Cref{eq:prob_form}. Our algorithm and analysis can be directly extended to the case where client objectives are unequally weighted, e.g., proportional to local dataset sizes~$|\mathcal{D}_i|$.

 \paragraph{\texttt{FedAvg}.} We focus on solving \cref{eq:prob_form} using \texttt{FedAvg} \citep{mcmahan2017communication, kairouz2019advances}. At round $t$ of \texttt{FedAvg}, the server sends the current global model $\bw^{(t)}$ to all clients. Upon receiving the global model, clients perform $\tau$ steps of local stochastic gradient descent (SGD) to compute their updates $\{\Delta_i^{(t)}\}_{i=1}^M$ for round $t$ as follows.
\begin{align}
   & \text{Perform Local SGD:}\quad \bw_i^{(t,k+1)} = \bw_i^{(t,k)} - \eta_l \nabla F_i(\bw_i^{(t,k)},\xi^{(t,k)}) \hspace{10pt} \forall k \in \{0,1,\dots, \tau-1\}
   \label{eq:local_sgd_update}\\
    & \text{Compute Local Difference:}\quad  \Delta_i^{(t)} = \bw^{(t)} - \bw_i^{(t,\tau)} \label{eq:local_update}
\end{align}
where $\bw_i^{(t,0)} = \bw^{(t)}$ for all $i \in [M]$, $\eta_l$ is the client step size and $\nabla F_i(\bw_i^{(t,k)},\xi^{(t,k)})$ represents a stochastic gradient computed on the minibatch $\xi_i^{(t,k)}$ sampled randomly from $\mathcal{D}_i$. 

\paragraph{Server Optimization in \texttt{FedAvg}.} \hspace{-8pt} In vanilla \texttt{FedAvg} \citep{mcmahan2017communication}, the global model would simply be updated as the average of the client local models, that is, $\bw^{(t+1)} = \frac{1}{M}\sum_{i=1}^M \bw_i^{(t,\tau)}$. To improve over this, recent work \citep{reddi2020adaptive, hsu2019measuring} has focused on optimizing the server aggregation process by treating the client updates $\Delta_i^{(t)}$ as ``pseudo-gradients'' and multiplying by a server step size when aggregating them as follows.
\begin{align}
    \text{Generalized \texttt{FedAvg} Global Update:} \hspace{20pt} \bw^{(t+1)} = \bw^{(t)} - \eta_g \bardeltat
    \label{eq:gen_fedavg_update}
\end{align}
where $\bardeltat = \frac{1}{M}\sum_{i=1}^M \Delta_i^{(t)}$ is the aggregated client update in round $t$ and $\eta_g$ acts as \textit{server step size}. Note that setting $\eta_g = 1$ recovers the vanilla \texttt{FedAvg} update.

While the importance of the server step size has been theoretically well established in these works, we find that its practical relevance has not been explored. In this work, we take a step towards bridging this gap between theory and practice by \textit{adaptively} tuning the value of $\eta_g$ that we use in every round.

\section{Proposed Algorithm: FedExP} 
\label{sec:proposed_algorithm}

Before discussing our proposed algorithm, we first highlight a useful and novel connection between \texttt{FedAvg} and the \texttt{POCS} algorithm used to find a point in the intersection of some convex sets.

\subsection{Motivation for Extrapolation}

\paragraph{Connection Between \texttt{FedAvg} and \texttt{POCS} in the Overparameterized Convex Regime.} \hspace{-8pt} Consider the case where the local objectives of the clients $\{F_i(\bw)\}_{i=1}^M$ are convex. In this case, we know that the set of minimizers of $F_i(\bw)$ given by $\mathcal{S}_i^* = \{\bw : \bw \in \argmin F_i(\bw)\}$ is also a convex set for all $i \in [M]$. Now let us assume that we are in the \textit{overparameterized regime} where $d$ is sufficiently larger than the total number of data points across clients. In this regime, the model can fit all the training data at clients simultaneously and hence be a minimizer for all local objectives. Thus we assume that the global minimum satisfies $\bw^* \in \mathcal{S}_i^* , \forall i \in [M]$. 
Our original problem in \cref{eq:prob_form} can then be reformulated as trying to \textit{find a point in the intersection of convex sets} $\{\mathcal{S}_i^*\}_{i=1}^M$ since $\bw^* \in \mathcal{S}_i^* , \forall i \in [M]$. One of the most popular algorithms to do so is the  \textit{Projection Onto Convex Sets} (\texttt{POCS}) algorithm \citep{gurin1967method}. In \texttt{POCS}, at every iteration the current model is updated as follows\footnote{We refer here to a parallel implementation of \texttt{POCS}. This is also known as Parallel Projection Method (\texttt{PPM}) and Simultaneous Iterative Reconstruction Technique (\texttt{SIRT}) in some literature \citep{combettes1997convex}.}.
\begin{align}
\textstyle
    \text{Generalized \texttt{POCS} update: } \hspace{20pt} \bw^{(t+1)}_{\texttt{POCS}} = \bw^{(t)}_{\texttt{POCS}} - \lambda \left(\frac{1}{M}\sum_{i=1}^M P_i(\bw^{(t)}_{\texttt{POCS}})-\bw^{(t)}_{\texttt{POCS}}\right)
\end{align}
where $P_i(\bw^{(t)}_{\texttt{POCS}})$ is a projection of $\bw^{(t)}_{\texttt{POCS}}$ on the set $\mathcal{S}_i^*$ and $\lambda$ is known as the relaxation coefficient \citep{combettes1997convex}. 

\paragraph{Extrapolation in \texttt{POCS}. } \citet{combettes1997convex} notes that \texttt{POCS} has primarily been used with $\lambda = 1$, with studies failing to demonstrate a systematic benefit of $\lambda < 1$ or $\lambda > 1$ \citep{mandel1984convergence}. This prompts \citet{combettes1997convex} to study an \textit{adaptive} method of setting $\lambda$, first introduced by \citet{pierra1984decomposition} as follows:
\vspace{-0.5em}
\begin{align*}
    \lambda^{(t)} = \frac{\sum_{i=1}^M \norm{P_i(\bw^{(t)})-\bw^{(t)}}}{M\norm{\frac{1}{M}\sum_{i=1}^M P_i(\bw^{(t)})-\bw^{(t)}}}\, .
\end{align*}
\citet{pierra1984decomposition} refer to the \texttt{POCS} algorithm with this adaptive $\lambda^{(t)}$ as Extrapolated Parallel Projection Method (\texttt{EPPM}). This is referred to as extrapolation since we always have $\lambda^{(t)} \geq 1$ by Jensen's inequality. The intuition behind \texttt{EPPM} lies in showing that the update with the proposed $\lambda^{(t)}$ always satisfies $\normsmall{\bw^{(t+1)}_{\texttt{POCS}}-\bw^*} < \normsmall{\bw^{(t)}_{\texttt{POCS}}-\bw^*}$, thereby achieving asymptotic convergence. Experimental results in \cite{pierra1984decomposition} and \cite{combettes1997convex} show that \texttt{EPPM} can give an order-wise speedup over \texttt{POCS}, motivating us to study this algorithm in the FL context.

\subsection{Incorporating Extrapolation in FL}
\label{subsection:extrapolation in FL}
 Note that to implement \texttt{POCS} we do not need to explicitly know the sets $\{\mathcal{S}_i^*\}_{i=1}^M$; we only need to know how to compute a \textit{projection} on these sets. From this point of view, we see that \texttt{FedAvg} proceeds similarly to \texttt{POCS}. In each round, clients receive $\bw^{(t)}$ from the server and run multiple SGD steps to compute an ``approximate projection'' $\bw_i^{(t,\tau)}$ of $\bw^{(t)}$ on their solution sets $\mathcal{S}_i^*$. These approximate projections are then aggregated at the server to update the global model. In this case, the relaxation coefficient $\lambda$ plays exactly the same role as the server step size $\eta_g$ in \texttt{FedAvg}.

Inspired by this observation and the idea of extrapolation in \texttt{POCS}, we seek to understand if a similar idea can be applied to tune the server step size $\eta_g$ in \texttt{FedAvg}. Note that the \texttt{EPPM} algorithm makes use of exact projections to prove convergence which is not available to us in FL settings. This is further complicated by the fact that the client updates are noisy due to the stochasticity in sampling minibatches. We find that in order to use an \texttt{EPPM}-like step size the use of exact projections can be relaxed to the following condition, which bounds the distance of the local models from the global minimum as follows.
\begin{align}
\text{Approximate projection condition in FL:}\hspace{10pt}
\label{prop_1}
\textstyle
    \frac{1}{M}\sum_{i=1}^M \norm{\bw_i^{(t,\tau)}-\bw^*} \leq \norm{\bw^{(t)}-\bw^*}
\end{align}
where $\bw^{(t)}$ and $\{\bw_i^{(t,\tau)}\}_{i=1}^{M}$ are the global and local client models, respectively, at round $t$ and $\bw^*$ is a global minimum. 
Intuitively, this condition suggests that after the local updates, the local models are closer to the optimum $\bw^*$ on average as compared to model $\bw^{(t)}$ at the beginning of that round.
We first show that this condition~\Cref{prop_1} holds in the overparameterized convex regime under some conditions. The full proofs for lemmas and theorems in this paper are included in \Cref{appendix:sec_proofs}.
\begin{lem}
\label{lem_1}
Let $F_i(\bw)$ be convex and $L$-smooth for all $i \in [M]$ and let $\bw^*$ be a common minimizer of all $F_i(\bw)$. Assuming clients run full-batch gradient descent to minimize their local objectives with $\eta_l \leq 1/L$, then \Cref{prop_1}  holds for all $t$ and $\tau \geq 1$. 
\end{lem}

In the case with stochastic gradient noise or when the model is underparameterized, although \Cref{prop_1} may not hold in general, we expect it to be satisfied at least during the initial phase of training when $\norm{\bw^{(t)}-\bw^*}$ is large and clients make common progress towards a minimum.

\setlength{\textfloatsep}{1em}%
\begin{algorithm} [t]
\caption{Proposed Algorithm: \texttt{FedExP} }\label{algo1}
\renewcommand{\algorithmicloop}{\textbf{Global server do:}}
\begin{algorithmic}[1]
\STATE {\bfseries Input:} $\bw^{(0)}$, number of rounds $T$, local iteration steps $\tau$, parameters $\eta_l,\epsilon$
\STATE {\bfseries For ${t=0,\ldots,T-1}$ communication rounds do}:
\STATE \hspace*{1em} {\bfseries Global server does:}\\
\STATE \hspace*{1em} Send $\bw^{(t)}$ to all clients
\STATE \hspace*{1em} {\bfseries Clients $i \in [M]$ in parallel do:}
\STATE \hspace*{2em} {Set $\bw_i^{(t,0)}\leftarrow \bw^{(t,0)}$}
\STATE {\hspace*{2em} {\bfseries For $k=0,\ldots,\tau-1$ local iterations do:}}\\
\STATE {\hspace*{3em} Update $\bw_i^{(t,k+1)}\leftarrow\bw_i^{(t,k)}-\eta_l\nabla F_i(\bw_i^{(t,k)},\xi_i^{(t,k)})$}\\
\STATE {\hspace*{2em} Send $\Delta_i^{(t)}\leftarrow \bw^{(t)}-\bw_i^{(t,\tau)}$ to the server}
\STATE {\hspace*{1em} {\bfseries Global server does:}}\\
\STATE {\hspace*{2em} \tikzmk{A}Compute $\bardeltat \!\leftarrow\!  \frac{1}{M}\sum_{i=1}^M \Delta_i^{(t)}$ and $\eta_g^{(t)} \!\leftarrow\!  \max\left\{1, \sum_{i=1}^M \normsmall{\Delta_i^{(t)}}\!\!\Big/2M\!\left(\norm{\bardeltat}\!+\!\epsilon \right) \!\right\}$}
\STATE {\hspace*{2em} Update global model with $\bw^{(t+1)}\leftarrow \bw^{(t)}-\eta_g^{(t)}\bardeltat$} \hfill \tikzmk{B} \boxit{pink}
\end{algorithmic} 
\end{algorithm} 

Given that \Cref{prop_1} holds, we now consider the generalized \texttt{FedAvg} update with a server step size $\eta_g^{(t)}$ in round $t$. Our goal is to find the value of $\eta_g^{(t)}$ that \textit{minimizes} the distance of $\bw^{(t+1)}$ to $\bw^*$:
\begin{align}
\textstyle
      \norm{\bw^{(t+1)}-\bw^*}  = \norm{\bw^{(t)}-\bw^*} + (\eta_g^{(t)})^2 \norm{\bardeltat} -2 \eta_g^{(t)} \left \langle \bw^{(t)}-\bw^*, \bardeltat \right \rangle .
      \label{eq:dist_w_FedExP}
\end{align}
Setting the derivative of the RHS of \cref{eq:dist_w_FedExP} to zero we have,
\begin{align}
\textstyle
    (\eta_g^{(t)})_{\text{opt}} &=  %
    \frac{\left \langle \bw^{(t)}-\bw^*, \bardeltat \right \rangle}{\norm{\bardeltat}} = \frac{\sum_{i=1}^M \left \langle \bw^{(t)} - \bw^*,\Delta_i^{(t)} \right\rangle}{M\norm{\bardeltat}} \geq \frac{\sum_{i=1}^M \normsmall{\Delta_i^{(t)}}}{2M \norm{\bardeltat}}
    \label{eq:eta_g_lb},
\end{align}
where the last inequality follows from $\langle \ba,\bb \rangle = \frac{1}{2}[ \norm{\ba} + \norm{\bb} - \norm{\ba-\bb}]$, definition of $\Delta_i^{(t)}$ in \Cref{eq:local_update} and \Cref{prop_1}. 
Note that depending on the values of $\{ \Delta_i^{(t)}\}_{i=0}^M$, we may have $(\eta_g^{(t)})_{\text{opt}} \gg 1$. Thus, we see that \Cref{prop_1} acts as a suitable replacement for projection to justify the use of extrapolation in FL settings.

\subsection{Proposed Algorithm}

Motivated by our findings above, we propose the following server step size for the generalized \texttt{FedAvg} update at each round:
\begin{align}
    (\eta_g^{(t)})_{\texttt{FedExP}} = \max\left\{1, \,\,\frac{\sum_{i=1}^M \normsmall{\Delta_i^{(t)}}}{2M(\norm{\bardeltat}+\epsilon)} \right\}.
    \label{eq:eta_g_fedexp}
\end{align}
We term our algorithm \textit{Federated Extrapolated Averaging} or \texttt{FedExP}, in reference to the original \texttt{EPPM} algorithm which inspired this work. 
Note that our proposed step size satisfies the property that $\big|(\eta_g^{(t)})_{\text{opt}} - (\eta_g^{(t)})_{\text{\texttt{FedExP}}}\big| \leq \big| (\eta_g^{(t)})_{\text{opt}} -1 \big|$ when \Cref{prop_1} holds, which can be seen by comparing \cref{eq:eta_g_lb} and \cref{eq:eta_g_fedexp}. Since \cref{eq:dist_w_FedExP} depends quadratically on $\eta_g^{(t)}$, we can show that in this case $\normsmall{\bw^{(t+1)} - (\eta_g^{(t)})_{\texttt{FedExP}}\bardeltat - \bw^*} \leq \normsmall{\bw^{(t+1)} - \bw^*}$, implying we are at least as close to the optimum as the \texttt{FedAvg} update. In the rest of the paper, we denote $(\eta_g^{(t)})_{\texttt{FedExP}}$ as $\eta_g^{(t)}$ when the context is clear.

\paragraph{Importance of Adding Small Constant to Denominator.} \hspace{-8pt} In the case where \Cref{prop_1} does not hold, using the lower bound established in \Cref{eq:eta_g_lb} can cause the proposed step size to blow up. This is especially true towards the end of training where we can have $\norm{\bardeltat} \approx 0$ but $\norm{\Delta_i^{(t)}} \neq 0$. Thus we propose to add a small positive constant $\epsilon$ to the denominator in \Cref{eq:eta_g_fedexp} to prevent this blow-up.
For a large enough $\epsilon$ our algorithm reduces to \texttt{FedAvg} and therefore tuning $\epsilon$ can be a useful tool to interpolate between vanilla averaging and extrapolation. Similar techniques exist in adaptive algorithms such as \texttt{Adam} \citep{kingma2014adam} and \texttt{Adagrad} \citep{duchi2011adaptive} to improve stability.

\paragraph{Compatibility with Partial Client Participation and Secure Aggregation.} Note that \texttt{FedExP} can be easily extended to support partial participation of clients by calculating $\eta_g^{(t)}$ using only the updates of participating clients, i.e., the averaging and division in \Cref{eq:eta_g_fedexp} will be only over the clients that participate in the round. Furthermore, since the server only needs to estimate the average of pseudo-gradient norms, $\eta_g^{(t)}$ can be computed with secure aggregation, similar to computing $\bardeltat$.

\paragraph{Connection with Gradient Diversity. } We see that our lower bound on $(\eta_g^{(t)})_{\text{opt}}$ naturally depends on the similarity of the client updates with each other.
In the case where $\tau=1$ and clients run full-batch gradient descent, our lower bound \Cref{eq:eta_g_lb} reduces to $\sum_{i=1}^M \norm{\nabla F_i(\bw^{(t)})}\big/2M\norm{\nabla F(\bw^{(t)})}$ which is used as a measure of data-heterogeneity in many FL works \citep{wang2020tackling, haddadpour2019convergence}. 
Our lower bound suggests using larger step-sizes as this gradient diversity increases, which can be a useful tool to speed up training in heterogeneous settings. This is an orthogonal approach to existing optimization methods to tackle heterogeneity such as \citet{karimireddy2020scaffold, li2020federated, acar2021federated}, which propose additional regularization terms or adding control variates to the local client objectives to limit the impact of heterogeneity.

\section{Convergence Analysis}
\label{sec:convergence}
Our analysis so far has focused on the overparameterized convex regime to motivate our algorithm. In this section we discuss the convergence of our algorithm in the presence of underparameterization and non-convexity. We would like to emphasize that \Cref{prop_1} is \textit{not needed} to show convergence of \texttt{FedExP}; it is only needed to motivate why \texttt{FedExP} might be beneficial. To show general convergence, we only require that $\eta_l$ be sufficiently small and the standard assumptions stated below. 

\paragraph{Challenge in incorporating stochastic noise and partial participation.} Our current analysis focuses on the case where clients are computing full-batch gradients in every step with full participation. This is primarily due to the difficulty in decoupling the effect of stochastic and sampling noise on $\eta_g^{(t)}$ and the pseudo-gradients $\{\Delta_i^{(t)}\}_{i=1}^M$. To be more specific, if we use $\xi^{(t)}$ to denote the randomness at round $t$, then $\Ex{(\eta_g^{(t)}) \bardeltat} \neq \Ex{(\eta_g^{(t)})}\Ex{\bardeltat}$ which significantly complicates the proof. This is purely a theoretical limitation.
Empirically, our results in \Cref{sec:experiments} show that \texttt{FedExP} performs well with both SGD and partial client participation.

\begin{assum}($L$-smoothness)
\label{assum_data_smooth}
Local objective $F_i(\bw)$ is  differentiable and $L$-smooth for all $i \in [M]$, i.e., $\Vert\nabla F_i(\bw) - \nabla F_i(\bw')\Vert \leq L\Vert\bw-\bw'\Vert$, $\forall \bw, \bw' \in \mathbb{R}^d$.
\end{assum}

\begin{assum}(Bounded data heterogenenity at optimum)
\label{assum_optima_dissim}
The norm of the client gradients at the global optima $\bw^*$ is bounded as follows: $\frac{1}{M}\sum_{i=1}^M \norm{\nabla F_i(\bw^*)} \leq \sigma^2_{*}$.
\end{assum}

\begin{thm} \textcolor{black}{($F_i$ are convex) 
Under Assumptions \ref{assum_data_smooth},\ref{assum_optima_dissim} and assuming clients compute full-batch gradients with full participation and $\eta_l \leq \frac{1}{6\tau L}$, the iterates $\{ \bw^{(t)}\}$ generated by \texttt{FedExP} satisfy,}
\begin{align}
     F(\bar{\bw}^{(T)})-F^* \leq  \underbrace{\bigO{\frac{\norm{\bw^{(0)}-\bw^*}}{\eta_l \tau \sum_{t=0}^{T-1}\eta_g^{(t)}}}}_{T_1 := \text{initialization error}} + \underbrace{\bigO{\eta_l^2 \tau(\tau-1) L\sigma^2_{*}}}_{T_2 := \text{client drift error}} + \underbrace{\bigO{\eta_l\tau \sigma^2_{*}}}_{T_3 := \text{noise at optimum}},
\end{align}
where $\eta_g^{(t)}$ is the \texttt{FedExP} server step size at round $t$ and $\bar{\bw}^{(T)}  = \frac{\sum_{t=0}^{T-1}\eta_g^{(t)}\bw^{(t)}}{\sum_{t=0}^{T-1}\eta_g^{(t)}}$.
\label{thm:theorem_1}
\end{thm}

For the non-convex case, we need the data heterogeneity to be bounded everywhere as follows.
\begin{assum}(Bounded global gradient variance) 
\label{assum_uniform_variance}
There exists a constant $\sigma^2_{g} >0$ such that the global gradient variance is bounded as follows.   
$\frac{1}{M}\sum_{i=1}^M \norm{\nabla F_i(\bw)-\nabla F(\bw)} \leq \sigma_g^2$, $\forall \bw \in \mathbb{R}^d$.
\end{assum}

\begin{thm} \textcolor{black}{($F_i$ are non-convex)
Under Assumptions \ref{assum_data_smooth}, \ref{assum_uniform_variance} and assuming clients compute full-batch gradients with full participation and $\eta_ l \leq \frac{1}{6\tau L}$, the iterates $\{\bw^{(t)}\}$ generated by \texttt{FedExP} satisfy,}
\begin{align}
    \min_{t \in [T]} \norm{\nabla F(\bw^{(t)})} \leq \underbrace{\bigO{\frac{F(\bw^{(0)})-F^*}{\eta_l \tau \sum_{t=0}^{T-1}\eta_g^{(t)}}}}_{T_1 := \text{initialization error}} + \underbrace{\bigO{\eta_l^2L^2(\tau-1)\tau \sigma_g^2}}_{T_2 := \text{client drift error}} + \underbrace{\bigO{\eta_l L\tau\sigma_g^2}}_{T_3 := \text{ global variance}},
\end{align}
where $\eta_g^{(t)}$ is the \texttt{FedExP} server step size at round $t$.
\label{thm:theorem_2}
\end{thm}

\paragraph{Discussion.}
 \textcolor{black}{In the convex case, the error of \texttt{FedAvg} can be bounded by $\bigO{\|\bw^{(0)}-\bw^*\|^2/\eta_l \tau T} + \bigO{\eta_l^2 \tau(\tau-1)L\sigma_*^2}$ \citep{khaled2020tighter} and in the non-convex case by $\bigO{(F(\bw^0)-F^*)/\eta_l\tau T} + \bigO{\eta_l^2L^2\tau(\tau-1)\sigma_g^2}$ \citep{wang2020tackling}.}
A careful inspection reveals that the impact of $T_1$ on convergence of \texttt{FedExP} is different from \texttt{FedAvg} (effect of $T_2$ is the same). 
We see that since $\sum_{t=0}^{T-1} \eta_g^{(t)} \geq T$, \texttt{FedExP} reduces $T_1$ faster than \texttt{FedAvg}.
However this comes at the price of an increased error floor due to $T_3$.
Thus, the larger step-sizes in \texttt{FedExP} help us reach the vicinity of an optimum faster, but can ultimately end up saturating at a higher error floor due to noise around the optimum. Note that the impact of the error floor can be controlled by setting the client step size $\eta_l$ appropriately. Moreover, in the overparameterized convex regime where $\sigma^2_{*} = 0$, the effect of $T_2$ and $T_3$ vanishes and thus \texttt{FedExP} clearly outperforms \texttt{FedAvg}. This aligns well with our initial motivation of using extrapolation in the overparameterized regime.

\begin{figure*}[t]
 \centering
    \hspace{-0.5em}
    \subfloat{\includegraphics[width=0.325\linewidth]{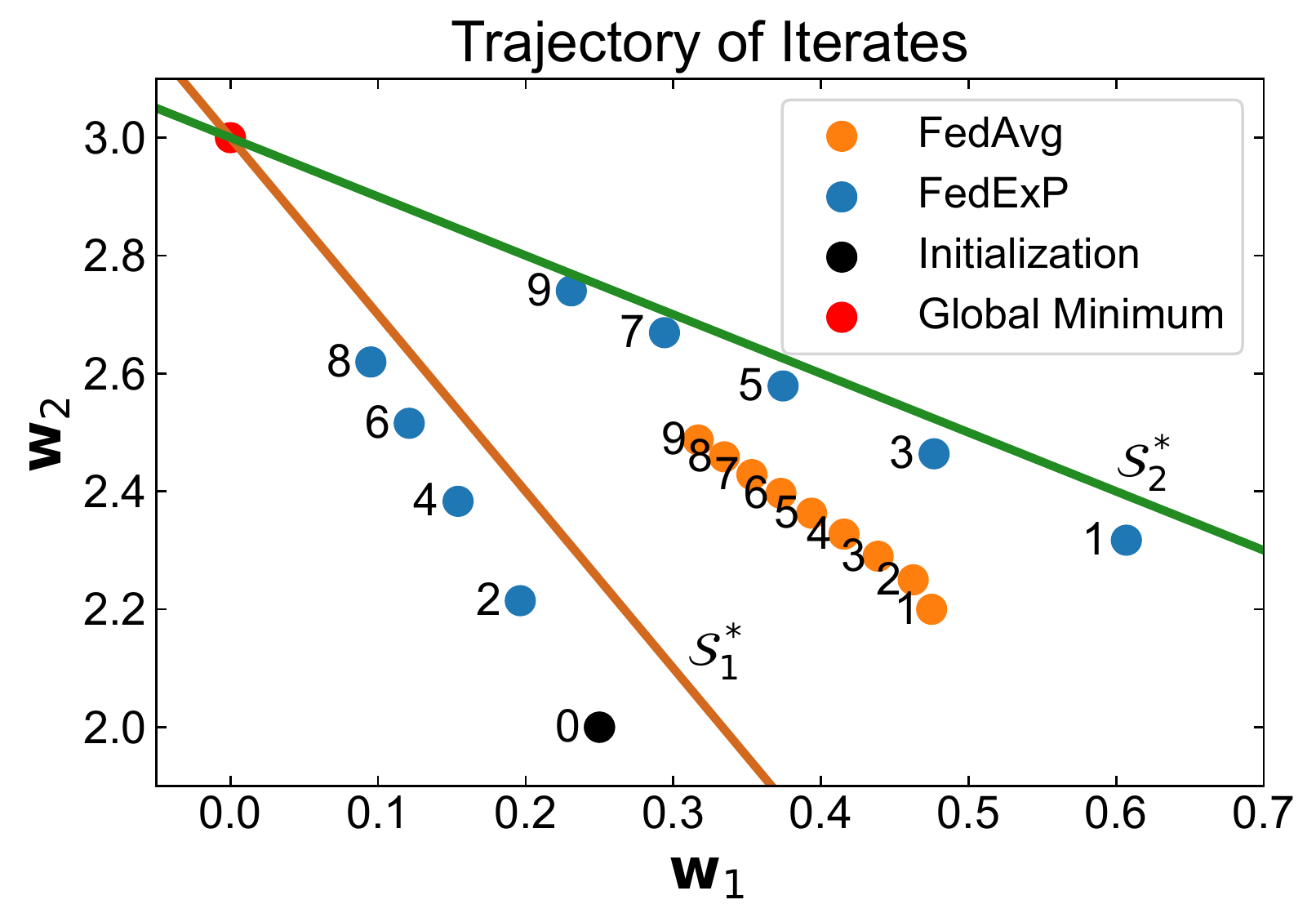}%
    }~
    \subfloat{\includegraphics[width=0.325\linewidth]{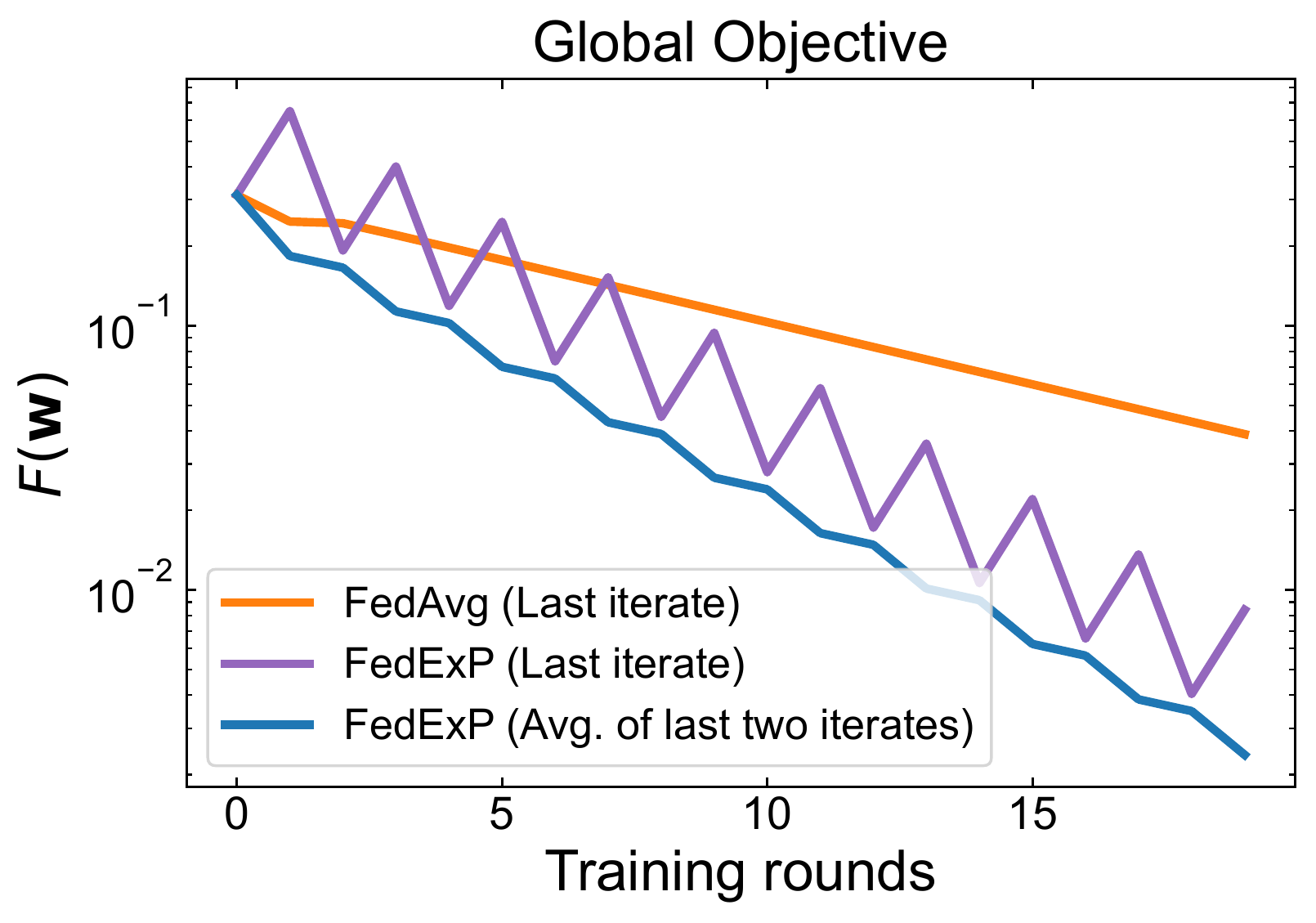}%
    }~
    \subfloat{\includegraphics[width=0.325\linewidth]{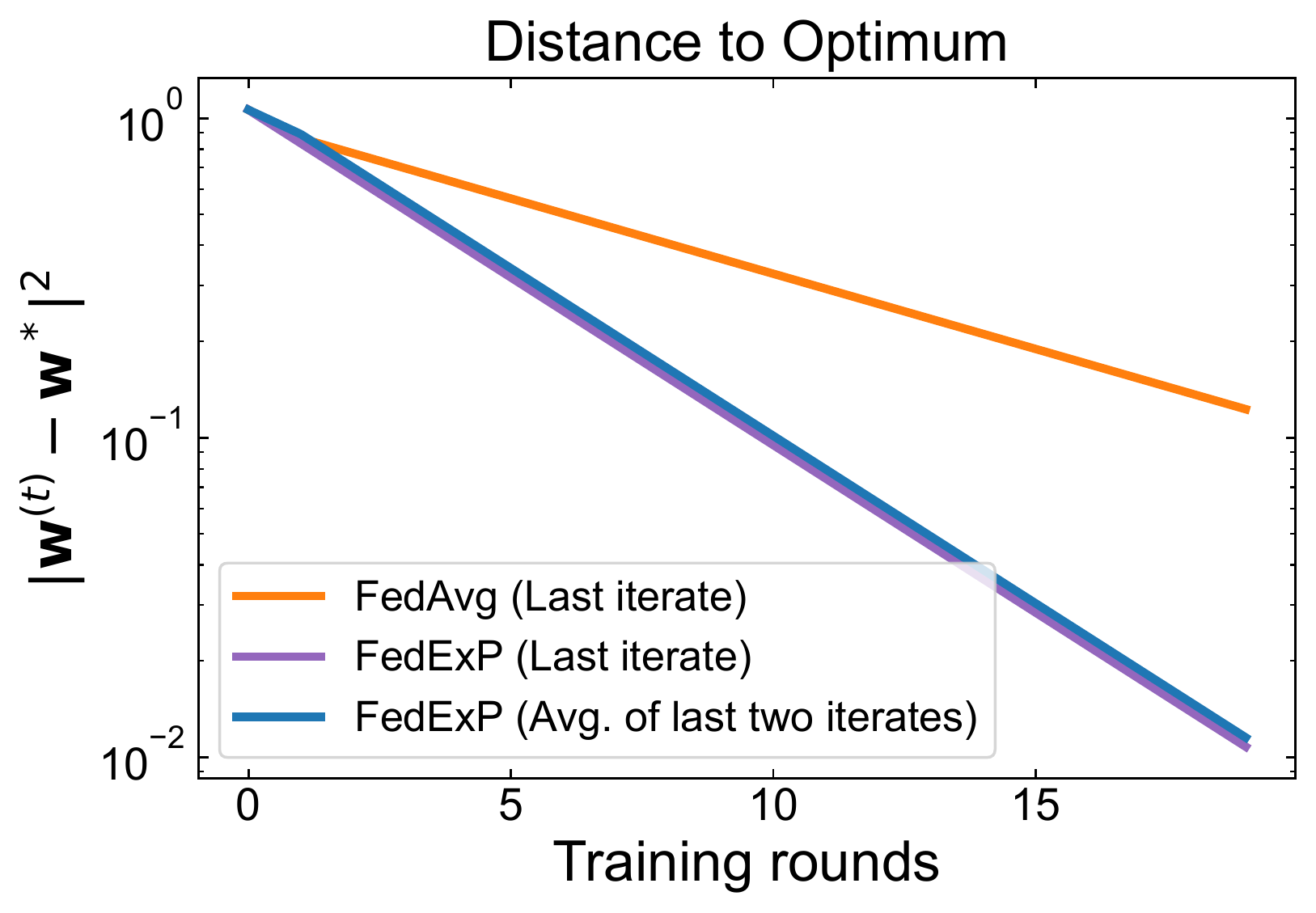}%
    }
    \vspace{-0.7em} 
    \caption{Training characteristics of \texttt{FedAvg} and \texttt{FedExP} for the 2-D toy problem in \Cref{sec:practical_insights}. The last iterate of \texttt{FedExP} has an oscillating behavior in $F(\bw)$ but monotonically decreases $\normsmall{\bw^{(t)}-\bw^*}$; the average of the last two iterates lies in a lower loss region than the last iterate.}
\label{fig:toy_problem}
\end{figure*}

\section{Further Insights into FedExP}
\label{sec:practical_insights}
In this section, we discuss some further insights into the training of \texttt{FedExP} and how we leverage these insights to improve the performance of \texttt{FedExP}.

\paragraph{\texttt{FedExP} monotonically decreases $\norm{\bw^{(t)}-\bw^*}$ but not necessarily $F(\bw^{(t)
}) - F(\bw^*)$.} Recall that our original motivation for the \texttt{FedExP} step size was aimed at trying to minimize the \textit{distance} to the optimum give by
$\norm{\bw^{(t+1)}-\bw^*}$, when \Cref{prop_1} holds. Doing so satisfies $\norm{\bw^{(t+1)}-\bw^*} \leq \norm{\bw^{(t)}-\bw^*}$ but does not necessarily satisfy $F(\bw^{(t+1)}) \leq F(\bw^{(t)})$. 

To better illustrate this phenomenon, we consider the following toy example in $\mathbb{R}^2$. We consider a setup with two clients, where the objective at each client is given as follows:
\begin{align}
    F_1(\bw) = (3w_1 + w_2 - 3)^2; \hspace{5pt} F_2(\bw) = (w_1 + w_2 - 3)^2.
\end{align}
We denote the set of minimizers of $F_1(\bw)$ and $F_2(\bw)$ by $\mathcal{S}_1^* = \{\bw: 3w_1 + w_2 = 3\}$ and $\mathcal{S}_2^* = \{\bw:w_1 + w_2 = 3\}$ respectively. Note that $\mathcal{S}_1^*$ and $\mathcal{S}_2^*$ intersect at the point $\bw^* = [0, 3]$, making it a global minimum. To minimize their local objectives, we assume clients run gradient descent with $\tau \rightarrow \infty$ in every round\footnote{ The local models will be an exact projection of the global model on the solution sets $\{\mathcal{S}_i^*\}_{i=1}^2$. In this case, the lower bound in \cref{eq:eta_g_lb} can be improved by a factor of 2 and therefore we use $\eta_g^{(t)} = (\|\Delta_1\|^2 + \|\Delta_2\|^2)/2\|\bardeltat\|^2$ for this experiment (see \Cref{app:subsec_linear_regression_proj} and \Cref{app:subsec_linear_regression_lb} for proof).}.
\Cref{fig:toy_problem} shows the trajectory of the iterates generated by \texttt{FedExP} and \texttt{FedAvg}. 
We see that while $\norm{\bw^{(t)}-\bw^*}$ decreases monotonically for \texttt{FedExP}, $F(\bw^{(t)})$ does not do so and in fact has an oscillating nature as we discuss below.

\paragraph{Understanding oscillations in $F(\bw^{(t)})$.} We see that the oscillations in $F(\bw^{(t)})$ are caused by \texttt{FedExP} iterates trying to minimize their distance from the solution sets $\mathcal{S}_1^*$ and $\mathcal{S}_2^*$ simultaneously. The initialization point $\bw^{(0)}$ is closer to $\mathcal{S}_1^*$ than $\mathcal{S}_2^*$, which causes the \texttt{FedExP} iterate at round 1 to move towards $\mathcal{S}_2^*$, then back towards $\mathcal{S}_1^*$ and so on. To understand why this happens, consider the case where $\Delta_1^{(t)} = 0, \Delta_2^{(t)} \neq 0$. In this case, we have $\eta_g^{(t)}= 2$ and therefore $\bw^{(t+1)} = \bw^{(t)} - 2\bardeltat = \bw_2^{(t,\tau)}$, which indicates that \texttt{FedExP} is now trying to minimize $\norm{\Delta_2^{(t+1)}}$. This gives us the intuition that the \texttt{FedExP} update in round $t$ is trying to minimize the objectives of the clients that have $\norm{\Delta_i^{(t)}} \gg 0$. While this leads to a temporary increase in global loss $F(\bw^{(t)})$ in some rounds as shown in \Cref{fig:toy_problem}, it is beneficial in the long run as it leads to a faster decrease in distance to the global optimum $\bw^*$.

\paragraph{Averaging last two iterates in \texttt{FedExP}.} Given the oscillating behavior of the iterates of \texttt{FedExP}, we find that measuring progress on $F(\bw)$ using the last iterate can be misleading. Motivated by this finding, we propose to set the final model as the \textit{average} of the last two iterates of \texttt{FedExP}. While the last iterate oscillates between regions that minimize the losses $F_1(\bw)$ and $F_2(\bw)$ respectively, the behavior of the average of the last two iterates is more stable and proceeds along a globally low loss region. 
Interestingly, we find that the benefits of averaging the iterates of \texttt{FedExP} also extend to training neural networks with multiple clients in practical FL scenarios (see \Cref{app:subsec_averaging_in_nn}). In practice, the number of iterates to average over could also be a hyperparameter for \texttt{FedExP}, but we find that averaging the last two iterates works well, and we use this for our other experiments.

\begin{figure}[t]
 \centering
 \includegraphics[width=1\linewidth]{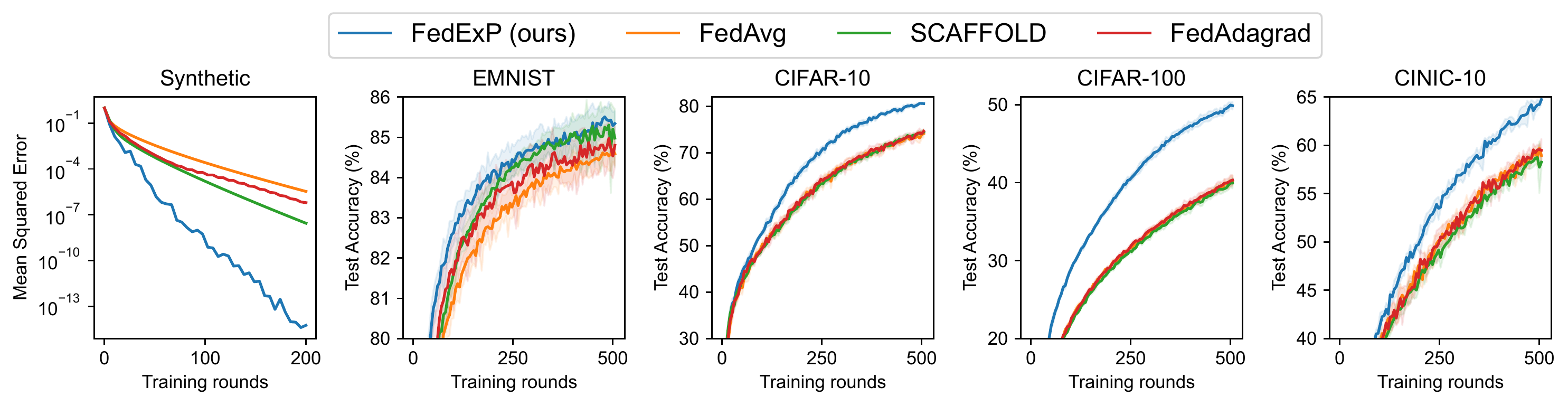}\vspace{-0.7em}    \caption{Experimental results on a synthetic linear regression experiments and a range of realistic FL tasks. \texttt{FedExP} consistently gives faster convergence compared to baselines while adding no extra computation, communication or storage at clients or server.}
\label{fig:exp_results}
\end{figure}

\section{Experiments}
\label{sec:experiments}

We evaluate the performance of \texttt{FedExP} on synthetic and real FL tasks. For our synthetic experiment, we consider a distributed overparameterized linear regression problem. This experiment aligns most closely with our theory and allows us to carefully examine the performance of \texttt{FedExP} when \cref{prop_1} holds. For realistic FL tasks, we consider image classification on the following datasets i)~EMNIST \citep{cohen2017emnist}, ii) CIFAR-10 \citep{krizhevsky2009learning}, iii) CIFAR-100 \citep{krizhevsky2009learning}, iv) CINIC-10 \citep{darlow2018cinic}. In all experiments, we compare against the following baselines i) \texttt{FedAvg}, ii) \texttt{SCAFFOLD} \citep{karimireddy2020scaffold}, and iii)  \texttt{FedAdagrad} \citep{reddi2020adaptive} which is a federated version of the popular \texttt{Adagrad} algorithm. To the best of our knowledge, we are not aware of any other baselines that adaptively tune the server step size in FL.

\paragraph{Experimental Setup. } \hspace{-1em}
For the synthetic experiment, we consider a setup with 20 clients, 30 samples at each client, and model size to be 1000, making this an overparameterized problem. The data at each client is generated following a similar procedure as the synthetic dataset in \cite{li2020federated}. We use the federated version of EMNIST available at \cite{caldas2018leaf}, which is naturally partitioned into 3400 clients. For CIFAR-10/100 we artifically partition the data into 100 clients, and for CINIC-10 we partition the data into 200 clients. In both cases, we follow a Dirichlet distribution with $\alpha = 0.3$ for the partitioning to model heterogeneity among client data \citep{hsu2019measuring}.
For EMNIST we use the same CNN architecture used in \citet{reddi2020adaptive}.
For CIFAR10, CIFAR100 and CINIC-10 we use a ResNet-18 model \citep{he2016deep}. For our baselines, we find the best performing $\eta_g$ and $\eta_l$ by grid-search tuning. For \texttt{FedExP} we optimize for $\epsilon$ and $\eta_l$ by grid search. We fix the number of participating clients to 20, minibatch size to 50 and number of local updates to 20 for all experiments. 
\textcolor{black}{In \Cref{app:sec_experiments}, we provide additional details and results, including the best performing hyperparameters, comparison with \texttt{FedProx} \citep{li2020federated}, and results for more rounds. }

\paragraph{\texttt{FedExP} comprehensively outperforms \texttt{FedAvg} and baselines.} Our experimental results in \Cref{fig:exp_results} demonstrate that \texttt{FedExP} clearly outperforms \texttt{FedAvg} and competing baselines that use the best performing $\eta_g$ and $\eta_l$ found by grid search. Moreover, \texttt{FedExP} does not require additional communication or storage at clients or server unlike \texttt{SCAFFOLD} and \texttt{FedAdagrad}. The order-wise improvement in the case of the convex linear regression experiment confirms our theoretical motivation for \texttt{FedExP} outlined in \Cref{subsection:extrapolation in FL}. In this case, since \cref{prop_1} is satisfied, we know that the \texttt{FedExP} iterates are always moving towards the optimum. For realistic FL tasks, we see a consistent speedup of over $1.4-2 \times$ over \texttt{FedAvg}. This verifies that \texttt{FedExP} also provides performance improvement in more general settings with realistic datasets and models.
Plots showing $\eta_g^{(t)}$ can be found in \Cref{app:subsec_addtl_results}. The key takeaway from our experiments is that adapting the server step size allows \texttt{FedExP} to take much larger steps in some (but not all) rounds compared to the constant optimum step size taken by our baselines, leading to a large speedup. 

\textbf{Comparison with \texttt{FedAdagrad}.} As discussed in \Cref{sec:introduction}, \texttt{FedAdagrad} and \texttt{FedExP} use different notions of adaptivity; \texttt{FedAdagrad} uses coordinate-wise adaptivity, while \texttt{FedExP} uses client-based adaptivity. We believe that the latter is more meaningful for FL settings as seen in our experiments. In many experiments, especially image classification tasks like CIFAR, the gradients produced are \textit{dense} with relatively little variance in coordinate-wise gradient magnitudes \citep{reddi2020adaptive,zhang2020adaptive}. In such cases, \texttt{FedAdagrad} is unable to leverage any coordinate-level information and gives almost the same performance as \texttt{FedAvg}.

\textbf{Comparison with \texttt{SCAFFOLD}. } We see that \texttt{FedExP} outperforms \texttt{SCAFFOLD} in all experiments, showing that adaptively tuning the server step size is sufficient to achieve speedup in FL settings. 
Furthermore, \texttt{SCAFFOLD} even fails to outperform \texttt{FedAvg} for the more difficult CIFAR and CINIC datasets. 
Several other papers have reported similar findings, including \citet{reddi2020adaptive, karimireddy2020mime, yu2022tct}. Several reasons have been postulated for this behavior, including the staleness of control variates \citep{reddi2020adaptive} and the difficulty in characterizing client drift in non-convex scenarios \citep{yu2022tct}. Thus, while theoretically attractive, simply using variance reduction techniques such as \texttt{SCAFFOLD} may not provide any speedup in practice.
\paragraph{Adding extrapolation to \texttt{SCAFFOLD}.} 
\begin{wrapfigure}{r}{0.28\textwidth} 
\vspace{-1em}
\centering 
\vspace{-1em}
\!\!\includegraphics[width=0.29\textwidth]{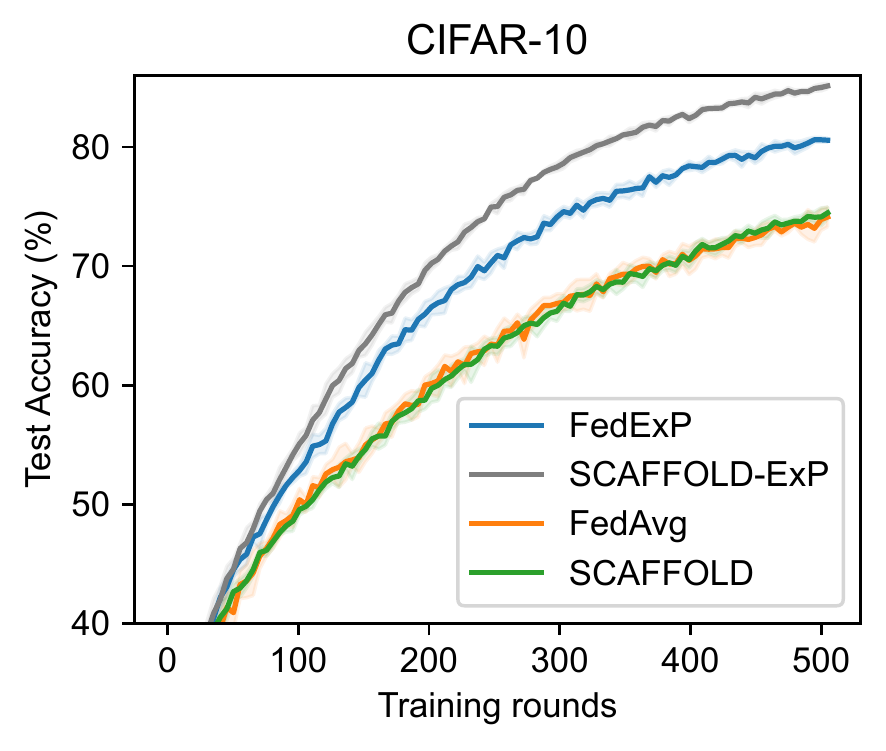}\vspace{-0.7em}\caption{Adding extrapolation to \texttt{SCAFFOLD} for greater speedup.}\vspace{-1em}
    \label{fig:scaffold+exp}
\end{wrapfigure}
We note that \texttt{SCAFFOLD} only modifies the Local SGD procedure at clients and keeps the global aggregation at the server unchanged. Therefore, it is easy to modify the \texttt{SCAFFOLD} algorithm to use extrapolation when updating the global model at the server (algorithm details in \Cref{app:sec_scaffold_exp}). \Cref{fig:scaffold+exp} shows the result of our proposed extrapolated \texttt{SCAFFOLD} on the CIFAR-10 dataset. Interestingly, we observe that while \texttt{SCAFFOLD} alone fails to outperform \texttt{FedAvg}, the extrapolated version of \texttt{SCAFFOLD} achieves the best performance among all algorithms. This result highlights the importance of carefully tuning the server step size to achieve the best performance for variance-reduction algorithms. \textcolor{black}{It is also possible to add extrapolation to algorithms with server momentum (\Cref{app:sec_server_momentum}).}
\vspace{-0.5em}

\begin{table}[t]

\centering
\caption{Table showing the average number of rounds to reach desired accuracy for \texttt{FedExP} and baselines. \texttt{FedExP} provides a consistent speedup over all baselines.}\vspace{-0.5em}
{\small
\begin{tabular}{c c c c c c}
\thickhline \\[-0.32cm]
Dataset & Target Acc. & \texttt{FedExP} & \texttt{FedAvg} & \texttt{SCAFFOLD} & \texttt{FedAdagrad}\\ [0.01cm]
\hline\\[-0.3cm]
EMNIST & $84\%$ & $186$ & $328\, (1.76 \times)$ & $232\, (1.24 \times)$ & $277\, (1.48 \times)$\\
CIFAR-10 & $72\%$ & $267$ & $434\, (1.62 \times)$ & $429\, (1.61 \times)$ & $419\, (1.56 \times)$\\
CIFAR-100 & $40\%$ & $242$ & $500\, (2.06 \times)$ & $>\!500\, (>\!2.06 \times)$ & $494\, (2.04 \times)$\\
CINIC-10 & $58\%$ & $318$ & $450\, (1.42 \times)$ & $470\, (1.48 \times)$ & $444\, (1.40 \times)$\\
\thickhline\\[-0.3cm]
\end{tabular}
}
\end{table}

\section{Conclusion \texorpdfstring{\vspace{-0.4em}}{}}

In this paper, we have proposed \texttt{FedExP}, a novel extension of \texttt{FedAvg} that adaptively determines the server step size used in every round of global aggregation in FL. Our algorithm is based on the key observation that \texttt{FedAvg} can be seen as an approximate variant of the \texttt{POCS} algorithm, especially for overparameterized convex objectives. This has inspired us to leverage the idea of extrapolation that is used to speed up \texttt{POCS} in a federated setting, resulting in \texttt{FedExP}. We have also discussed several theoretical and empirical perspectives of \texttt{FedExP}. In particular, we have explained some design choices in \texttt{FedExP} and how it can be used in practical scenarios with partial client participation and secure aggregation. We have also shown the convergence of \texttt{FedExP} for possibly underparameterized models and non-convex objectives.  Our experimental results have shown that \texttt{FedExP} consistently outperforms baseline algorithms with virtually no additional computation or communication at clients or server. We have also shown that the idea of extrapolation can be combined with other techniques, such as the variance-reduction method in \texttt{SCAFFOLD}, for greater speedup. Future work will study the convergence analysis of \texttt{FedExP} with stochastic gradient noise and the incorporation of extrapolation into a wider range of algorithms used in FL.

\section*{Acknowledgments}
This work was supported in part by NSF grants CCF 2045694, CNS-2112471, ONR N00014-23-1-2149, and the CMU David Barakat and LaVerne Owen-Barakat Fellowship.


\bibliography{Bibliography}
\bibliographystyle{iclr2023_conference}

\clearpage
\begin{center}
\LARGE \textsc{Appendix}
\end{center}

\appendix

\startcontents[sections]
\printcontents[sections]{l}{1}{\setcounter{tocdepth}{2}}

\clearpage

\input{appendix}

\label{appendix}

\end{document}

%% file: math_commands.tex
\usepackage{amsmath,amsfonts,bm}

\def\eqref#1{equation~\ref{#1}}

\def\1{\bm{1}}

\newcommand{\Ex}[1]{\mathbb{E}_{\xi^{(t)}}\left[{#1}\right]}

\DeclareMathOperator*{\argmin}{arg\,min}

\newcommand{\norm}[1]{\left\lVert#1\right\rVert^{2}}
\newcommand{\normsmall}[1]{\big\lVert#1\big\rVert^{2}}

\newcommand{\bigO}[1]{\mathcal{O}\left({#1}\right)}

\newcommand{\bardeltat}{\bar{\Delta}^{(t)}}

\def\ba{{\mathbf a}}

\newcommand{\bA}{{\bf A}}
\newcommand{\bv}{{\bf v}}
\newcommand{\bmm}{{\bf m}}
\newcommand{\bI}{{\bf I}}
\newcommand{\bSigma}{{\bf \Sigma}}
\newcommand{\bU}{{\bf U}}
\newcommand{\bV}{{\bf V}}
\newcommand{\bx}{{\bf x}}

\newcommand{\bb}{{\bf b}}

\newcommand{\bh}{{\bf h}}

\newcommand{\bw}{{\bf w}}

\newcommand{\bc}{{\bf c}}

\DeclareMathAlphabet{\mathsfit}{\encodingdefault}{\sfdefault}{m}{sl}
\SetMathAlphabet{\mathsfit}{bold}{\encodingdefault}{\sfdefault}{bx}{n}

%% file: appendix.tex
\section{Additional Related Work}
\label{app_sec:related_work}

In this section, we provide further discussion on some additional related work that complements our discussion in \Cref{sec:introduction}.

\paragraph{Adaptive Step Size in Gradient Descent.} Here we briefly discuss methods for tuning the step size in gradient descent and the challenges in applying them to the FL setting. Early methods to tune the step size in gradient descent were based on line search (or backtracking) strategies \citep{armijo1966minimization,goldstein1977optimization}. However, these strategies need to repeatedly compute the function value or gradient within an iteration, making them computationally expensive. Another popular class of adaptive step sizes is based on the Polyak step size \citep{POLYAK196914, hazan2019revisiting, loizou2021stochastic}. Similar to \texttt{FedExP}, the Polyak step size is derived from trying to minimize the distance to the optimum for convex functions. However it is not clear how this can be extended to the federated setting where we only have access to pseudo-gradients. Also, the Polyak step size requires knowledge of the function value at the optimum which is hard to estimate. Another related class of step sizes is the Barzilai-Borwein stepsize \citep{barzilai1988two}. However, to the best of our knowledge, these are known to provably work only for quadratic functions \citep{raydan1993barzilai,burdakov2019stabilized} only. A recent work \citep{malitsky2019adaptive} alleviates some of the concerns associated with these classical methods by setting the step size as an approximation of the inverse local Lipschitz constant; however it is again not clear how this intuition can be applied to the federated setting. An orthogonal line of work has focused on methods that adapt to the geometry of the data using gradient information in previous iterations, the most popular among them being  \texttt{Adagrad} \citep{duchi2011adaptive} and its extensions \texttt{RMSProp} \citep{tieleman2012lecture} and \texttt{Adadelta} \citep{zeiler2012adadelta}. There exist federated counterparts of these algorithms, namely \texttt{FedAdagrad}; however, as we show in our experiments these methods can fail to even outperform \texttt{FedAvg} in standard FL tasks.

\paragraph{Overparameterization in FL.} Inspired by the success of analyzing deep neural networks in the neural tangent kernel (NTK) regime \citep{jacot2018neural, arora2019exact, allen2019learning}, recent work has looked at studying the convergence of overparameterized neural networks in the FL setting. \cite{huang2021fl} and \cite{deng2022local} show that for a sufficiently wide neural network and proper step size conditions, \texttt{FedAvg} will converge to a globally optimal solution even in the presence of data heterogeneity. We note that these works are primarily concerned with convergence analysis, whereas our focus is on developing a practical algorithm that is inspired by characteristics in the overparameterized regime for speeding up FL training. Another recent line of work has looked at utilizing NTK style Jacobian features for learning a FL model in just a few rounds of communication \citep{yu2022tct, yue2022neural}. While interesting, these approaches are orthogonal to our current work.

\clearpage

\color{black}

\section{Table of Notation and Schematic}
\label{app_sec:table_and_schematic}

\subsection{Table of Notation}

\captionsetup[table]{labelfont={color=black},font={color=black}}

\begin{table}[h]
\centering
\caption{Summary of notation used in paper} \vspace{-0.5em}
\color{black}
\begin{tabular}{c l}
\thickhline \\[-0.2cm]
Symbol & Description\\ [0.05cm]
\hline\\[-0.2cm]
$\| \; \|$ & $L_2$ norm\\
$M$ & Number of clients\\
$\ell(\cdot, \cdot)$ & Loss function\\
$\mathcal{D}_i$ & Dataset at $i$-th client\\
$F_i(\bw)$ & Local objective at $i$-th client\\
$F(\bw)$ & Global objective at server\\
$\eta_l$ & Client step size\\
$\eta_g$ & Server step size\\
$\bw^{(t)}$ & Global model at round $t$\\
$\eta_g^{(t)}$ & \texttt{FedExP} server step size at round $t$\\
$\bw_i^{(t,k)}$& Local model at $i$-th client at $t$-th round and $k$-th iteration\\
$\tau$ & Number of local SGD steps\\
$\Delta_i^{(t)}$ & Update of $i$-th client at round $t$\\
$\bardeltat$ & Average of client updates at round $t$\\
$\mathcal{S}_i^*$ & Set of minimizers of $F_i(\bw)$\\
$T$ & Number of communication rounds\\
$\epsilon$ & Small constant added to denominator of \texttt{FedExP} step size\\
$\bw^*$ & Global minimum\\
$F^*$ & Minimum value of global objective\\
$L$ & $L$-smoothness constant used in \Cref{assum_data_smooth}\\
$\sigma_*^2$ & Upper bound on variance of client gradients at optimum (see \Cref{assum_optima_dissim}) \\
$\sigma^2$ & Upper bound on variance of client gradients (see \Cref{assum_uniform_variance})\\ 
\thickhline\\
[-0.2cm]
\end{tabular}
\end{table}

\subsection{Schematic of Client-Server communication in FedExP}
\begin{figure}[h]
\color{black}
 \centering
 \subfloat{
 \includegraphics[width=0.70\linewidth]{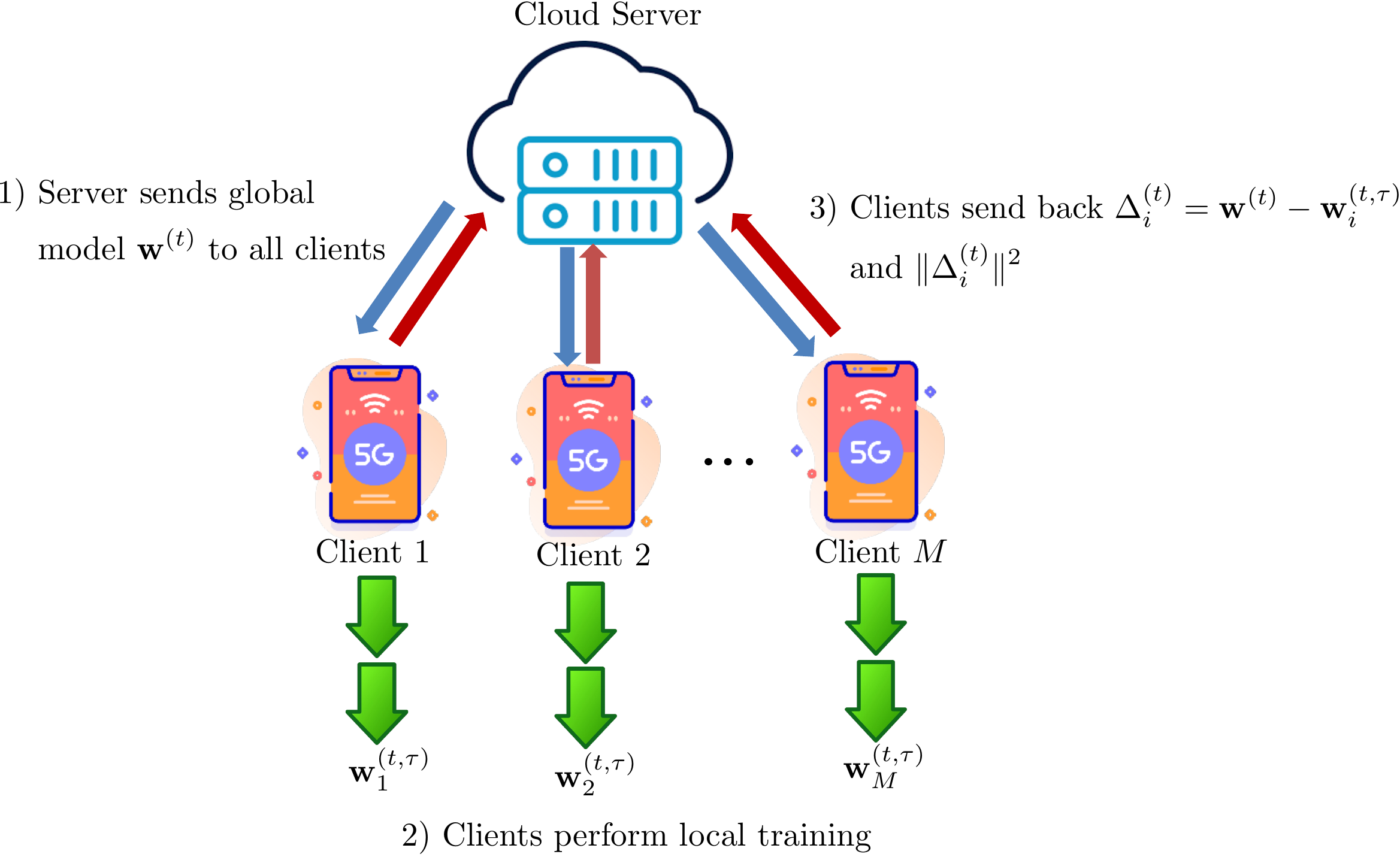}
}~
\caption{\textcolor{black}{Schematic of client-server communication in \texttt{FedExP}. }}
\label{fig:fedexpSchematic}
\end{figure}

At each round $t$, the server first sends global model $\bw^{(t)}$ to all clients. Clients perform local optimization on $\bw^{(t)}$ to compute their local models $\bw_i^{(t,\tau)}$ and send back their update $\Delta_i^{(t)} = \bw_i^{(t)} - \bw_i^{(t,\tau)}$ and norm of update $\norm{\Delta_i^{(t)}}$ to the server. This procedure is illustrated in Figure~\ref{fig:fedexpSchematic}.

\captionsetup[figure]{labelfont={color=black},font={color=black}}

\captionsetup[figure]{labelfont={color=black},font={color=black}}

\captionsetup[table]{labelfont={color=black},font={color=black}}

\newpage

\color{black}

\section{Proofs}
\label{appendix:sec_proofs}
We first state some preliminary lemmas that will used throughout the proofs.

\begin{lem}(Jensen's inequality)
For any $\ba_i \in \mathbb{R}^d, i\in\{1,2,\ldots,M\}$:
\begin{align}
    \norm{ \frac{1}{M}\sum_{i=1}^M \ba_i } &\leq \frac{1}{M}\sum_{i=1}^M \norm{\ba_i },
    \label{eq:Jensen1}\\
    \norm{ \sum_{i=1}^M \ba_i } &\leq M \sum_{i=1}^M \norm{ \ba_i }.
    \label{eq:Jensen2}
\end{align}
\end{lem}

We also note the following known result related to the Bregman divergence.
\begin{lem}\citep{khaled2020tighter}
\label{lemma: bregman_div}
If $F$ is smooth and convex, then
\begin{align}
    \norm{\nabla F(\bw) - \nabla F(\bw')} \leq 2L(F(\bw)-F(\bw') - \langle \nabla F(\bw'),\bw-\bw' \rangle).
\end{align}
\end{lem}

\begin{lem}(Co-coercivity of convex smooth function)
If $F$ is $L$-smooth and convex then,
\begin{align}
\label{lemma:co-coercivity}
\left\langle \nabla F(\bw) - \nabla F(\bw'),\bw-\bw' \right\rangle \geq \frac{1}{L}\norm{\nabla F(\bw) - \nabla F(\bw')}.
\end{align}
\end{lem}
A direct consequence of this lemma is,
\begin{align}
\label{lemma:co-coercivity_consequence}
\left\langle \nabla F(\bw),\bw-\bw^* \right\rangle \geq \frac{1}{L}\norm{\nabla F(\bw)}
\end{align}
where $\bw^*$ is a minimizer of $F(\bw)$.

\subsection{Proof of Lemma~\ref{lem_1}}
\label{app:subsec_lemma1}

Let $F_i(\bw)$ be the local objective at a client and $\bw^*$ be the global minimum. From the overparameterization assumption, we know that $\bw^*$ is also a minimizer for $F_i(\bw)$. We have,
\begin{align}
    \norm{\bw_i^{(t,k)} - \bw^*} &= \norm{\bw_i^{(t,k-1)} - \eta_l \nabla F (\bw_i^{(t,k-1)}) - \bw^*}\\
    & = \norm{\bw_i^{(t,k-1)} \!-\! \bw^*} \!\!- 2\eta_l \langle \nabla F (\bw_i^{(t,k-1)}),\bw_i^{(t,k-1)} \!-\! \bw^*  \rangle + \eta_l^2 \norm{\nabla F (\bw_i^{(t,k-1)})}\\
    & \leq \norm{\bw_i^{(t,k-1)} - \bw^*} - \frac{2\eta_l}{L} \norm{\nabla F (\bw_i^{(t,k-1)})} + \eta_l^2 \norm{\nabla F (\bw_i^{(t,k-1)})} \label{lemma1:1}\\
    & \leq \norm{\bw_i^{(t,k-1)} - \bw^*} - \frac{\eta_l}{L}\norm{\nabla F (\bw_i^{(t,k-1)})} \label{lemma1:2}
\end{align}
where \cref{lemma1:1} follows from \cref{lemma:co-coercivity_consequence} and \cref{lemma1:2} follows from $\eta_l \leq \frac{1}{L}$. Summing the above inequality from $k=0$ to $\tau-1$ we have,
\begin{align}
     \norm{\bw_i^{(t,\tau)} - \bw^*} \leq \norm{\bw^{(t)}-\bw^*} - \frac{\eta_l}{L}\sum_{k=0}^{\tau-1}\norm{\nabla F (\bw_i^{(t,k)})}.
\end{align}
Thus we have,
\begin{align}
    \frac{1}{M}\sum_{i=1}^M  \norm{\bw_i^{(t,\tau)} - \bw^*} &\leq \norm{\bw^{(t)}-\bw^*} - \frac{\eta_l}{ML}\sum_{i=1}^M\sum_{k=0}^{\tau-1}\norm{\nabla F (\bw_i^{(t,k)})}\\
    & \leq \norm{\bw^{(t)}-\bw^*}.
\end{align}
This completes the proof of this lemma. \qed

\subsection{Convergence Analysis for Convex Objectives}
\label{app:subsec_convex}

\textcolor{black}{
Our proof technique is inspired by \cite{khaled2020tighter} with some key differences.
 The biggest difference is the incorporation of the adaptive \texttt{FedExP} server step sizes which \citet{khaled2020tighter} does not account for. Another difference is that we provide convergence guarantees in terms of number of rounds $T$ while \citet{khaled2020tighter} focus on number of iterations $T'=T\tau$. We highlight the specific steps where we made adjustments to the analysis of \citet{khaled2020tighter} below.}

\textcolor{black}{We begin by modifying \citet[Lemma 11 and Lemma 13]{khaled2020tighter} to bound client drift in every round instead of every iteration.}

\begin{lem} (Bounding client aggregate gradients)
\label{lemma:bounding client aggregate gradients}
\begin{align}
    \frac{1}{M}\sum_{i=1}^M \sum_{k=0}^{\tau-1}\norm{\nabla F_i(\bw_i^{(t,k)})} \leq \frac{3L^2}{M}\sum_{i=1}^M \sum_{k=0}^{\tau-1}\norm{\bw_i^{(t,k)}\!-\!\bw^{(t)}} + 6\tau L(F(\bw^{(t)}) - F(\bw^*)) + 3\tau \sigma_*^2 \, .
\end{align}
\end{lem}

\textbf{Proof of \Cref{lemma:bounding client aggregate gradients}:}
\begin{align}
      &\frac{1}{M}\sum_{i=1}^M \sum_{k=0}^{\tau-1}\norm{\nabla F_i(\bw_i^{(t,k)})} \nonumber \\
      & = \frac{1}{M}\sum_{i=1}^M \sum_{k=0}^{\tau-1} \norm{\nabla F_i(\bw_i^{(t,k)}) - \nabla F_i(\bw^{(t)}) + \nabla F_i(\bw^{(t)}) - \nabla F_i(\bw^{*}) + \nabla F_i(\bw^{*})}\\
      & \leq \frac{3}{M}\sum_{i=1}^M \sum_{k=0}^{\tau-1}\norm{\nabla F_i(\bw_i^{(t,k)}) - \nabla F_i(\bw^{(t)})} + \frac{3}{M}\sum_{i=1}^M \sum_{k=0}^{\tau-1}\norm{\nabla F_i(\bw^{(t)}) - \nabla F_i(\bw^{*})} \\
      & \hspace{5pt} + \frac{3}{M}\sum_{i=1}^M \sum_{k=0}^{\tau-1} \norm{\nabla F_i(\bw^{*})}\nonumber\\
     & \leq \frac{3L^2}{M}\sum_{i=1}^M \sum_{k=0}^{\tau-1}\norm{\bw_i^{(t,k)} - \bw^{(t)}} + 6\tau L(F(\bw^{(t)}) - F^*) + 3\tau \sigma^2_{*} \, . \label{lemma5:1}
\end{align}
The first term in \cref{lemma5:1} follows from $L$-smoothness of $F_i(\bw)$, the second term follows from \Cref{lemma: bregman_div} and the third term follows from bounded noise at optimum. \qed

\begin{lem} (Bounding client drift)
\label{lemma:bounding client drift}
\begin{align}
   \frac{1}{M}\sum_{i=1}^M \sum_{k=0}^{\tau-1}\norm{\bw^{(t)}-\bw_i^{(t,k)}} \leq 12\eta_l^2 \tau^2 (\tau-1)L (F(\bw^{(t)})-F(\bw^*)) + 6\eta_l^2 \tau^2 (\tau-1)\sigma_*^2  \, .
\end{align}
\end{lem}

\textbf{Proof of \Cref{lemma:bounding client drift}:}
\begin{align}
 & \frac{1}{M}\sum_{i=1}^M\sum_{k=0}^{\tau-1} \norm{\bw^{(t)}-\bw_i^{(t,k)}} \nonumber \\
 & = \eta_l^2 \frac{1}{M}\sum_{i=1}^M \sum_{k=0}^{\tau-1} \norm{\sum_{l=0}^{k-1} \nabla F_i(\bw_i^{(t,l)})}\\
 & \leq \eta_l^2 \frac{1}{M}\sum_{i=1}^M \sum_{k=0}^{\tau-1} k \sum_{l=0}^{k-1} \norm{ \nabla F_i(\bw_i^{(t,l)})}\\
 & \leq \eta_l^2 \tau (\tau-1) \frac{1}{M}\sum_{i=1}^M \sum_{k=0}^{\tau-1} \norm{\nabla F_i(\bw_i^{(t,k)})}\\
  & \leq 3\eta_l^2 \tau (\tau-1)L^2 \frac{1}{M}\sum_{i=1}^M \sum_{k=0}^{\tau-1} \norm{\bw^{(t)}-\bw_i^{(t,k)}}  + 6\eta_l^2 \tau^2 (\tau-1) L(F(\bw^{(t)})-F(\bw^*)) \label{lemma6:1} \\
  & \hspace{5pt}+ 3\eta_l^2 \tau^2 (\tau-1) \sigma_{*}^2  \nonumber \\
  & \leq \frac{1}{2M}\sum_{i=1}^M \sum_{k=0}^{\tau-1} \norm{\bw^{(t)}-\bw_i^{(t,k)}}  + 6\eta_l^2 \tau^2 (\tau-1) L(F(\bw^{(t)})-F(\bw^*))  \label{lemma6:2}\\
  & \hspace{5pt}+ 3\eta_l^2 \tau^2 (\tau-1) \sigma_{*}^2 \nonumber
\end{align}
where \cref{lemma6:1} uses \Cref{lemma:bounding client aggregate gradients} and \cref{lemma6:2} uses $\eta_l \leq \frac{1}{6\tau L}$.

Therefore we have,
\begin{align}
   \frac{1}{M}\sum_{i=1}^M \sum_{k=0}^{\tau-1}\norm{\bw^{(t)}-\bw_i^{(t,k)}} \leq 12\eta_l^2 \tau^2 (\tau-1)L (F(\bw^{(t)})-F(\bw^*)) + 6\eta_l^2 \tau^2 (\tau-1)\sigma_*^2  \, .
\end{align}
\qed

\textbf{Proof of Theorem 1:}

We define the following auxiliary variables that will used in the proof.

\begin{align}
    \text{Aggregate Client Gradient:} \hspace{10pt} \bh_i^{(t)} = \sum_{k=0}^{\tau-1}\nabla F_i(\bw_i^{(t,k)}).
\end{align}

We also define
$\bar{\bh}^{(t)} = \frac{1}{M}\sum_{i=1}^M \bh_i^{(t)}$.

Recall that the update of the global model can be written as $\bw^{(t+1)} = \bw^{(t)} - \eta_g^{(t)}\eta_l \bar{\bh}^{(t)}$.

We have
\begin{align}
\norm{\bw^{(t+1)}-\bw^*} &= \norm{\bw^{(t)}-\eta_g^{(t)}\eta_l \bar{\bh}^{(t)} - \bw^*}\\
& = \norm{\bw^{(t)}-\bw^*} - 2\eta_g^{(t)} \eta_l \left \langle \bw^t-\bw^*, \bar{\bh}^{(t)} \right \rangle + (\eta_g^{(t)})^2\eta_l^2 \norm{\bar{\bh}^{(t)}}\\
& \leq \norm{\bw^{(t)}-\bw^*} - 2\eta_g^{(t)} \eta_l \left \langle \bw^t-\bw^*, \bar{\bh}^{(t)} \right \rangle + \eta_g^{(t)}\eta_l^2\frac{1}{M}\sum_{i=1}^M \norm{\bh_i^{(t)}} \label{thm1-1}
\end{align}
where \cref{thm1-1} follows from $\eta_g^{(t)} \leq \frac{\sum_{i=1}^M \norm{ \bh_i^{(t)}}}{M\norm{\bar{\bh}^{(t)}}}$.
\textcolor{black}{Inequality \cref{thm1-1} is a key step in our proof and the differentiating factor in our approach from \cite{khaled2020tighter}. Following a similar technique as \cite{khaled2020tighter} to bound $(\eta_g^{(t)})^2\eta_l^2 \norm{\bar{\bh}^{(t)}}$ will end up requiring the condition $\eta_l \leq 1/8L\eta_g^{(t)}$, which cannot be satisfied in our setup due to the adaptive choice of $\eta_g^{(t)}$. Therefore we first upper bound $(\eta_g^{(t)})^2\eta_l^2 \norm{\bar{\bh}^{(t)}}$ by $\eta_g^{(t)}\eta_l^2\frac{1}{M}\sum_{i=1}^M \norm{\bh_i^{(t)}}$ and focus on further bounding this quantity in the rest of the proof, which does not require the aforementioned condition. Note that this comes at the expense of the additional $T_3$ error seen in our final  convergence bound in \Cref{thm:theorem_1}.}

Therefore,

\begin{align}
    \norm{\bw^{(t+1)}-\bw^*} \leq \norm{\bw^{(t)}-\bw^*} - 2\eta_g^{(t)} \eta_l \underbrace{\left \langle \bw^t-\bw^*, \bar{\bh}^{(t)} \right \rangle}_{T_1} + \eta_g^{(t)}\eta_l^2 \underbrace{\frac{1}{M}\sum_{i=1}^M \norm{\bh_i^{(t)}}}_{T_2} .\label{convex_start_ineq}
\end{align}

\textbf{Bounding $T_2$}

We have,
\begin{align}
    T_2 &= \frac{1}{M}\sum_{i=1}^M \norm{\bh_i^{(t)}}\\
    & = \frac{1}{M}\sum_{i=1}^M \norm{\sum_{k=0}^{\tau-1} \nabla F_i(\bw_i^{(t,k)})}\\
    & \leq \frac{\tau}{M}\sum_{i=1}^M \sum_{k=0}^{\tau-1} \norm{\nabla F_i(\bw_i^{(t,k)})} \label{thm1-2}\\
     & \leq \frac{3\tau L^2}{M}\sum_{i=1}^M \sum_{k=0}^{\tau-1}\norm{\bw_i^{(t,k)} - \bw^{(t)}} + 6\tau^2 L(F(\bw^{(t)}) - F^*) + 3\tau^2 \sigma^2_{*} \label{thm1-4}
\end{align}
where \cref{thm1-2} follows from Jensen's inequality and and \cref{thm1-4} follows from \Cref{lemma:bounding client aggregate gradients}.

\textbf{Bounding $T_1$}
\begin{align}
    T_1 &= \frac{1}{M}\sum_{i=1}^M \left \langle \bw^t-\bw^*, \bh_i^{(t)} \right \rangle\\
    & = \frac{1}{M}\sum_{i=1}^M \sum_{k=0}^{\tau-1} \left \langle \bw^{(t)}-\bw^*, \nabla F_i(\bw_i^{(t,k)}) \right\rangle . \label{t1_convex_simplify}
\end{align}
We have,
\begin{align}
   \left \langle \bw^{(t)}-\bw^*, \nabla F_i(\bw_i^{(t,k)}) \right\rangle  = \left \langle \bw^{(t)}-\bw_i^{(t,k)}, \nabla F_i(\bw_i^{(t,k)}) \right\rangle + \left \langle \bw_i^{(t,k)} - \bw^*, \nabla F_i(\bw_i^{(t,k)}) \right\rangle .
\end{align}

From $L$-smoothness of $F_i$ we have,
\begin{align}
    \left \langle \bw^{(t)}-\bw_i^{(t,k)}, \nabla F_i(\bw_i^{(t,k)}) \right\rangle \geq F_i(\bw^{(t)}) - F_i(\bw_i^{(t,k)}) -  \frac{L}{2}\norm{\bw^{(t)}-\bw_i^{(t,k)}} .
\end{align}

From convexity of $F_i$ we have,
\begin{align}
  \left \langle \bw_i^{(t,k)} - \bw^*, \nabla F_i(\bw_i^{(t,k)}) \right\rangle \geq  F_i(\bw_i^{(t,k)}) - F_i(\bw^*)  .
\end{align}
Therefore, adding the above inequalities we have,
\begin{align}
  \left \langle \bw^{(t)}-\bw^*, \nabla F_i(\bw_i^{(t,k)}) \right\rangle \geq F_i(\bw^{(t)}) - F_i(\bw^*) -  \frac{L}{2}\norm{\bw^{(t)}-\bw_i^{(t,k)}} . \label{t1_final_ineq}  
\end{align}

Substituting \Cref{t1_final_ineq} in \Cref{t1_convex_simplify} we have,
\begin{align}
   T_1 \geq \tau(F(\bw^{(t)}) - F(\bw^*)) - \frac{L}{2M}\sum_{i=1}^M \sum_{k=0}^{\tau-1}\norm{\bw^{(t)} - \bw_i^{(t,k)}}.
\end{align}
\textcolor{black}{Here we would like to note that the bound for $T_1$ is our contribution and is needed in our proof due to the relaxation in \cref{thm1-1}. The bound for $T_2$ follows a similar technique as  \citet[Lemma~12]{khaled2020tighter}.}

Substituting the bounds for $T_1$ and $T_2$ in \Cref{convex_start_ineq} we have,
\begin{align}
    \norm{\bw^{(t+1)}-\bw^*} &\leq \norm{\bw^{(t)}-\bw^*} -2\eta_g^{(t)}\eta_l \tau(1-3\eta_l \tau L) (F(\bw^{(t)})-F(\bw^*)) + 3\eta_g^{(t)}\eta_l^2 \tau^2 \sigma_{*}^2 \nonumber\\
    & \hspace{5pt} + (3\eta_g^{(t)}\eta_l^2 \tau L^2 + \eta_g^{(t)}\eta_l L) \frac{1}{M}\sum_{i=1}^M \sum_{k=0}^{\tau-1}\norm{\bw_i^{(t,k)}-\bw^{(t)}} \nonumber \\
    & \leq \norm{\bw^{(t)}-\bw^*} -\eta_g^{(t)}\eta_l\tau (F(\bw^{(t)})-F(\bw^*)) + 3\eta_g^{(t)}\eta_l^2 \tau^2 \sigma_{*}^2 \label{thm1:1}\\
    & \hspace{5pt} + 2\eta_g^{(t)}\eta_l L \frac{1}{M}\sum_{i=1}^M \sum_{k=0}^{\tau-1}\norm{\bw_i^{(t,k)}-\bw^{(t)}} \nonumber \\
    & \leq \norm{\bw^{(t)}-\bw^*} -\eta_g^{(t)}\eta_l\tau (F(\bw^{(t)})-F(\bw^*)) + 3\eta_g^{(t)}\eta_l^2 \tau^2 \sigma_{*}^2 \label{thm1:2}\\
    & \hspace{5pt} + 24\eta_g^{(t)}\eta_l^3 \tau^2(\tau-1)L^2 (F(\bw^{(t)})-F(\bw^*)) + 12\eta_g^{(t)}\eta_l^3 \tau^2(\tau-1) L \sigma^2_{*} \nonumber \\
    & \leq \norm{\bw^{(t)}-\bw^*} -\frac{\eta_g^{(t)}\eta_l\tau}{3} (F(\bw^{(t)})-F(\bw^*)) + 3\eta_g^{(t)}\eta_l^2 \tau^2 \sigma_{*}^2\\
    &\hspace{5pt} + 12\eta_g^{(t)}\eta_l^3\tau^2(\tau-1)L\sigma_{*}^2  \nonumber\label{thm1:3}
\end{align}
where both \cref{thm1:1} and \cref{thm1:3} use $\eta_l \leq \frac{1}{6 \tau L}$, and \cref{thm1:2} uses \Cref{lemma:bounding client drift}.

Rearranging terms and averaging over all rounds we have,
\begin{align}
    \frac{\sum_{t=0}^{T-1}\eta_g^{(t)} F(\bw^{(t)})-F(\bw^*)}{\sum_{t=0}^{T-1}\eta_g^{(t)}} &\leq \frac{3\norm{\bw^{(0)}-\bw^*}}{\sum_{t=0}^{T-1}\eta_g^{(t)}\eta_l \tau} + 9\eta_l\tau\sigma_*^2 + 36\eta_l^2\tau(\tau-1)L\sigma_*^2 \, .
\end{align}

This implies,
\begin{align}
    F(\bar{\bw}^{(T)}) - F(\bw^*) \leq \bigO{\frac{\norm{\bw^{(0)}-\bw^*}}{\eta_l \tau \sum_{t=0}^{T-1}\eta_g^{(t)}}} + \bigO{\eta_l^2 \tau(\tau-1) L\sigma^2_{*}} + \bigO{\eta_l\tau \sigma^2_{*}}
\end{align}
where $\bar{\bw}^{(T)} = \frac{\sum_{t=0}^{T-1}\eta_g^{(t)}\bw^{(t)}}{\sum_{t=0}^{T-1} \eta_g^{(t)}}$.
This completes the proof. 
\qed

\subsection{Convergence Analysis for Non-Convex Objectives}
\label{app:subsec_non_convex}

Our proof technique is inspired by \citet{wang2020tackling} and we use one of their intermediate results to bound client drift in non-convex settings as we describe below. \textcolor{black}{ We highlight the specific steps where we made adjustments to the analysis of \citet{wang2020tackling} below.}

We begin by defining the following auxiliary variables that will used in the proof.
\begin{align}
    \text{Normalized Gradient:} \hspace{10pt} \bh_i^{(t)} = \frac{1}{\tau}\sum_{k=0}^{\tau-1}\nabla F_i(\bw_i^{(t,k)}).
\end{align}
We also define $\bar{\bh}^{(t)} = \frac{1}{M}\sum_{i=1}^M \bh_i^{(t)}$.

\begin{lem} (Bounding client drift in Non-Convex Setting)
\begin{align}
    \frac{1}{M}\sum_{i=1}^M \norm{\nabla F_i(\bw^{(t)})-\bh_i^{(t)}} & \leq \frac{1}{8}\norm{\nabla F(\bw^{(t)}} + 5\eta_l^2L^2\tau(\tau-1)\sigma_g^2\, .
\end{align}
\label{lemma_bounded_drift_nonconvex}
\end{lem}

\textbf{Proof of \Cref{lemma_bounded_drift_nonconvex}:}
Let $D = 4\eta_l^2 L^2 \tau(\tau-1)$. We have the following bound from equation (87) in \cite{wang2020tackling},
\begin{align}
    \frac{1}{M}\sum_{i=1}^M \norm{\nabla F_i(\bw^{(t)})-\bh_i^{(t)}} \leq \frac{D}{1-D} \norm{\nabla F(\bw^{(t)})} + \frac{D \sigma_g^2}{1-D}.
\end{align}   

From $\eta_l \leq \frac{1}{6\tau L}$ we have $D \leq \frac{1}{9}$ which implies $\frac{1}{1-D} \leq \frac{9}{8}$ and $\frac{D}{1-D} \leq \frac{1}{8}$.

Therefore we have,
\begin{align}
    \frac{1}{M}\sum_{i=1}^M \norm{\nabla F_i(\bw^{(t)})-\bh_i^{(t)}} &\leq \frac{1}{8}\norm{\nabla F(\bw^{(t)})} + \frac{9D}{8}\sigma_g^2\\
    & \leq \frac{1}{8}\norm{\nabla F(\bw^{(t)}} + 5\eta_l^2L^2\tau(\tau-1)\sigma_g^2\, .
\end{align}
\qed

\textbf{Proof of Theorem 2:}

The update of the global model can be written as follows,
\begin{align}
    \bw^{(t+1)} = \bw^{(t)} - \eta_g^{(t)}\eta_l\tau \bar{\bh}^{(t)}.
\end{align}

Now using the Lipschitz-smoothness assumption we have,
\begin{align}
    F(\bw^{(t+1)}) - F(\bw^{(t)}) & \leq - \eta_g^{(t)}\eta_l \tau \left\langle \nabla F(\bw^{(t)}), \bar{\bh}^{(t)}\right\rangle + \frac{(\eta_g^{(t)})^2 \eta_l^2 \tau^2 L}{2}\norm{\bar{\bh}^{(t)}}\\
    & \leq -\eta_g^{(t)}\eta_l \tau \left\langle \nabla F(\bw^{(t)}), \bar{\bh^{(t)}}\right\rangle + \frac{\eta_g^{(t)}\eta_l^2 \tau^2 L}{2M}\sum_{i=1}^M\norm{\bh_i^{(t)}} \label{thm2-1}
\end{align}
where \cref{thm2-1} uses $\eta_g^{(t)} \leq \frac{\sum_{i=1}^M \norm{\bh_i^{(t)}}}{M\norm{\bar{\bh}^{(t)}}}$. \textcolor{black}{As in the convex case, inequality \cref{thm2-1} is a key step in our proof and the differentiating factor in our approach from \cite{wang2020tackling}. Following a similar technique as \cite{wang2020tackling} to bound $(\eta_g^{(t)})^2\eta_l^2\tau^2L\norm{\bar{\bh}^{(t)}}/2$ will need the condition $\eta_l \leq 1/2L\tau\eta_g^{(t)}$, which cannot be satisfied in our setup due to the adaptive choice of $\eta_g^{(t)}$. Therefore we first upper bound $(\eta_g^{(t)})^2\eta_l^2 \tau^2L^2\norm{\bar{\bh}^{(t)}}$ by $\eta_g^{(t)}\eta_l^2\tau^2L\frac{1}{M}\sum_{i=1}^M \norm{\bh_i^{(t)}}\Big/2$ and focus on further bounding this quantity in the rest of the proof, which does not require the aforementioned condition. Note that this comes at the expense of the additional $T_3$ error seen in our final  convergence bound in \Cref{thm:theorem_2}.}

Therefore we have,
\begin{align}
     F(\bw^{(t+1)}) - F(\bw^{(t)}) &\leq   -\eta_g^{(t)}\eta_l \tau \underbrace{\left\langle \nabla F(\bw^{(t)}), \bar{\bh}^{(t)} \right\rangle}_{T_1} + \frac{\eta_g^{(t)}\eta_l^2 \tau^2 L}{2M}\underbrace{\sum_{i=1}^M \norm{\bh_i^{(t)}}}_{T_2}.
     \label{thm2-2}
\end{align}

\textbf{Bounding $T_1$}

We have,
\begin{align}
    T_1
    & = \left\langle \nabla F(\bw^{(t)}), \frac{1}{M}\sum_{i=0}^M \bh_i^{(t)}\right\rangle \\
    & = \frac{1}{2}\norm{\nabla F(\bw^{(t)})} + \frac{1}{2}\norm{\frac{1}{M}\sum_{i=1}^M \bh_i^{(t)}} - \frac{1}{2}\norm{\nabla F(\bw^{(t)}) - \frac{1}{M}\sum_{i=1}^M \bh_i^{(t)}} \label{thm2-3-1}\\
    & \geq \frac{1}{2}\norm{\nabla F(\bw^{(t)})} - \frac{1}{2M}\sum_{i=1}^M \norm{\nabla F_i(\bw^{(t)}) - \bh_i^{(t)}} \label{thm2-3-2}
\end{align}
where \cref{thm2-3-1} uses $\left\langle\ba,\bb\right\rangle = 
\frac{1}{2}\norm{\ba} + \frac{1}{2}\norm{\bb} - \frac{1}{2}\norm{\ba - \bb}$ and \cref{thm2-3-2} uses Jensen's inequality and the definition of the global objective function $F$. 

\textbf{Bounding $T_2$}

We have,
\begin{align}
    T_2 &= \frac{1}{M}\sum_{i=1}^M \norm{\bh_i^{(t)}}\\
    & = \frac{1}{M}\sum_{i=1}^M \norm{\bh_i^{(t)}-\nabla F_i(\bw^{(t)}) +\nabla F_i(\bw^{(t)}) - \nabla F(\bw^{(t)}) + \nabla F(\bw^{(t)}) }\\
    & \leq \frac{3}{M}\sum_{i=1}^M\left( \norm{ \bh_i^{(t)}-\nabla F_i(\bw^{(t)})} + \norm{\nabla F_i(\bw^{(t)}) - \nabla F(\bw^{(t)})} + \norm{\nabla F(\bw^{(t)})}\right) \label{thm2-6}\\
    & \leq \frac{3}{M}\sum_{i=1}^M \norm{\bh_i^{(t)}-\nabla F_i(\bw^{(t)})} + 3 \sigma^2_g + 3\norm{\nabla F(\bw^{(t)})} \label{thm2-6-1} 
\end{align}
where \cref{thm2-6} uses Jensen's inequality, \cref{thm2-6-1} uses bounded data heterogeneity assumption.

\textcolor{black}{Here we would like to note that the bound for $T_2$ is our contribution and is needed in our proof due to the relaxation in \cref{thm1-1}. The bound for $T_1$ follows a similar technique as in \cite{wang2020tackling}.}

Substituting the $T_1$ and $T_2$ bounds into \cref{thm2-2}, we have,
\begin{align}
   F(\bw^{(t+1)}) - F(\bw^{(t)}) &\leq   -\eta_g^{(t)}\eta_l \tau\Bigg(\frac{1}{2}\norm{\nabla F(\bw^{(t)})} + \frac{1}{2M}\sum_{i=1}^M \norm{\nabla F_i(\bw^{(t)})-\bh_i^{(t)}} \\ 
   & \hspace{10pt} + \frac{\eta_l \tau L}{2} \left(3\sigma_g^2 + 3\norm{\nabla F(\bw^{(t)})} +  \frac{3}{M}\sum_{i=1}^M \norm{\bh_i^{(t)}-\nabla F_i(\bw^{(t)})} \right) \Bigg)\nonumber \\
   & \leq -\eta_g^{(t)}\eta_l \tau \left(\frac{1}{4}\norm{\nabla F(\bw^{(t)})} + \frac{1}{M}\sum_{i=1}^M \norm{\nabla F_i(\bw^{(t)})-\bh_i^{(t)}} + 3\eta_l \tau L\sigma_g^2 \right)\label{thm2-7}\\
   & \leq -\eta_g^{(t)}\eta_l \tau \left(\frac{1}{8}\norm{\nabla F(\bw^{(t)}} + 3\eta_l \tau L \sigma_g^2 + 5\eta_l^2 L^2 \tau(\tau-1)\sigma_g^2 \right)\label{thm2-8}
\end{align}

where \cref{thm2-7} uses $\eta_l \leq \frac{1}{6\tau L}$, \cref{thm2-8} uses \Cref{lemma_bounded_drift_nonconvex}.

Thus rearranging terms and averaging over all rounds we have,
\begin{align}
    \frac{\sum_{t=0}^{T-1}\eta_g^{(t)} \norm{\nabla F(\bw^{(t)})}}{\sum_{t=0}^{T-1}\eta_g^{(t)}} &\leq \frac{8(F(\bw^{(0)})-F^*)}{\sum_{t=0}^{T-1}\eta_g^{(t)}\eta_l \tau} +  40\eta_l^2L^2\tau(\tau-1)\sigma_g^2 + 24 \eta_l L \tau \sigma_g^2 \, .
\end{align}
This implies,
\begin{align}
    & \min_{t \in [T]} \norm{\nabla F(\bw^{(t)})} \leq \bigO{\frac{(F(\bw^{(0)})-F^*)}{\sum_{t=0}^{T-1}\eta_g^{(t)}\eta_l \tau}} + \bigO{\eta_l^2L^2\tau(\tau-1)\sigma_g^2} + \bigO{\eta_l L \tau \sigma_g^2} .
\end{align}

This completes the proof.
\qed

\subsection{Exact Projection with Gradient Descent for Linear Regression}
\label{app:subsec_linear_regression_proj}

Let $F(\bw) = \norm{\bA \bw - \bb}$ where $\bA$ is a $(n \times d)$ matrix and $\bb$ is a $n$ dimensional vector. We assume that $d \geq n$ here and $\bA$ has rank $n$. The singular value decomposition (SVD) of $\bA$ can be written as,
\begin{align}
    \bA = \bU \bSigma \bV^{\top} = \bU \begin{bmatrix}
\bSigma_1 & \mathbf{0}
\end{bmatrix}
\begin{bmatrix}
\bV_1^{\top}\\
\bV_2^{\top}
\end{bmatrix} & = \bU \bSigma_1 \bV_1^{\top}
\end{align}
where $\bU$ is an $(n \times n)$ orthogonal matrix, $\bSigma$ is an $(n \times n)$ diagonal matrix, $\bV_1$ is a $(d \times n)$ matrix with orthogonal columns and $\bV_2$ is a $(d \times (d-n))$ matrix with orthogonal columns. Here $\bV_1$ is a basis for the row space of $\bA$, while $\bV_2$ is a basis for the null space of $\bA$. We first prove the following lemmas about the set of minimizers of $F(\bw)$ and the projection on this set.

\begin{lem} The set of minimizers of $F(\bw)$ is given by,

\begin{align}
    \mathcal{S}^*  = \{\bV_2\bV_2^{\top}\bw + \bV_1 \bSigma_1^{-1} \bU^{\top}\bb| \bw \in \mathbb{R}^d \}.
\end{align}

\end{lem}

\textbf{Proof.}
Let $\bw = \bV_2\bV_2^{\top}\bx + \bV_1 \bSigma_1^{-1} \bU^{\top}\bb$ for some $\bx \in \mathbb{R}^d$. We have, 
\begin{align}
    \bA\bw &= \bU \bSigma_1 \bV_1^{\top}(\bV_2\bV_2^{\top}\bx + \bV_1 \bSigma_1^{-1} \bU^{\top}\bb)\\
    & = \bb
\end{align}
where the last line uses $\bV_1^{\top}\bV_2 = 0, \bV_1^{\top}\bV_1 = \bI, \bU\bU^{\top} = \bI$. This implies $\norm{\bA\bw - \bb} = 0$. Thus any $\bw$ in $\mathcal{S}^*$ is a minimizer of $F(\bw)$.

Now let $\bw^*$ be a minimizer of $F(\bw)$, implying $\bA\bw^* = \bU \bSigma_1 \bV_1^{\top} \bw^* = \bb$. We have,
\begin{align}
    \bw^* &= \bV_2\bV_2^{\top}\bw^* + \bV_1\bV_1^{\top}\bw^* \label{eq:lemma8_first}\\
    & = \bV_2\bV_2^{\top}\bw^* + \bV_1 \bSigma_1^{-1} \bU^{\top}\bb \label{eq:lemma8_second}
\end{align}
where \cref{eq:lemma8_first} uses $\bV_1\bV_1^{\top} + \bV_2\bV_2^{\top} = \bI$ and \cref{eq:lemma8_second} uses $\bU \bSigma_1 \bV_1^{\top} \bw^* = \bb$.
Thus any minimizer of $F(\bw)$ must lie in $\mathcal{S}^*$.

Combining the above statements we have,
\begin{align}
    \bw \text{ is a minimizer of } F(\bw) \iff \bw \in \mathcal{S}^*.
\end{align}
which completes the proof.
\qed

\begin{lem}
The projection of any $\bw \in \mathbb{R}^{d}$ on $\mathcal{S}^*$ is given by,
\begin{align}
    P_{\mathcal{S^*}}(\bw) = \argmin_{\bw' \in \mathcal{S}^*} \norm{\bw - \bw'} & = \bV_2 \bV_2^{\top}\bw + \bV_1 \bSigma_1^{-1} \bU^{\top}\bb  .
\end{align}
\end{lem}

\textbf{Proof.}
When $\bw \in \mathcal{S}^*$, it is easy to see that this holds. Therefore we consider the case where $\bw \notin \mathcal{S}^*$. Let $\bx = \bV_2 \bV_2^{\top}\bw + \bV_1 \bSigma_1^{-1} \bU^{\top}\bb$ and $P_{\mathcal{S}^*}(\bw) = \bV_2 \bV_2^{\top}\bw_0 + \bV_1 \bSigma_1^{-1} \bU^{\top}\bb$ where $\bw_0 \neq \bw$. We have,
\begin{align}
    & \norm{\bw - \bV_2 \bV_2^{\top}\bw_0 - \bV_1 \bSigma_1^{-1} \bU^{\top}\bb} \\
    & = \norm{\bV_2 \bV_2^{\top}(\bw-\bw_0) + \bV_1 \bV_1^{\top}\bw - \bV_1 \bSigma_1^{-1} \bU^{\top}\bb} \hspace{10pt} (\bV_1\bV_1^{\top} + \bV_2\bV_2^{\top} = \bI)\\
    & = \norm{\bV_2 \bV_2^{\top}(\bw-\bw_0)} + \norm{ \bV_1 \bV_1^{\top}\bw - \bV_1 \bSigma_1^{-1} \bU^{\top}\bb} \label{eq:exact_proj_cross_term}\\
    & = \norm{\bV_2 \bV_2^{\top}(\bw-\bw_0)} + \norm{\bw -\bx } \label{eq:exact_proj_x}\\
    & > \norm{\bw -\bx }
\end{align}
leading to a contradiction. The cross term in \cref{eq:exact_proj_cross_term} is zero since $\bV_1^{\top}\bV_2 = \mathbf{0}$. Equation \cref{eq:exact_proj_x} follows by the definition of $\mathbf{x}$. \qed

We now show that running gradient descent on $F(\bw)$ starting from $\bw$ with a sufficiently small step size converges to $P_{\mathcal{S}^*}(\bw)$.

\begin{lem}
Let $\bw^{(0)},\bw^{(1)},\dots$ be the iterates generated by running gradient descent on $F(\bw)$ with $\bw^{(0)} = \bw$ and learning rate $\eta_l \leq \lambda_{\max}$, where $\lambda_{\max}$ is the largest eigen value of $\bA^{\top}\bA$. Then $\lim_{T \rightarrow \infty} \bw^{(T)} = P_{\mathcal{S}^*}(\bw)$.
\end{lem}

\textbf{Proof.} 
By the gradient descent update we have,
\begin{align}
    \bw^{(t+1)} &= \bw^{(t)} - \eta_l(\bA^{\top}\bA \bw^{(t)} - \bA^{\top}\bb)\\
    & = (\bI - \eta_l\bA^{\top}\bA)\bw^{(t)}+ \eta_l\bA^{\top}\bb .
\end{align}
Therefore,
\begin{align}
    \bw^{(T)} &= (\bI - \eta_l\bA^{\top}\bA)^{T}\bw^{(0)}+ \eta_l\sum_{t=0}^{T-1}(\bI - \eta_l\bA^{\top}\bA)^{t}\bA^{\top}\bb\\
    & = \bV(\bI - \eta_l\bSigma^{\top}\bSigma)^{T}\bV^{\top}\bw^{(0)} + \eta_l\sum_{t=0}^{T-1}\bV(\bI - \eta_l\bSigma^{\top}\bSigma)^{t}\bSigma^{\top}\bU^{\top}\bb\\
    & = (\bV_1 (\bI - \eta_l \bSigma_1^2)^T\bV_1 + \bV_2\bV_2^{\top})\bw^{(0)} + \eta_l \bV_1 \left(\sum_{t=0}^{T-1}(\bI - \eta_l \bSigma_1^2)^t\right)\bSigma_1 \bU^{\top}\bb .
\end{align}

In the limit $T \rightarrow \infty$ and with $\eta_l \leq \lambda_{\max}$, we have,
\begin{align}
    \lim_{T \rightarrow \infty} (\bI - \eta_l \bSigma_1^2)^T = \mathbf{0} \text{ and } \lim_{T \rightarrow \infty} \sum_{t=0}^{T-1}(\bI - \eta_l \bSigma_1^2)^t = \frac{1}{\eta_l}\bSigma_{1}^{-2}.
\end{align}
Thus,
\begin{align}
    \lim_{T \rightarrow \infty} \bw^{(T)} &= \bV_2 \bV_2^{\top}\bw^{(0)}+ \bV_1 \bSigma_1^{-1} \bU^{\top}\bb \\
    &= P_{\mathcal{S}^*}(\bw^{(0)})\\
    & = P_{\mathcal{S}^*}(\bw).
\end{align}
\qed

\subsubsection{Improving Lower Bound in \cref{eq:eta_g_lb} in the Case of Exact Projections}
\label{app:subsec_linear_regression_lb}

Let $\mathcal{S}_i^*$ be convex and let $\bw^* \in \mathcal{S}_i$ for all $i \in [M]$. We assume that $\bw_i^{(t,\tau)} = P_{\mathcal{S}_i^*}(\bw^{(t)}) \; \forall i \in [M]$, i.e., the local models are an exact projection of $\bw^{(t)}$ on their respective solution sets. From \cref{eq:eta_g_lb} we have,
\begin{align}
\textstyle
    (\eta_g^{(t)})_{\text{opt}} &=  %
    \frac{\left \langle \bw^{(t)}-\bw^*, \bardeltat \right \rangle}{\norm{\bardeltat}} = \frac{\sum_{i=1}^M \left \langle \bw^{(t)} - \bw^*,\Delta_i^{(t)} \right\rangle}{M\norm{\bardeltat}}.
\end{align}
We can lower bound $\left \langle \bw^{(t)} - \bw^*,\Delta_i^{(t)} \right\rangle$ as follows,
\begin{align}
   \left \langle \bw^{(t)} - \bw^*,\Delta_i^{(t)} \right\rangle &=   \left \langle \bw^{(t)} - \bw_i^{(t,\tau)} + \bw_i^{(t,\tau)} - \bw^*,\bw^{(t)}-\bw_i^{(t,\tau)} \right\rangle\\
   & = \norm{\bw^{(t)}-\bw_i^{(t,\tau)}} + \left\langle \bw_i^{(t,\tau)}-\bw^*,\bw^{(t)}-\bw_i^{(t,\tau)} \right\rangle \\
   & \geq \norm{\bw^{(t)}-\bw_i^{(t,\tau)}} \label{eq:cross_term_geq_0}\\
   & = \norm{\Delta_i^{(t)}}\\
\end{align}
where \cref{eq:cross_term_geq_0} uses the fact that $\left\langle \bw_i^{(t,\tau)}-\bw^*,\bw^{(t)}-\bw_i^{(t,\tau)} \right\rangle \geq 0$ following the properties of projection \citep{fundament}.

Thus we have,
\begin{align}
\textstyle
    (\eta_g^{(t)})_{\text{opt}} &\geq %
    \frac{\sum_{i=1}^M \normsmall{\Delta_i^{(t)}}}{M \norm{\bardeltat}}
    \label{eq:eta_g_improv_lb}
\end{align}

Note here the improvement by a factor of 2 in the lower bound compared to \cref{eq:eta_g_lb}.

\newpage

\section{Additional Experiments and Setup Details}
\label{app:sec_experiments}

Our code is available at the following link \url{https://github.com/Divyansh03/FedExP}. 

\subsection{Impact of Averaging Iterates for Neural Networks}
\label{app:subsec_averaging_in_nn}
\begin{figure}[H]
 \centering
 \includegraphics[height=0.29\linewidth]{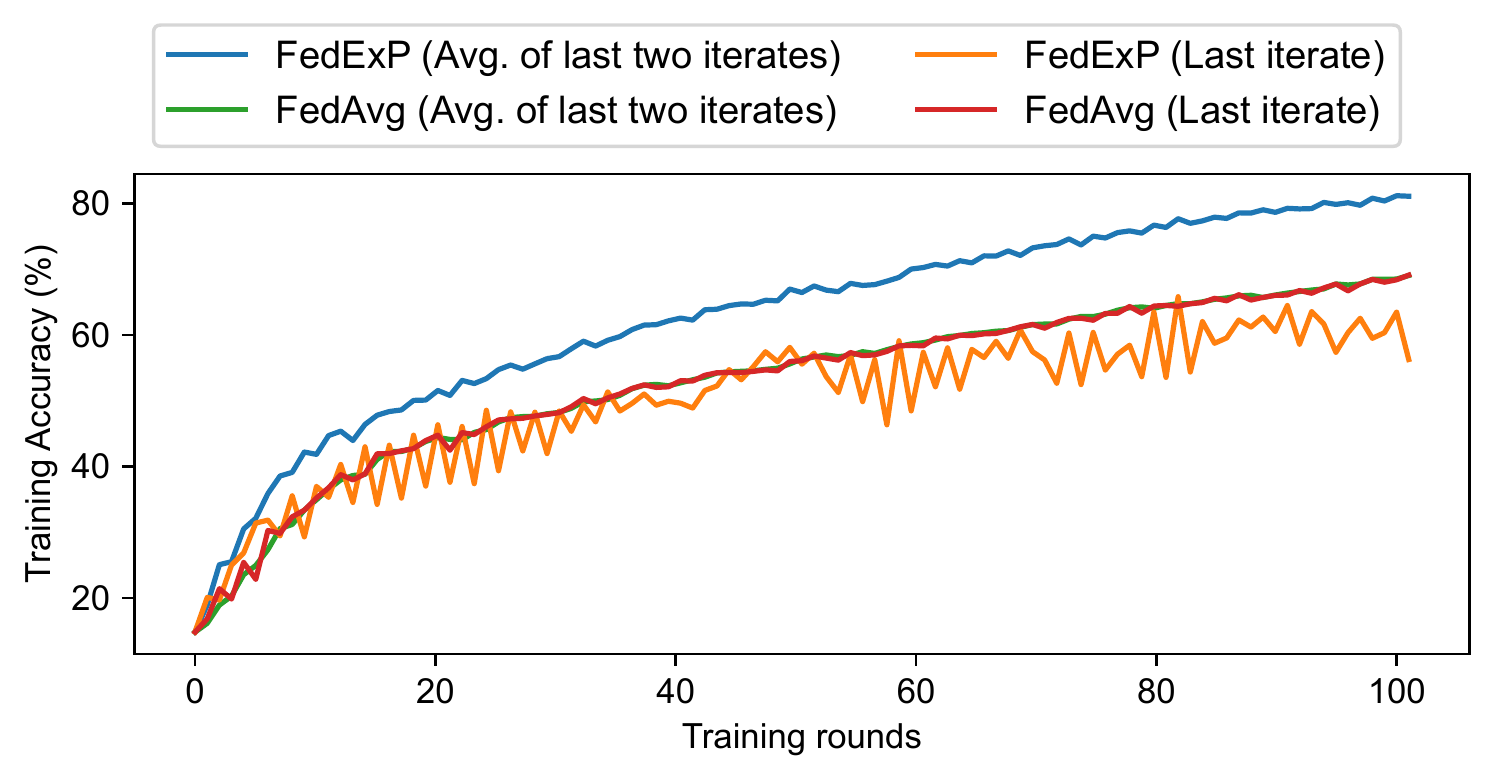}  
 \includegraphics[height=0.29\linewidth]{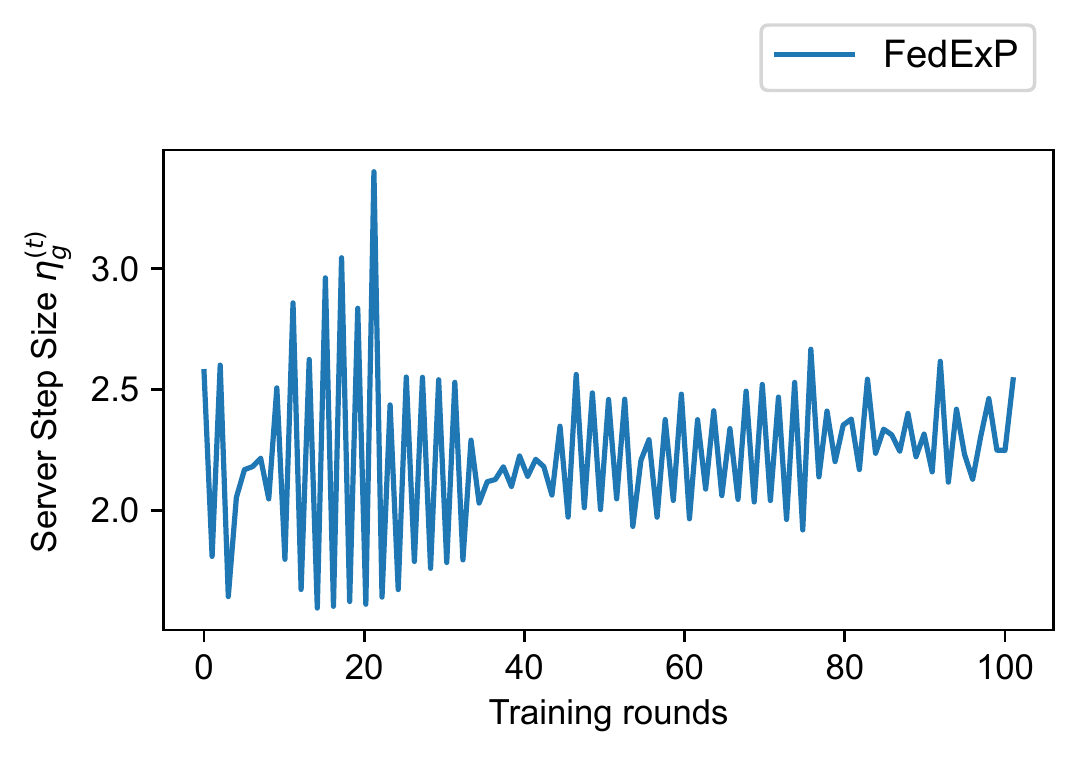}  \caption{Benefit of averaging the last two iterates for \texttt{FedExP} in training a CNN model on \mbox{CIFAR-10}. 
 Note that averaging does not make significant difference for \texttt{FedAvg}. }
\label{fig:toy_averaging_iterates}
\end{figure}

As discussed in \Cref{sec:practical_insights}, we find that setting the final \texttt{FedExP} model as the average of the last two iterates also improves performance when training neural networks in practical FL scenarios. To demonstrate this, we consider an experiment on the CIFAR-10 dataset with 10 clients, where the data at each client is distributed using a Dirichlet distribution with $\alpha = 0.3$. We set the number of local steps to be $\tau = 20$ and train a CNN model having the same architecture as outlined in \cite{mcmahan2017communication} with full client participation. \Cref{fig:toy_averaging_iterates} shows the training accuracy as a function of the last iterate and the average of last two iterates for \texttt{FedAvg} and \texttt{FedExP}. We see that the last iterate of \texttt{FedExP} has an oscillating behavior that can hide improvements in training accuracy. On the other hand, the average of the last two iterates of \texttt{FedExP} produces a more stable training curve and shows a considerable improvement in the final accuracy. Note however that this improvement only shows for \texttt{FedExP}; averaging iterates does not make significant difference for \texttt{FedAvg}.

\subsection{Dataset Details}
\label{app:subsec_dataset_details}
Here we provide more details about the datasets used in \Cref{sec:experiments}.

\paragraph{Synthetic Linear Regression.} In this case we assume that the local objective of each client is given by $F_i(\bw) = \norm{\bA_i \bw - \bb_i}$ where $\bA_i \in \mathbb{R}^{(30 \times 1000)}$, $\bb_i \in \mathbb{R}^{30}$ and $\bw \in \mathbb{R}^{1000}$. We set the number of clients to be $M = 20$. Note that since $d \geq \sum_{i=1}^M n_i$, this is an overparameterized convex problem. To generate $\bA_i$ and $\bb_i$, we follow a similar process as \cite{li2020federated}. We have $(\bA_i)_{j:} \sim \mathcal{N}(\bmm_i,\bI_d)$ and $(\bb_i)_j = \bw_i^{\top}(\bA_i)_{j:}$ where $\bmm_i \sim \mathcal{N}(u_i,1), \bw_i \sim \mathcal{N}(y_i,1), u_i \sim \mathcal{N}(0,0.1), y_i \sim \mathcal{N}(0,0.1)$. 

\paragraph{EMNIST.} EMNIST is an image classification task consisting of handwritten characters associated with 62 labels. The federated EMNIST dataset available at \cite{caldas2018leaf} is naturally partitioned into 3400 clients based on the identities of the character authors. The number of training and test samples is 671,585 and 77,483 respectively. 

\paragraph{CIFAR-10/100.} CIFAR-10 is a natural image dataset consisting of 60,000 32x32 images divided into 10 classes. CIFAR-100 uses a finer labeling of the CIFAR images to divide them into 100 classes making it a harder dataset for image classification. In both cases the number of training examples and test examples is 50,000 and 10,000 respectively. To simulate a federated setting, we artificially partition the training data into 100 clients following the procedure outlined in \cite{hsu2019measuring}.  

\paragraph{CINIC-10.} CINIC-10 is a natural image dataset that can be used as a direct replacement of CIFAR for machine learning tasks. It is intended to act as a harder dataset than CIFAR-10 while being easier than CIFAR-100. The number of training and test examples is both 90,000. We partition the training data into 200 clients in this case, following a similar procedure as for CIFAR.

\subsection{Hyperparameter Details}
\label{app:subsec_additional_experiments}

For our baselines, we find the best performing $\eta_g$ and $\eta_l$ by grid-search tuning. For \texttt{FedExP} we search for $\epsilon$ and $\eta_l$. This is done by running algorithms for 50 rounds and finding the parameters that achieve the highest training accuracy averaged over the last 10 rounds. We provide details of the grid used below for each experiment below.

\paragraph{Grid for Synthetic.}~\\
For \texttt{FedAvg} and \texttt{SCAFFOLD}, the grid for $\eta_g$ is $\{10^{0},10^{0.5},10^{0.5},10^{1},10^2\}$. For \texttt{FedAdagrad}, the grid for $\eta_g$ is $\{10^{-1},10^{-0.5},10^{-0},10^{0.5},10^{1}\}$. For \texttt{FedExP} we keep $\epsilon = 0$ in this experiment as \cref{prop_1} is satisfied in this case. The grid for $\eta_l$ is  $\{10^{-2},10^{-1.5},10^{-1},10^{-0.5},10^{0}\}$ for all algorithms.

\paragraph{Grid for Neural Network Experiments.}~\\
For \texttt{FedAvg} and \texttt{SCAFFOLD} the grid for $\eta_g$ is $\{10^{-1},10^{-0.5},10^{0},10^{0.5},10^1\}$. For \texttt{FedAdagrad}, the grid for $\eta_g$ is $\{10^{-2},10^{-1.5},10^{-1},10^{-0.5},10^{0}\}$. For \texttt{FedExP} the grid for $\epsilon$ is $\{10^{-3},10^{-2.5},10^{-2},10^{-1.5},10^{-1}\}$. The grid for $\eta_l$ is $\{10^{-2},10^{-1.5},10^{-1},10^{-0.5},10^{0}\}$ for all algorithms.

We use lower values of $\eta_g$ in the grid for \texttt{FedAdagrad} based on observations from \cite{reddi2020adaptive} which show that \texttt{FedAdagrad} performs better with smaller values of the server step size. We provide details of the best performing hyperparameters below.

\begin{table}[!h]
\caption{Base-10 logarithm of the best combination of $\epsilon$ and $\eta_l$ for \texttt{FedExP} and combination of $\eta_l$ and $\eta_g$ for baselines. For the synthetic dataset we keep $\epsilon = 0$ for \texttt{FedExP}.}
\centering
  \begin{tabular}{lSSSSSSSS}
    \toprule
    \multirow{2}{*}{Dataset} &
      \multicolumn{2}{c}{ \texttt{FedExP}} &
      \multicolumn{2}{c}{\texttt{FedAvg}} &
      \multicolumn{2}{c}{\texttt{SCAFFOLD}} &
      \multicolumn{2}{c}{\texttt{FedAdagrad}} \\
      & {$\epsilon$} & {$\eta_l$} & {$\eta_g$} & {$\eta_l$} & {$\eta_g$} & {$\eta_l$} & {$\eta_g$} & {$\eta_l$}\\
      \midrule
    Synthetic & * & -1 &  1 & -1 & 1 & -1 & -1 & -1 \\
    EMNIST & -1 & -0.5 & 0 & -0.5 & 0 & -0.5 & -0.5 & -0.5\\
    CIFAR-10 & -3 & -2 &  0 & -2 & 0 & -2 & -1 & -2 \\
    CIFAR-100 & -3 & -2 &  0 & -2 & 0 & -2 & -1 & -2 \\
    CINIC-100 & -3 & -2 &  0 & -2 & 0 & -2 & -1 & -2 \\
    \bottomrule
  \end{tabular}
\end{table}

Other hyperparameters are kept the same for all algorithms. In particular, we apply a weight decay of 0.0001 for all algorithms and decay $\eta_l$ by a factor of 0.998 in every round. We also use gradient clipping to improve stability of the algorithms as done in previous works \citep{acar2021federated}. In all experiments we fix the number of participating clients to be 20, minibatch size to be 50 (for the synthetic dataset this reduces to full-batch gradient descent) and number of local updates $\tau$ to be 20.

\color{black}

\subsection{Sensitivity of FedExP to $\epsilon$}

To evaluate the sensitivity of \texttt{FedExP} to $\epsilon$, we compute the training accuracy of \texttt{FedExP} after 500 rounds for varying $\epsilon$ and on different tasks. For each task, we fix $\eta_l$ to be the value used in our experiments in Section 6 and only vary $\epsilon$. The results are summarized below.

\captionsetup[table]{labelfont={color=black},font={color=black}}

\begin{table}[H]
\centering
\caption{Training accuracy obtained by \texttt{FedExP} with different choices of $\epsilon$ after 500 rounds of training on various tasks. Value of $\eta_l$ is fixed for each task ($10^{-0.5}$ for EMNIST and $10^{-2}$ for others). Results averaged over last 10 rounds.}\vspace{-0.5em}
\color{black}
{\small
\begin{tabular}{c c c c c c c c}
\thickhline \\[-0.2cm]
Dataset & $\epsilon \!=\! 10^{-3}$ & $\epsilon \!=\! 10^{-2.5} $ & $\epsilon\!=\!10^{-2}$ & $\epsilon \!=\! 10^{-1.5}$ & $\epsilon \!=\! 10^{-1}$ \\ [0.05cm]
\hline\\[-0.2cm]
EMNIST & $85.40$ & $\mathbf{86.26}$ & $85.73$ & $85.49$ & $84.90$\\
CIFAR-10 & $\mathbf{84.79}$ & $77.82$ & 
$77.63$ & $77.66$ & $77.64$\\
CIFAR-100 & $\mathbf{59.01}$ & $44.76$ & $44.21$ & $44.37$ & $44.40$\\
CINIC-10 & $\mathbf{66.31}$ & $60.93$ & $61.05$ &  $60.47$ & $60.96$\\
\thickhline\\[-0.2cm]
\end{tabular}
}
\end{table}

We see that the sensitivity of $\epsilon$ is similar to that of the $\tau$ parameter which is added to the denominator of \texttt{FedAdam} and \texttt{FedAdagrad} \citep{reddi2020adaptive} to prevent the step size from blowing up. Keeping $\epsilon$ too large reduces the adaptivity of the method and makes the behavior similar to \texttt{FedAvg}. At the same time, keeping $\epsilon$ too small may not also be beneficial always as seen in the case of EMNIST.  In practice, we find that a grid search for $\epsilon$ in the range $\{10^{-3},10^{-2.5},10^{-2},10^{-1.5},10^{-1}\}$ usually suffices to yield a good value of $\epsilon$. A general rule of thumb would be to start with $\epsilon = 10^{-3}$ and increase $\epsilon$ till the performance drops.

\color{black}
\subsection{Additional Results}
\label{app:subsec_addtl_results}

In this section, we provide additional results obtained from our experiments.

\begin{figure}[H]
 \centering
 \includegraphics[width=1\linewidth]{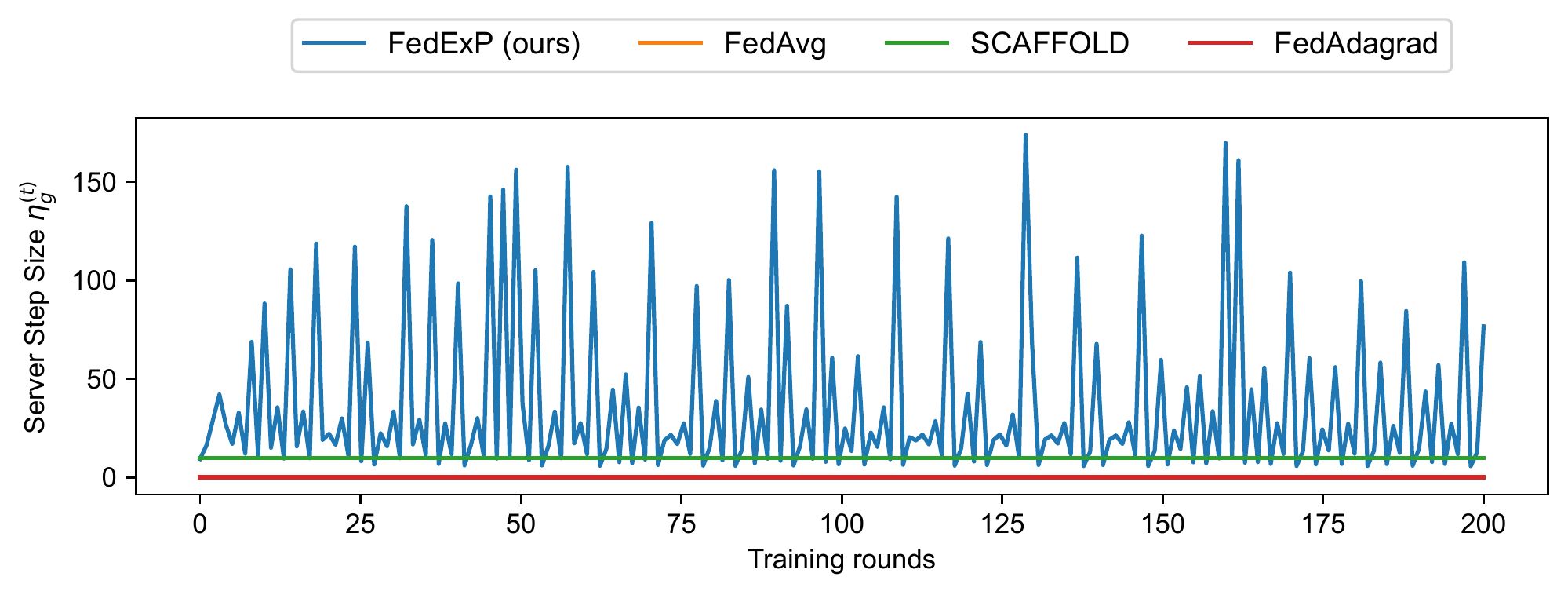}  \caption{Global learning rates for synthetic data with linear regression. Results from a single instance of experiment.}
\label{fig:addtl_results_synthetic}
\end{figure}

\paragraph{Synthetic Linear Regression.}
Note that for the synthetic linear regression experiments there is no test data. Also note that there is no randomness in this experiment since clients compute full-batch gradients with full participation. We provide the plot of $\eta_g^{(t)}$ for \texttt{FedExP} in \Cref{fig:addtl_results_synthetic}. We see that \texttt{FedExP} takes much larger steps in some (but not all) rounds compared to the constant optimum step size taken by our baselines, leading to a large speedup. Recall that we also let $\epsilon = 0$ in this experiment (since it aligns with our theory) which also explains the larger values of $\eta_g^{(t)}$ taken by \texttt{FedExP} in this case. 

\begin{figure}[H]
 \centering
 \includegraphics[width=1\linewidth]{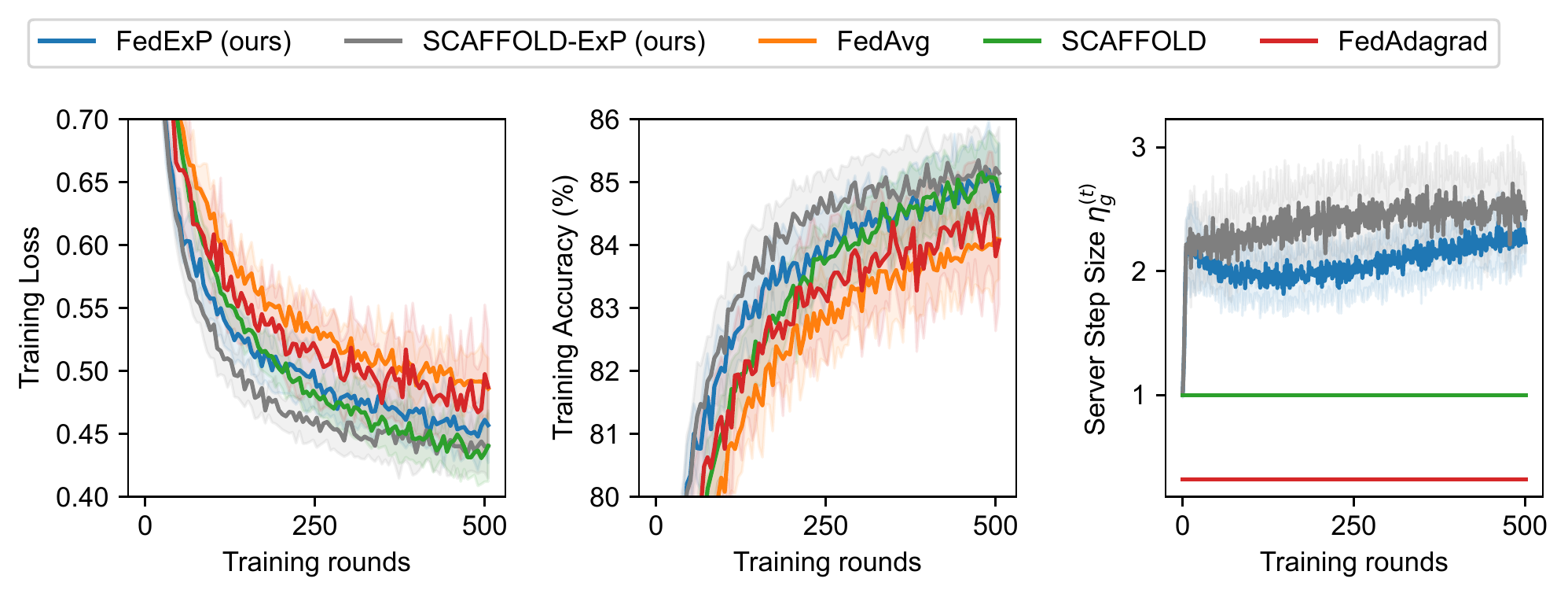}  \caption{Additional results for EMNIST dataset. Mean and standard deviation from experiments with $20$ different random seeds. The shaded areas show the standard deviation.}
\label{fig:addtl_results_emnist}
\end{figure}

\paragraph{EMNIST.} For EMNIST we observe that \texttt{SCAFFOLD} gives slightly better training loss than \texttt{FedExP} towards the end of training. As described in \Cref{sec:experiments}, extrapolation can be combined with the variance-reduction in \texttt{SCAFFOLD} (the resulting algorithm is referred to as \texttt{SCAFFOLD-ExP}) to further improve performance. This gives the best result in this case as shown in \Cref{fig:addtl_results_emnist}.

\begin{figure}[H]
 \centering
 \includegraphics[width=1\linewidth]{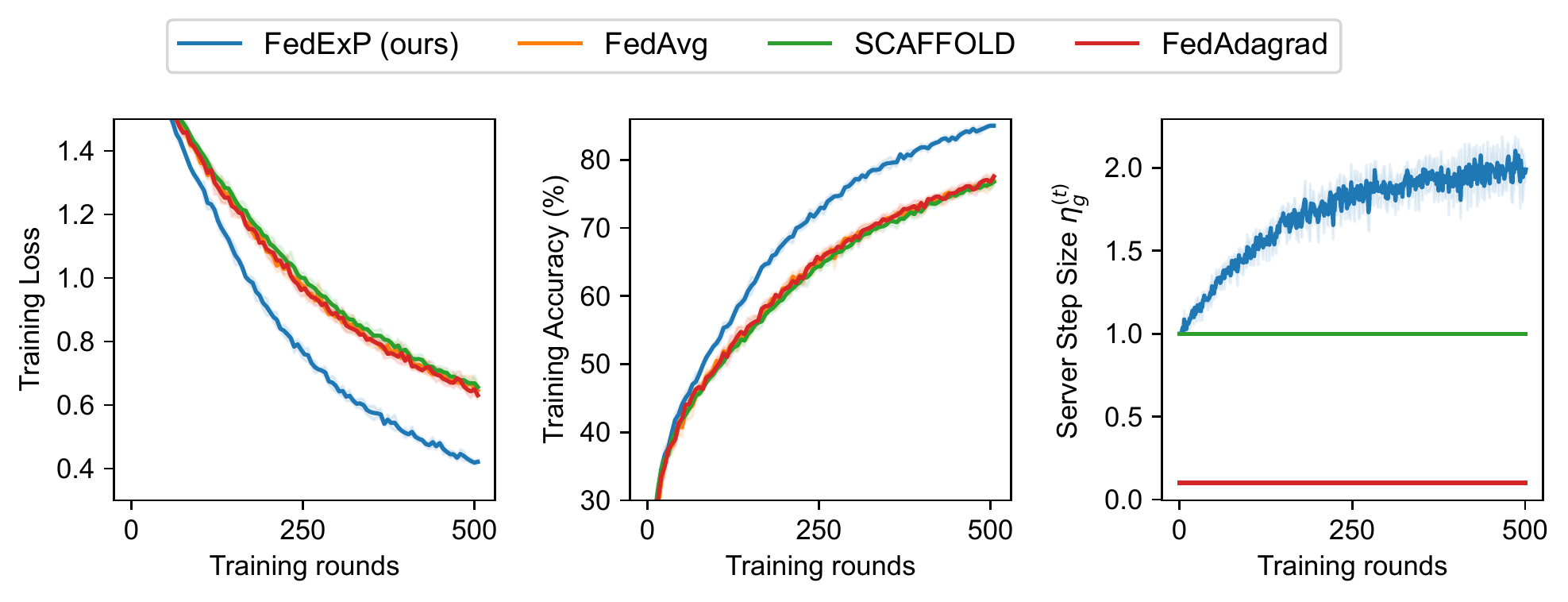}\vspace{-0.5em}  \caption{Additional results for CIFAR-10 dataset. Mean and standard deviation from experiments with $5$ different random seeds. The shaded areas show the standard deviation.}
\label{fig:addtl_results_cifar10}
\end{figure}

\begin{figure}[H]
 \centering
 \includegraphics[width=1\linewidth]{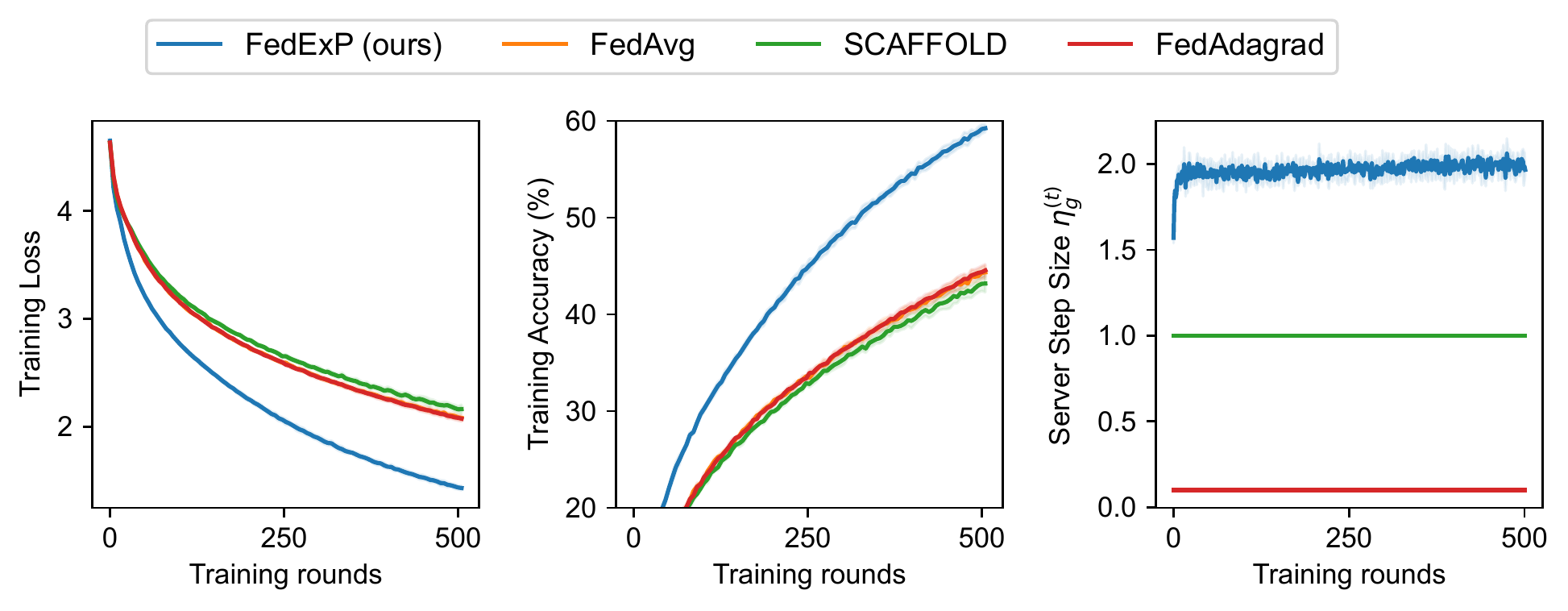}\vspace{-0.5em}  \caption{Additional results for CIFAR-100 dataset. Mean and standard deviation from experiments with $5$ different random seeds. The shaded areas show the standard deviation.}
\label{fig:addtl_results_cifar100}
\end{figure}

\begin{figure}[H]
 \centering
 \includegraphics[width=1\linewidth]{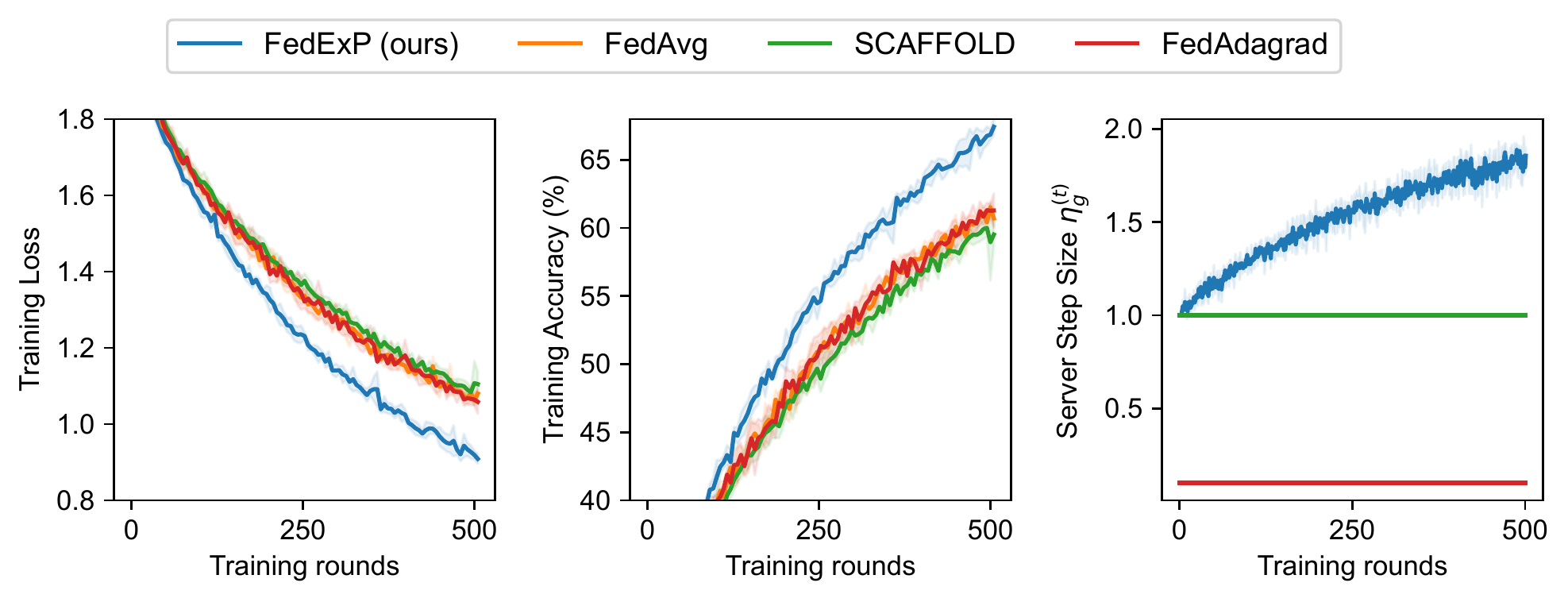}\vspace{-0.5em}  \caption{Additional results for CINIC-10 dataset. Mean and standard deviation from experiments with $5$ different random seeds. The shaded areas show the standard deviation.}
\label{fig:addtl_results_cinic10}
\end{figure}

\paragraph{CIFAR-10, CIFAR-100 and CINIC-10.}
From \Cref{fig:exp_results} and Figures~\ref{fig:addtl_results_cifar10}--\ref{fig:addtl_results_cinic10}, we see that \texttt{FedExP} comprehensively outperforms baselines in these cases, achieving almost $10\%$--$20\%$ higher accuracy than the closest baseline by the end of training. The margin of improvement is most in CIFAR-100, which can be considered as the toughest dataset in our experiments. This points to the practical utility of \texttt{FedExP} even in challenging FL scenarios.  

\newpage

\captionsetup[figure]{labelfont={color=black},font={color=black}}

\color{black}

\begin{figure}[H]
 \centering
 \includegraphics[width=1\linewidth]{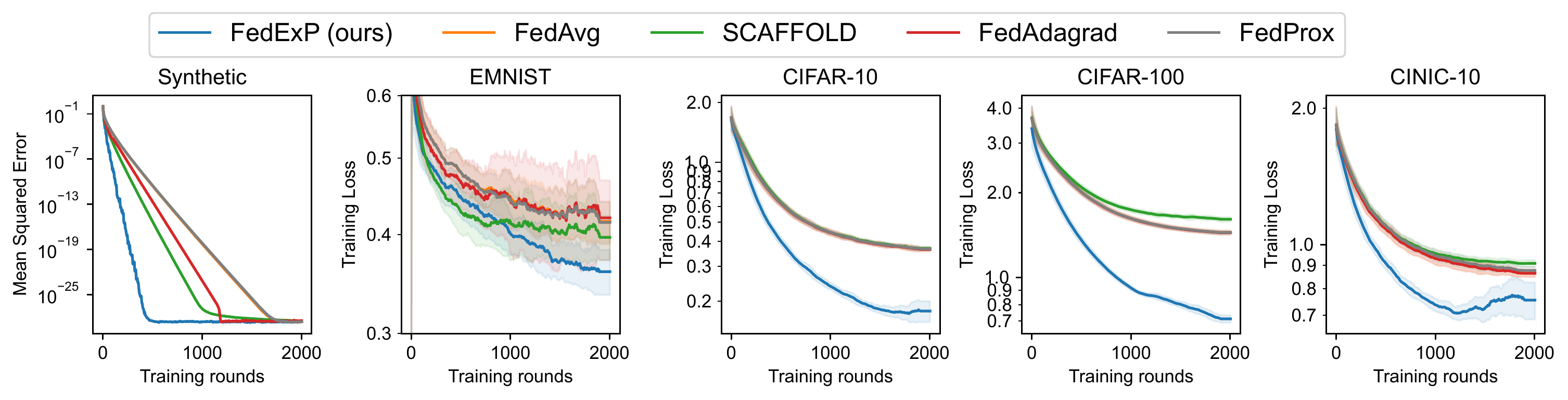}  \caption{Training loss results of \texttt{FedExP}, \texttt{FedAvg}, \texttt{SCAFFOLD}, \texttt{FedAdagrad} and \texttt{FedProx} on the Synthetic, EMNIST, CIFAR-10,CIFAR-100 and CINIC-10 datasets for $2000$ rounds.}
\label{fig:addtl_results_2000_runs_train_loss}
\end{figure}

\begin{figure}[H]
 \centering
 \includegraphics[width=1\linewidth]{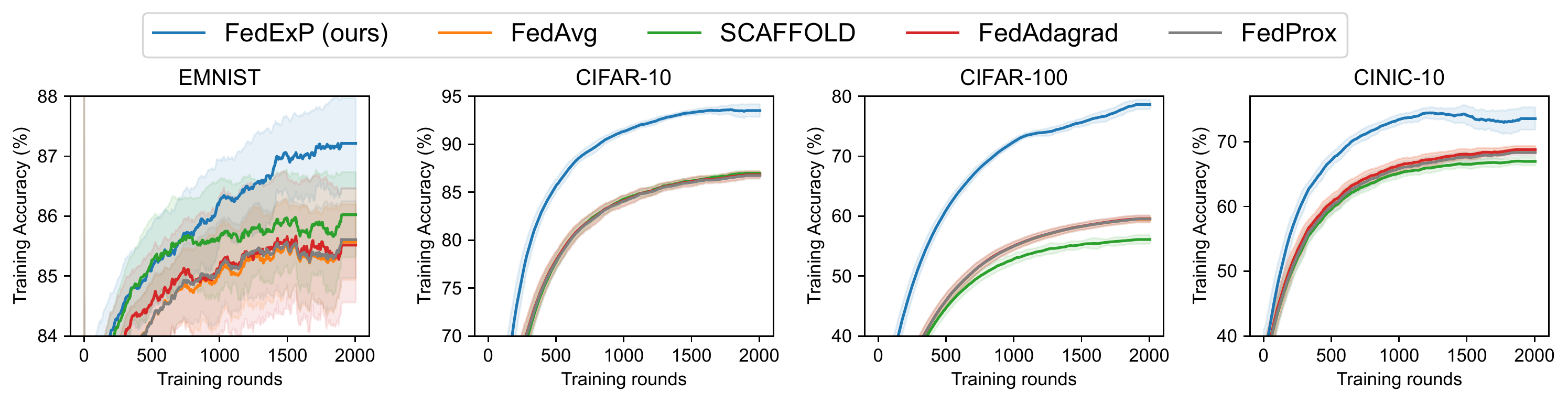}  \caption{Training accuracy results of \texttt{FedExP}, \texttt{FedAvg}, \texttt{SCAFFOLD}, \texttt{FedAdagrad} and \texttt{FedProx} on the EMNIST, CIFAR-10, CIFAR-100 and CINIC-10 datasets for $2000$ rounds.}
\label{fig:addtl_results_2000_runs_train_acc}
\end{figure}

\begin{figure}[H]
 \centering
 \includegraphics[width=1\linewidth]{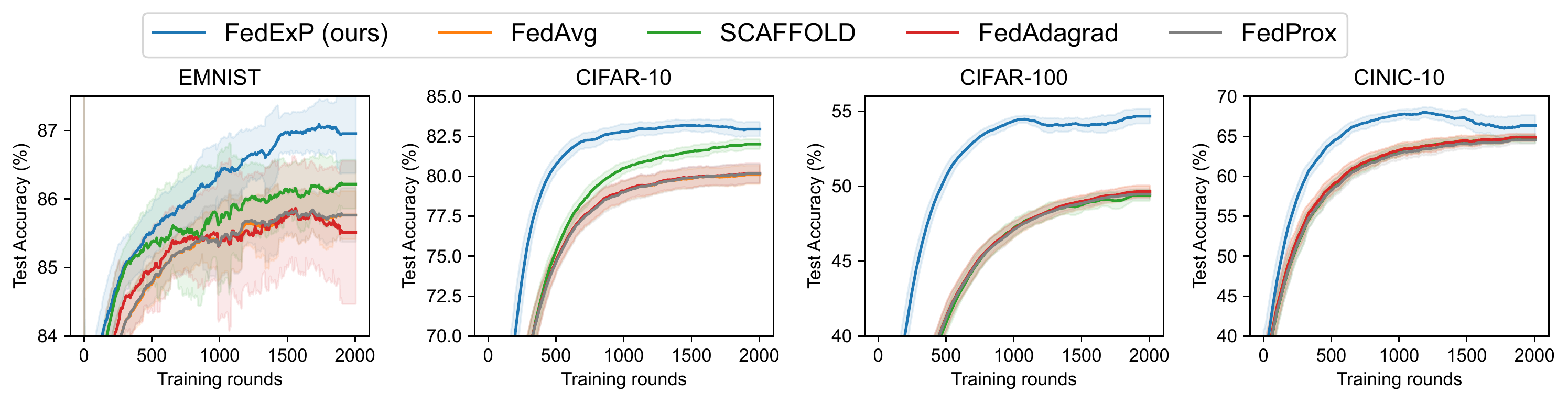}  \caption{Test accuracy results of \texttt{FedExP}, \texttt{FedAvg}, \texttt{SCAFFOLD}, \texttt{FedAdagrad} and \texttt{FedProx} on the EMNIST, CIFAR-10, CIFAR-100 and CINIC-10 datasets for $2000$ rounds.}
\label{fig:addtl_results_2000_runs_test_acc}
\end{figure}

\textbf{Long-Term Behavior of Algorithms and Comparison with \texttt{FedProx}.}
To evaluate the long-term behavior of different algorithms, we ran the experiments for $2000$ rounds. Here, we also consider an additional algorithm, namely \texttt{FedProx}, for comparison.  For fair comparison, we have tuned the $\mu$ parameter of \texttt{FedProx} for each dataset, by doing a grid search over the range $\{10^{-3}, 10^{-2},10^{-1},1\}$ as done in the original \texttt{FedProx} paper \citep{li2020federated}.
The results of EMNIST, CIFAR-10, CIFAR-100, and CINIC-10 in Figures~\ref{fig:addtl_results_2000_runs_train_loss}--\ref{fig:addtl_results_2000_runs_test_acc} and \Cref{table:2000rounds} are from experiments with $3$ different random seeds. Except for the synthetic dataset, the plots show mean and standard deviation values across all the random seeds and also over a moving average window of size $20$.

\captionsetup[table]{labelfont={color=black},font={color=black}}

\begin{table}[H]
\centering
\caption{Test accuracy obtained by \texttt{FedExP} and baselines after 2000 rounds of training on various tasks. Results are averaged across 3 random seeds and last $20$ rounds.}\label{table:2000rounds}\vspace{-0.5em}
\color{black}
\begin{tabular}{c c c c c c}
\thickhline \\[-0.2cm]
Dataset & \texttt{FedExP} & \texttt{FedAvg} & \texttt{SCAFFOLD} & \texttt{FedAdagrad} & \texttt{FedProx}\\ [0.05cm]
\hline\\[-0.2cm]
EMNIST & $\mathbf{86.96} \pm 0.58$ & $85.78 \pm 0.35$ & $86.22 \pm 0.35$ & $85.53 \pm 1.04$ & $85.77 \pm 0.39$\\
CIFAR-10 & $\mathbf{82.94} \pm 0.42$ & $80.10 \pm 0.56$ & 
$82.02 \pm 0.30$ & $80.21 \pm 0.60$ & $80.16 \pm 0.59$\\
CIFAR-100 & $\mathbf{54.65} \pm 0.49$ & $49.63 \pm 0.37$ & $49.40 \pm 0.38$ & $49.64 \pm 0.39$ & $49.47 \pm 0.31$\\
CINIC-10 & $\mathbf{66.45} \pm 1.28$ & $64.87 \pm 0.44$ & $64.61 \pm 0.49$ &  $64.87 \pm 0.44$ & $64.52 \pm 0.45$\\
\thickhline\\[-0.2cm]
\end{tabular}
\end{table}

We see that \texttt{FedExP} continues to outperform baselines including \texttt{FedProx} in the long-term behavior as well. 
\clearpage

\color{black}

\section{Combining Extrapolation with \texttt{SCAFFOLD}}
\label{app:sec_scaffold_exp}

As described in \Cref{sec:experiments}, the extrapolation step can be added to the \texttt{SCAFFOLD} algorithm in a similar way as \texttt{FedExP}. The detailed steps of this \texttt{SCAFFOLD-ExP} algorithm are shown in
\Cref{algo:scaffoldexp}.

\begin{algorithm} [H]
\caption{\texttt{SCAFFOLD-ExP} }\label{algo:scaffoldexp}
\renewcommand{\algorithmicloop}{\textbf{Global server do:}}
\begin{algorithmic}[1]
\STATE {\bfseries Input:} $\bw^{(0)}$, control variate $\bc^{(0)}$, $\bc_i^{(0)}, \forall i\in[M]$, number of rounds $T$, local iteration steps $\tau$, parameters $\eta_l,\epsilon$
\STATE {\bfseries For ${t=0,\ldots,T-1}$ communication rounds do}:
\STATE \hspace*{1em} {\bfseries Global server do:}\\
\STATE \hspace*{1em} Send $\bw^{(t)}$, $\bc^{(t)}$ to all clients
\STATE \hspace*{1em} {\bfseries Clients $i \in [M]$ in parallel do:}
\STATE \hspace*{2em} {Set $\bw_i^{(t,0)}\leftarrow \bw^{(t,0)}$}
\STATE {\hspace*{2em} {\bfseries For $k=0,\ldots,\tau-1$ local iterations do:}}\\
\STATE {\hspace*{3em} Update $\bw_i^{(t,k+1)}\leftarrow\bw_i^{(t,k)}-\eta_l\left(\nabla F_i(\bw_i^{(t,k)},\xi_i^{(t,k)}) - \bc_i^{(t)} + \bc^{(t)}\right)$} \\
\STATE {\hspace*{2em} Compute $\Delta_i^{(t)}\leftarrow \bw^{(t)}-\bw_i^{(t,\tau)}$ and $\Psi_i^{(t)}\leftarrow \bc^{(t)} - \frac{1}{\tau \eta_l} \Delta_i^{(t)}$}
\STATE {\hspace*{2em} Send $\Delta_i^{(t)}$ and $\Psi_i^{(t)}$ to the server}
\STATE {\hspace*{2em} Update local control variate $\bc_i^{(t+1)}\leftarrow \bc_i^{(t)} - \Psi_i^{(t)}$}
\STATE {\hspace*{1em} {\bfseries Global server do:}}\\
\STATE {\hspace*{2em} \tikzmk{A}Compute $\bardeltat \!\leftarrow\!  \frac{1}{M}\sum_{i=1}^M \Delta_i^{(t)}$ and $\eta_g^{(t)} \!\leftarrow\!  \max\left\{1, \sum_{i=1}^M \normsmall{\Delta_i^{(t)}}\!\!\Big/2M\!\left(\norm{\bardeltat}\!+\!\epsilon \right) \!\right\}$}
\STATE {\hspace*{2em} Update global model with $\bw^{(t+1)}\leftarrow \bw^{(t)}-\eta_g^{(t)}\bardeltat$} \hfill \tikzmk{B} \boxit{pink}
\STATE {\hspace*{2em} Compute $\bar{\Psi}^{(t)}\leftarrow \frac{1}{M}\sum_{i=1}^M \Psi_i^{(t)}$} 
\STATE {\hspace*{2em} Update global control variate with $\bc^{(t+1)}\leftarrow \bc^{(t)}-\bar{\Psi}^{(t)}$} 
\end{algorithmic} 
\end{algorithm} 

\clearpage

\color{black}

\section{Combining Extrapolation with Server Momentum}
\label{app:sec_server_momentum}

We begin by recalling some notation from our work. The vector $\mathbf{w}^{(t)}$ is the global model at round $t$ and $\bar{\Delta}^{(t)}$ is the average of client updates at round $t$. The server momentum update at round $t$ can be written as $\mathbf{v}^{(t)} = \bar{\Delta}^{(t)} + \beta \mathbf{v}^{(t-1)}$ (let $\mathbf{v}^{-1} = \mathbf{0})$ and the global model update can be written as $\mathbf{w}^{(t+1)} = \mathbf{w}^{(t)}-\eta_g^{(t)}\mathbf{v}^{(t)}$. Our goal is now to find $\eta_g^{(t)}$ that minimizes $\norm{\bw^{(t+1)}-\bw^*}$. We have,
\begin{align}
  \norm{\bw^{(t+1)}-\bw^*} = \norm{\bw^{(t)}-\bw^*} + (\eta_g^{(t)})^2\norm{\bv^{(t)}} - 2\eta_g^{(t)}\langle \bw^{(t)}-\bw^*,\bv^{(t)} \rangle .
  \label{eq:mom_dist}
\end{align}
Setting the derivative of the RHS of \cref{eq:mom_dist} to zero we have,
\begin{align}(\eta_g^{(t)})_{\text{opt}} = \frac{\left\langle \bw^{(t)}-\bw^*,\bv^{(t)} \right\rangle }{\norm{\bv^{(t)}}}.
\end{align}

Our goal now is to find a lower bound on $\langle \bw^{(t)}-\bw^*,\bv^{(t)} \rangle$. We have the following lemma.

\begin{lem}
Assume that $\left\langle \bw^{(t)}-\bw^*,\bar{\Delta}^{(t)} \right\rangle \geq m^{(t)} = \sum_{i=1}^M \norm{\Delta_i^{(t)}}/M$ (see \Cref{app:subsec_linear_regression_lb}) for all $t \geq 0$ and $\eta_g^{(r)} \leq (m^{(r)} + \sum_{k=0}^{r-1}(\beta/2)^{r-k}m^{(k)})/2\norm{\bv^{(r)}}$ for all $r < t-1$. Then,
\begin{align}
   \left\langle \bw^{(t)}-\bw^*,\bv^{(t)} \right\rangle \geq m^{(t)} + \sum_{k=0}^{t-1}(\beta/2)^{t-k}m^{(k)},
\end{align}
which implies,
\begin{align}
(\eta_g^{(t)})_{\text{opt}} \geq \frac{m^{(t)} + \sum_{k=0}^{t-1}(\beta/2)^{t-k}m^{(k)}}{2\norm{\bv^{(t)}}}.
\end{align}
\end{lem}

\textbf{Proof.} We proceed via a proof by induction. The statement clearly holds at $t=0$ since $\left\langle \bw^{(0)}-\bw^*,\bv^{(0)} \right\rangle = \left\langle \bw^{(0)}-\bw^*,\bar{\Delta}^{(0)} \right\rangle \geq m^{(0)} $.

Now assuming the lemma holds at $t-1$ we have,
\begin{align}
  \left\langle \bw^{(t)}-\bw^*,\bv^{(t)} \right\rangle &=    \left\langle \bw^{(t)}-\bw^*,\bar{\Delta}^{(t)} \right\rangle + \beta \left\langle \bw^{(t)}-\bw^*,\bv^{(t-1)} \right\rangle \\
  & =    \left\langle \bw^{(t)}-\bw^*,\bar{\Delta}^{(t)} \right\rangle + \beta \left\langle \bw^{(t-1)}-\eta_g^{(t-1)}\bv^{(t-1)} -\bw^*,\bv^{(t-1)} \right\rangle \\
  & \geq m^{(t)} + \beta \left[\left\langle \bw^{(t-1)} - \bw^*, \bv^{(t-1)} \right\rangle - \eta_g^{(t-1)} \norm{\bv^{(t-1)}} \right]\\
  & \geq m^{(t)} + \sum_{k=0}^{t-1}(\beta/2)^{t-k}m^{(k)},
\end{align}
where the last line follows from the fact that $\left\langle \bw^{(t-1)} - \bw^*, \bv^{(t-1)} \right\rangle \geq m^{(t-1)} + \sum_{k=0}^{t-2}(\beta/2)^{t-1-k}m^{(k)}$ and $\eta_g^{(t-1)} \leq (m^{(t-1)} + \sum_{k=0}^{t-2}(\beta/2)^{t-1-k}m^{(k)})/2\norm{\bv^{(t-1)}}$.
\qed

Thus we propose to keep the following server step size when using server momentum,
\begin{align}
    \eta_g^{(t)} = \frac{m^{(t)} + \sum_{k=0}^{t-1}(\beta/2)^{t-k}m^{(k)}}{2(\norm{\bv^{(t)}}+\epsilon)},
\end{align}
where $m^{(t)} = \sum_{i=1}^M \norm{\Delta_i^{(t)}}/M$. Note that we also add a small constant $\epsilon$ to the denominator to prevent the step size from blowing up as done for \texttt{FedExP}. We call server momentum with this step size as \texttt{FedExP-M}.

\begin{figure}[H]
 \centering
 \includegraphics[width=0.7\linewidth]{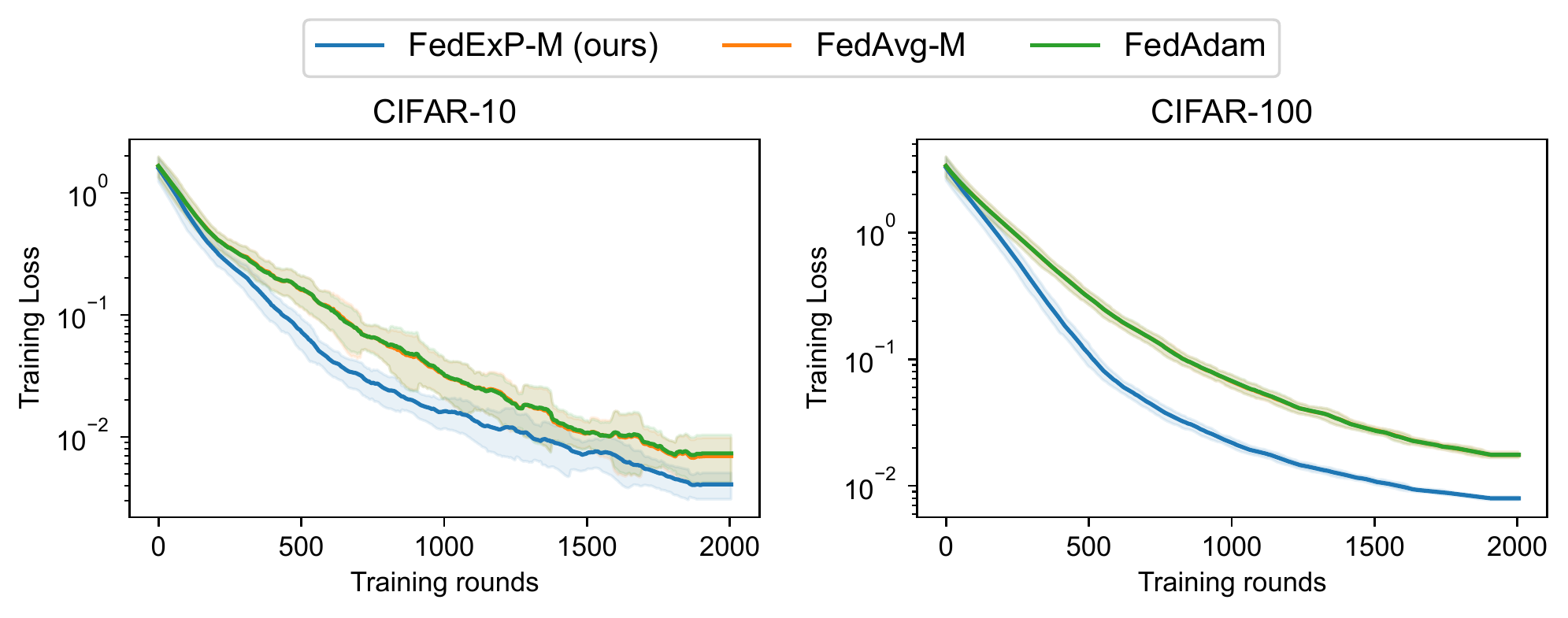}  \caption{Training loss results of \texttt{FedExP-M}, \texttt{FedAdam} and \texttt{FedAvg-M} on the CIFAR10 and CIFAR100 datasets.}
 \vspace{-1em}
\label{fig:addtl_results_momentum_train_loss}
\end{figure}

\begin{figure}[H]
 \centering
 \includegraphics[width=0.7\linewidth]{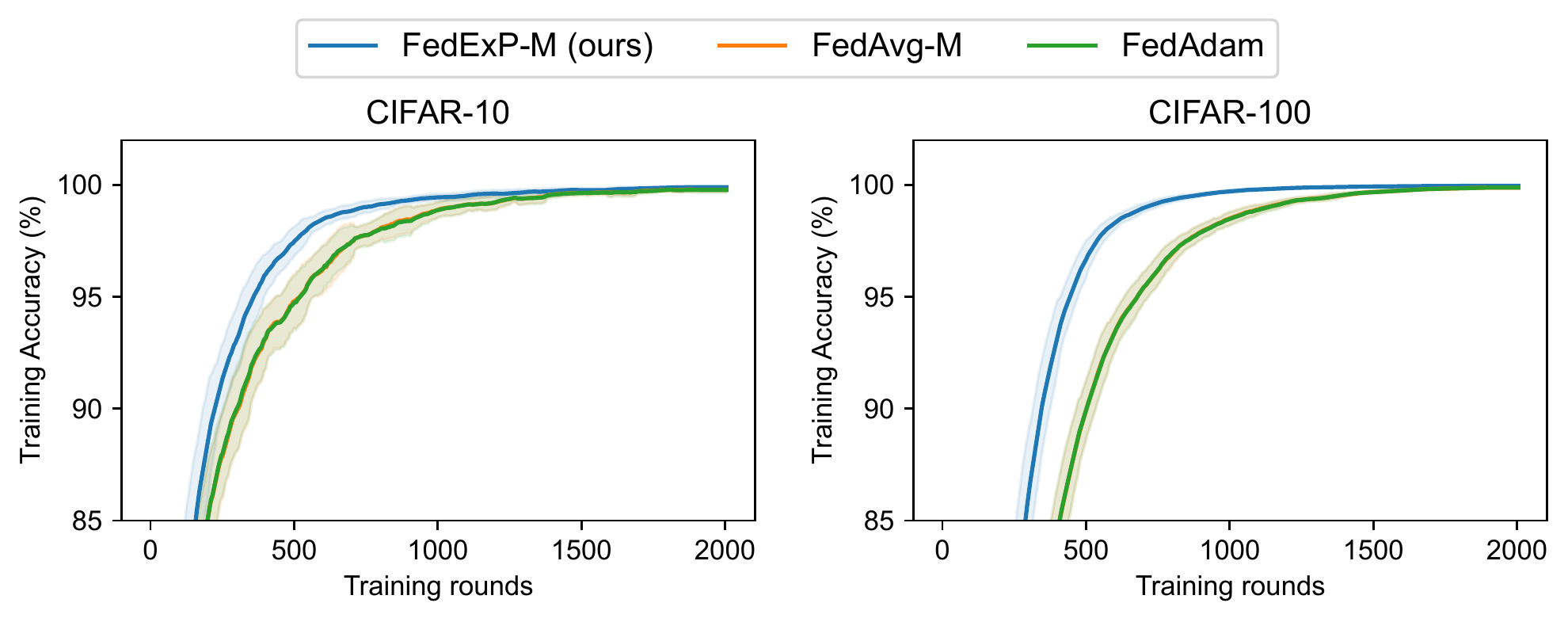}  \caption{Training accuracy results of \texttt{FedExP-M}, \texttt{FedAdam} and \texttt{FedAvg-M} on the CIFAR10 and CIFAR100 datasets.}
 \vspace{-1em}
\label{fig:addtl_results_momentum_train_acc}
\end{figure}

\begin{figure}[H]
 \centering
 \includegraphics[width=0.7\linewidth]{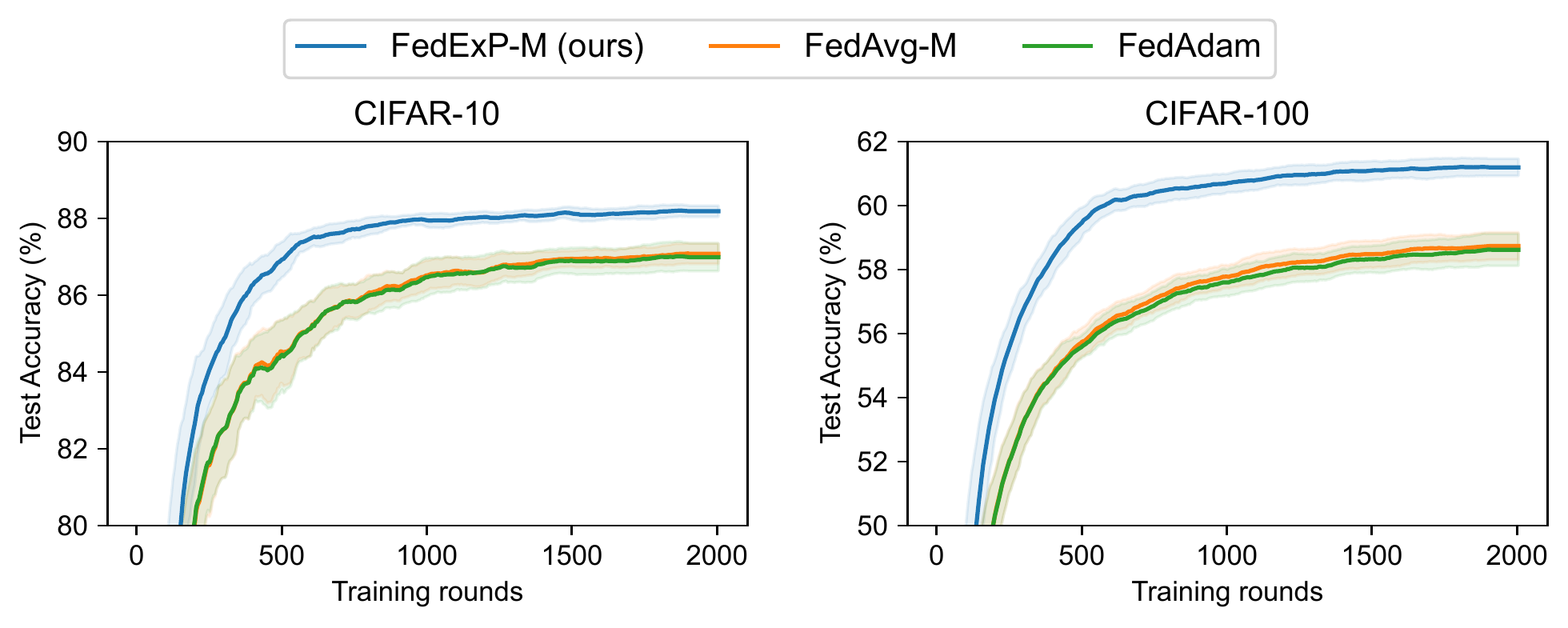}  \caption{Test accuracy results of \texttt{FedExP-M}, \texttt{FedAdam} and \texttt{FedAvg-M} on the CIFAR10 and CIFAR100 datasets.}
 \vspace{-1em}
\label{fig:addtl_results_momentum_test_acc}
\end{figure}

We compare the performance of \texttt{FedExP-M} with \texttt{FedAdam} and \texttt{FedAvg-M} (\texttt{FedAvg} with server momentum) on the CIFAR-10 and CIFAR-100 datasets as shown in Figures~\ref{fig:addtl_results_momentum_train_loss}--\ref{fig:addtl_results_momentum_test_acc}, where the mean and standard deviation values are computed over $3$ random seeds and a moving average window of size $20$. The experimental setup is the same as described in \Cref{sec:experiments}. The hyperparameters $\eta_l,\epsilon$ for \texttt{FedExP-M} and $\eta_l,\eta_g$ for \texttt{FedAdam} and \texttt{FedAvg-M} were tuned following a similar process as described in \Cref{app:subsec_additional_experiments}, and their resulting values are in \Cref{table:momentumHyperParams}. 

\captionsetup[table]{labelfont={color=black},font={color=black}}

\begin{table}[H]
\caption{Base-10 logarithm of the best combination of $\epsilon$ and $\eta_l$ for \texttt{FedExP-M} and combination of $\eta_l$ and $\eta_g$ for \texttt{FedAdam} and \texttt{FedAvg-M}.}
\label{table:momentumHyperParams}
\centering
\small
\color{black}
  \begin{tabular}{lSSSSSSSS}
    \toprule
    \multirow{2}{*}{Dataset} &
      \multicolumn{2}{c}{ \texttt{FedExP}} &
      \multicolumn{2}{c}{\texttt{FedAdam}} &
      \multicolumn{2}{c}{\texttt{FedAvgm-M}}\\
      & {$\epsilon$} & {$\eta_l$} & {$\eta_g$} & {$\eta_l$} & {$\eta_g$} & {$\eta_l$} &\\
      \midrule
    CIFAR-10 & -3 & -2 &  -2 & -2 & 0 & -2 \\
    CIFAR-100 & -3 & -2 &  -2 & -2 & 0 & -2\\
    \bottomrule
  \end{tabular}
\end{table}

Our result shows that server momentum can be successfully combined with extrapolation for the best speed-up among all baselines. The behavior of \texttt{FedAdam} and \texttt{FedAvg-M} are quite similar in these experiments which can be attributed to the dense nature of the gradients in image classification as discussed in \Cref{sec:experiments}. We note that this is only a preliminary result and future work will look to study the effect of combining server momentum and extrapolation more rigorously.

%% file: main.bbl
\begin{thebibliography}{53}
\providecommand{\natexlab}[1]{#1}
\providecommand{\url}[1]{\texttt{#1}}
\expandafter\ifx\csname urlstyle\endcsname\relax
  \providecommand{\doi}[1]{doi: #1}\else
  \providecommand{\doi}{doi: \begingroup \urlstyle{rm}\Url}\fi

\bibitem[Acar et~al.(2021)Acar, Zhao, Matas, Mattina, Whatmough, and
  Saligrama]{acar2021federated}
Durmus Alp~Emre Acar, Yue Zhao, Ramon Matas, Matthew Mattina, Paul Whatmough,
  and Venkatesh Saligrama.
\newblock Federated learning based on dynamic regularization.
\newblock In \emph{International Conference on Learning Representations}, 2021.

\bibitem[Allen-Zhu et~al.(2019)Allen-Zhu, Li, and Liang]{allen2019learning}
Zeyuan Allen-Zhu, Yuanzhi Li, and Yingyu Liang.
\newblock Learning and generalization in overparameterized neural networks,
  going beyond two layers.
\newblock \emph{Advances in Neural Information Processing Systems}, 32, 2019.

\bibitem[Armijo(1966)]{armijo1966minimization}
Larry Armijo.
\newblock Minimization of functions having {Lipschitz} continuous first partial
  derivatives.
\newblock \emph{Pacific Journal of mathematics}, 16\penalty0 (1):\penalty0
  1--3, 1966.

\bibitem[Arora et~al.(2019)Arora, Du, Hu, Li, Salakhutdinov, and
  Wang]{arora2019exact}
Sanjeev Arora, Simon~S Du, Wei Hu, Zhiyuan Li, Russ~R Salakhutdinov, and
  Ruosong Wang.
\newblock On exact computation with an infinitely wide neural net.
\newblock \emph{Advances in Neural Information Processing Systems}, 32, 2019.

\bibitem[Barzilai \& Borwein(1988)Barzilai and Borwein]{barzilai1988two}
Jonathan Barzilai and Jonathan~M Borwein.
\newblock Two-point step size gradient methods.
\newblock \emph{IMA journal of numerical analysis}, 8\penalty0 (1):\penalty0
  141--148, 1988.

\bibitem[Bonawitz et~al.(2016)Bonawitz, Ivanov, Kreuter, Marcedone, McMahan,
  Patel, Ramage, Segal, and Seth]{bonawitz2016practical}
K.~A. Bonawitz, Vladimir Ivanov, Ben Kreuter, Antonio Marcedone, H.~Brendan
  McMahan, Sarvar Patel, Daniel Ramage, Aaron Segal, and Karn Seth.
\newblock Practical secure aggregation for federated learning on user-held
  data.
\newblock In \emph{NeurIPS Workshop on Private Multi-Party Machine Learning},
  2016.

\bibitem[Boyd \& Dattarro(2003)Boyd and Dattarro]{fundament}
Stephen Boyd and Jon Dattarro.
\newblock Alternating projections, 2003.
\newblock \url{https://web.stanford.edu/class/ee392o/alt_proj.pdf}.

\bibitem[Burdakov et~al.(2019)Burdakov, Dai, and Huang]{burdakov2019stabilized}
Oleg Burdakov, Yu-Hong Dai, and Na~Huang.
\newblock Stabilized {Barzilai-Borwein} method.
\newblock \emph{Journal of Computational Mathematics}, 37\penalty0
  (6):\penalty0 916--936, 2019.

\bibitem[Caldas et~al.(2019)Caldas, Duddu, Wu, Li, Kone{\v{c}}n{\`y}, McMahan,
  Smith, and Talwalkar]{caldas2018leaf}
Sebastian Caldas, Sai Meher~Karthik Duddu, Peter Wu, Tian Li, Jakub
  Kone{\v{c}}n{\`y}, H~Brendan McMahan, Virginia Smith, and Ameet Talwalkar.
\newblock Leaf: A benchmark for federated settings.
\newblock In \emph{Workshop on Federated Learning for Data Privacy and
  Confidentiality}, 2019.

\bibitem[Charles \& Kone{\v{c}}n{\`y}(2020)Charles and
  Kone{\v{c}}n{\`y}]{charles2020outsized}
Zachary Charles and Jakub Kone{\v{c}}n{\`y}.
\newblock On the outsized importance of learning rates in local update methods.
\newblock \emph{arXiv preprint arXiv:2007.00878}, 2020.

\bibitem[Cohen et~al.(2017)Cohen, Afshar, Tapson, and
  Van~Schaik]{cohen2017emnist}
Gregory Cohen, Saeed Afshar, Jonathan Tapson, and Andre Van~Schaik.
\newblock Emnist: Extending mnist to handwritten letters.
\newblock In \emph{2017 International Joint Conference on Neural Networks
  (IJCNN)}, pp.\  2921--2926. IEEE, 2017.

\bibitem[Combettes(1997)]{combettes1997convex}
Patrick~L Combettes.
\newblock Convex set theoretic image recovery by extrapolated iterations of
  parallel subgradient projections.
\newblock \emph{IEEE Transactions on Image Processing}, 6\penalty0
  (4):\penalty0 493--506, 1997.

\bibitem[Darlow et~al.(2018)Darlow, Crowley, Antoniou, and
  Storkey]{darlow2018cinic}
Luke~N Darlow, Elliot~J Crowley, Antreas Antoniou, and Amos~J Storkey.
\newblock {CINIC-10} is not {Imagenet} or {CIFAR-10}.
\newblock \emph{arXiv preprint arXiv:1810.03505}, 2018.

\bibitem[Deng et~al.(2022)Deng, Kamani, and Mahdavi]{deng2022local}
Yuyang Deng, Mohammad~Mahdi Kamani, and Mehrdad Mahdavi.
\newblock Local {SGD} optimizes overparameterized neural networks in polynomial
  time.
\newblock In \emph{International Conference on Artificial Intelligence and
  Statistics}, pp.\  6840--6861. PMLR, 2022.

\bibitem[Duchi et~al.(2011)Duchi, Hazan, and Singer]{duchi2011adaptive}
John Duchi, Elad Hazan, and Yoram Singer.
\newblock Adaptive subgradient methods for online learning and stochastic
  optimization.
\newblock \emph{Journal of Machine Learning Research}, 12\penalty0 (7), 2011.

\bibitem[Goldstein(1977)]{goldstein1977optimization}
AA~Goldstein.
\newblock Optimization of {Lipschitz} continuous functions.
\newblock \emph{Mathematical Programming}, 13\penalty0 (1):\penalty0 14--22,
  1977.

\bibitem[Gurin et~al.(1967)Gurin, Polyak, and Raik]{gurin1967method}
Leonid~Georgievich Gurin, Boris~Teodorovich Polyak, and {\`E}~V Raik.
\newblock The method of projections for finding the common point of convex
  sets.
\newblock \emph{Zhurnal Vychislitel'noi Matematiki i Matematicheskoi Fiziki},
  7\penalty0 (6):\penalty0 1211--1228, 1967.

\bibitem[Haddadpour \& Mahdavi(2019)Haddadpour and
  Mahdavi]{haddadpour2019convergence}
Farzin Haddadpour and Mehrdad Mahdavi.
\newblock On the convergence of local descent methods in federated learning.
\newblock \emph{arXiv preprint arXiv:1910.14425}, 2019.

\bibitem[Hazan \& Kakade(2019)Hazan and Kakade]{hazan2019revisiting}
Elad Hazan and Sham Kakade.
\newblock Revisiting the {Polyak} step size.
\newblock \emph{arXiv preprint arXiv:1905.00313}, 2019.

\bibitem[He et~al.(2016)He, Zhang, Ren, and Sun]{he2016deep}
Kaiming He, Xiangyu Zhang, Shaoqing Ren, and Jian Sun.
\newblock Deep residual learning for image recognition.
\newblock In \emph{Proceedings of the IEEE Conference on Computer Vision and
  Pattern Recognition}, pp.\  770--778, 2016.

\bibitem[Horv{\'a}th et~al.(2022)Horv{\'a}th, Mishchenko, and
  Richt{\'a}rik]{horvath2022adaptive}
Samuel Horv{\'a}th, Konstantin Mishchenko, and Peter Richt{\'a}rik.
\newblock Adaptive learning rates for faster stochastic gradient methods.
\newblock \emph{arXiv preprint arXiv:2208.05287}, 2022.

\bibitem[Hsu et~al.(2019)Hsu, Qi, and Brown]{hsu2019measuring}
Tzu-Ming~Harry Hsu, Hang Qi, and Matthew Brown.
\newblock Measuring the effects of non-identical data distribution for
  federated visual classification.
\newblock \emph{arXiv preprint arXiv:1909.06335}, 2019.

\bibitem[Huang et~al.(2021)Huang, Li, Song, and Yang]{huang2021fl}
Baihe Huang, Xiaoxiao Li, Zhao Song, and Xin Yang.
\newblock {FL-NTK}: A neural tangent kernel-based framework for federated
  learning analysis.
\newblock In \emph{International Conference on Machine Learning}, pp.\
  4423--4434. PMLR, 2021.

\bibitem[Jacot et~al.(2018)Jacot, Gabriel, and Hongler]{jacot2018neural}
Arthur Jacot, Franck Gabriel, and Cl{\'e}ment Hongler.
\newblock Neural tangent kernel: Convergence and generalization in neural
  networks.
\newblock \emph{Advances in Neural Information Processing Systems}, 31, 2018.

\bibitem[Johnson et~al.(2020)Johnson, Agrawal, Gu, and
  Guestrin]{johnson2020adascale}
Tyler Johnson, Pulkit Agrawal, Haijie Gu, and Carlos Guestrin.
\newblock Adascale {SGD}: A user-friendly algorithm for distributed training.
\newblock In \emph{International Conference on Machine Learning}, pp.\
  4911--4920. PMLR, 2020.

\bibitem[Kadhe et~al.(2020)Kadhe, Rajaraman, Koyluoglu, and
  Ramchandran]{kadhe2020fastsecagg}
Swanand Kadhe, Nived Rajaraman, O~Ozan Koyluoglu, and Kannan Ramchandran.
\newblock {FastSecAgg}: Scalable secure aggregation for privacy-preserving
  federated learning.
\newblock \emph{arXiv preprint arXiv:2009.11248}, 2020.

\bibitem[Kairouz et~al.(2021)Kairouz, McMahan, Avent, Bellet, Bennis, Bhagoji,
  Bonawitz, Charles, Cormode, Cummings, et~al.]{kairouz2019advances}
Peter Kairouz, H~Brendan McMahan, Brendan Avent, Aur{\'e}lien Bellet, Mehdi
  Bennis, Arjun~Nitin Bhagoji, Kallista Bonawitz, Zachary Charles, Graham
  Cormode, Rachel Cummings, et~al.
\newblock Advances and open problems in federated learning.
\newblock \emph{Foundations and Trends{\textregistered} in Machine Learning},
  14\penalty0 (1--2):\penalty0 1--210, 2021.

\bibitem[Karimireddy et~al.(2019)Karimireddy, Rebjock, Stich, and
  Jaggi]{karimireddy2019error}
Sai~Praneeth Karimireddy, Quentin Rebjock, Sebastian Stich, and Martin Jaggi.
\newblock Error feedback fixes {S}ign{SGD} and other gradient compression
  schemes.
\newblock In \emph{Proceedings of the 36th International Conference on Machine
  Learning}, volume~97, pp.\  3252--3261. PMLR, 2019.

\bibitem[Karimireddy et~al.(2020{\natexlab{a}})Karimireddy, Jaggi, Kale, Mohri,
  Reddi, Stich, and Suresh]{karimireddy2020mime}
Sai~Praneeth Karimireddy, Martin Jaggi, Satyen Kale, Mehryar Mohri, Sashank~J
  Reddi, Sebastian~U Stich, and Ananda~Theertha Suresh.
\newblock Mime: Mimicking centralized stochastic algorithms in federated
  learning.
\newblock \emph{arXiv preprint arXiv:2008.03606}, 2020{\natexlab{a}}.

\bibitem[Karimireddy et~al.(2020{\natexlab{b}})Karimireddy, Kale, Mohri, Reddi,
  Stich, and Suresh]{karimireddy2020scaffold}
Sai~Praneeth Karimireddy, Satyen Kale, Mehryar Mohri, Sashank Reddi, Sebastian
  Stich, and Ananda~Theertha Suresh.
\newblock Scaffold: Stochastic controlled averaging for federated learning.
\newblock In \emph{International Conference on Machine Learning}, pp.\
  5132--5143. PMLR, 2020{\natexlab{b}}.

\bibitem[Khaled et~al.(2020)Khaled, Mishchenko, and
  Richt{\'a}rik]{khaled2020tighter}
Ahmed Khaled, Konstantin Mishchenko, and Peter Richt{\'a}rik.
\newblock Tighter theory for local sgd on identical and heterogeneous data.
\newblock In \emph{International Conference on Artificial Intelligence and
  Statistics}, pp.\  4519--4529. PMLR, 2020.

\bibitem[Kingma \& Ba(2015)Kingma and Ba]{kingma2014adam}
Diederik~P. Kingma and Jimmy Ba.
\newblock Adam: A method for stochastic optimization.
\newblock In \emph{ICLR (Poster)}, 2015.
\newblock URL \url{http://arxiv.org/abs/1412.6980}.

\bibitem[Krizhevsky et~al.(2009)Krizhevsky, Hinton,
  et~al.]{krizhevsky2009learning}
Alex Krizhevsky, Geoffrey Hinton, et~al.
\newblock Learning multiple layers of features from tiny images.
\newblock 2009.

\bibitem[Li et~al.(2020)Li, Sahu, Zaheer, Sanjabi, Talwalkar, and
  Smith]{li2020federated}
Tian Li, Anit~Kumar Sahu, Manzil Zaheer, Maziar Sanjabi, Ameet Talwalkar, and
  Virginia Smith.
\newblock Federated optimization in heterogeneous networks.
\newblock \emph{Proceedings of Machine Learning and Systems}, 2:\penalty0
  429--450, 2020.

\bibitem[Loizou et~al.(2021)Loizou, Vaswani, Laradji, and
  Lacoste-Julien]{loizou2021stochastic}
Nicolas Loizou, Sharan Vaswani, Issam~Hadj Laradji, and Simon Lacoste-Julien.
\newblock Stochastic polyak step-size for sgd: An adaptive learning rate for
  fast convergence.
\newblock In \emph{International Conference on Artificial Intelligence and
  Statistics}, pp.\  1306--1314. PMLR, 2021.

\bibitem[Malinovsky et~al.(2022)Malinovsky, Mishchenko, and
  Richt{\'a}rik]{malinovsky2022server}
Grigory Malinovsky, Konstantin Mishchenko, and Peter Richt{\'a}rik.
\newblock Server-side stepsizes and sampling without replacement provably help
  in federated optimization.
\newblock \emph{arXiv preprint arXiv:2201.11066}, 2022.

\bibitem[Malitsky \& Mishchenko(2020)Malitsky and
  Mishchenko]{malitsky2019adaptive}
Yura Malitsky and Konstantin Mishchenko.
\newblock Adaptive gradient descent without descent.
\newblock In \emph{Proceedings of the 37th International Conference on Machine
  Learning}, volume 119 of \emph{PMLR}, pp.\  6702--6712, 2020.

\bibitem[Mandel(1984)]{mandel1984convergence}
Jan Mandel.
\newblock Convergence of the cyclical relaxation method for linear
  inequalities.
\newblock \emph{Mathematical programming}, 30\penalty0 (2):\penalty0 218--228,
  1984.

\bibitem[McMahan et~al.(2017)McMahan, Moore, Ramage, Hampson, and
  y~Arcas]{mcmahan2017communication}
Brendan McMahan, Eider Moore, Daniel Ramage, Seth Hampson, and Blaise~Aguera
  y~Arcas.
\newblock Communication-efficient learning of deep networks from decentralized
  data.
\newblock In \emph{Artificial Intelligence and Statistics}, pp.\  1273--1282.
  PMLR, 2017.

\bibitem[Mishchenko et~al.(2022)Mishchenko, Malinovsky, Stich, and
  Richtarik]{mishchenko2022proxskip}
Konstantin Mishchenko, Grigory Malinovsky, Sebastian Stich, and Peter
  Richtarik.
\newblock {P}rox{S}kip: Yes! {L}ocal gradient steps provably lead to
  communication acceleration! {F}inally!
\newblock In \emph{Proceedings of the 39th International Conference on Machine
  Learning}, volume 162, pp.\  15750--15769. PMLR, 2022.

\bibitem[Mitra et~al.(2021)Mitra, Jaafar, Pappas, and Hassani]{mitra2021linear}
Aritra Mitra, Rayana Jaafar, George~J Pappas, and Hamed Hassani.
\newblock Linear convergence in federated learning: Tackling client
  heterogeneity and sparse gradients.
\newblock \emph{Advances in Neural Information Processing Systems},
  34:\penalty0 14606--14619, 2021.

\bibitem[Pierra(1984)]{pierra1984decomposition}
Guy Pierra.
\newblock Decomposition through formalization in a product space.
\newblock \emph{Mathematical Programming}, 28\penalty0 (1):\penalty0 96--115,
  1984.

\bibitem[Polyak(1969)]{POLYAK196914}
Boris~Teodorovich Polyak.
\newblock Minimization of unsmooth functionals.
\newblock \emph{USSR Computational Mathematics and Mathematical Physics},
  9\penalty0 (3):\penalty0 14--29, 1969.
\newblock ISSN 0041-5553.
\newblock \doi{https://doi.org/10.1016/0041-5553(69)90061-5}.

\bibitem[Raydan(1993)]{raydan1993barzilai}
Marcos Raydan.
\newblock On the barzilai and borwein choice of steplength for the gradient
  method.
\newblock \emph{IMA Journal of Numerical Analysis}, 13\penalty0 (3):\penalty0
  321--326, 1993.

\bibitem[Reddi et~al.(2021)Reddi, Charles, Zaheer, Garrett, Rush,
  Kone{\v{c}}n{\'y}, Kumar, and McMahan]{reddi2020adaptive}
Sashank~J. Reddi, Zachary Charles, Manzil Zaheer, Zachary Garrett, Keith Rush,
  Jakub Kone{\v{c}}n{\'y}, Sanjiv Kumar, and Hugh~Brendan McMahan.
\newblock Adaptive federated optimization.
\newblock In \emph{International Conference on Learning Representations}, 2021.

\bibitem[Tieleman et~al.(2012)Tieleman, Hinton, et~al.]{tieleman2012lecture}
Tijmen Tieleman, Geoffrey Hinton, et~al.
\newblock Lecture 6.5-{RMSProp}: Divide the gradient by a running average of
  its recent magnitude.
\newblock \emph{COURSERA: Neural networks for machine learning}, 4\penalty0
  (2):\penalty0 26--31, 2012.

\bibitem[Wang et~al.(2020)Wang, Liu, Liang, Joshi, and Poor]{wang2020tackling}
Jianyu Wang, Qinghua Liu, Hao Liang, Gauri Joshi, and H~Vincent Poor.
\newblock Tackling the objective inconsistency problem in heterogeneous
  federated optimization.
\newblock \emph{Advances in Neural Information Processing Systems},
  33:\penalty0 7611--7623, 2020.

\bibitem[Yin et~al.(2018)Yin, Pananjady, Lam, Papailiopoulos, Ramchandran, and
  Bartlett]{yin2018gradient}
Dong Yin, Ashwin Pananjady, Max Lam, Dimitris Papailiopoulos, Kannan
  Ramchandran, and Peter Bartlett.
\newblock Gradient diversity: a key ingredient for scalable distributed
  learning.
\newblock In \emph{International Conference on Artificial Intelligence and
  Statistics}, pp.\  1998--2007. PMLR, 2018.

\bibitem[Yu et~al.(2022)Yu, Wei, Karimireddy, Ma, and Jordan]{yu2022tct}
Yaodong Yu, Alexander Wei, Sai~Praneeth Karimireddy, Yi~Ma, and Michael~I
  Jordan.
\newblock {TCT}: Convexifying federated learning using bootstrapped neural
  tangent kernels.
\newblock \emph{arXiv preprint arXiv:2207.06343}, 2022.

\bibitem[Yue et~al.(2022)Yue, Jin, Pilgrim, Wong, Baron, and
  Dai]{yue2022neural}
Kai Yue, Richeng Jin, Ryan Pilgrim, Chau-Wai Wong, Dror Baron, and Huaiyu Dai.
\newblock Neural tangent kernel empowered federated learning.
\newblock In \emph{International Conference on Machine Learning}, pp.\
  25783--25803. PMLR, 2022.

\bibitem[Zeiler(2012)]{zeiler2012adadelta}
Matthew~D Zeiler.
\newblock Adadelta: an adaptive learning rate method.
\newblock \emph{arXiv preprint arXiv:1212.5701}, 2012.

\bibitem[Zhang et~al.(2017)Zhang, Bengio, Hardt, Recht, and
  Vinyals]{zhang2017understanding}
Chiyuan Zhang, Samy Bengio, Moritz Hardt, Benjamin Recht, and Oriol Vinyals.
\newblock Understanding deep learning requires rethinking generalization.
\newblock In \emph{International Conference on Learning Representations}, 2017.

\bibitem[Zhang et~al.(2020)Zhang, Karimireddy, Veit, Kim, Reddi, Kumar, and
  Sra]{zhang2020adaptive}
Jingzhao Zhang, Sai~Praneeth Karimireddy, Andreas Veit, Seungyeon Kim, Sashank
  Reddi, Sanjiv Kumar, and Suvrit Sra.
\newblock Why are adaptive methods good for attention models?
\newblock \emph{Advances in Neural Information Processing Systems},
  33:\penalty0 15383--15393, 2020.

\end{thebibliography}
